\RequirePackage{fix-cm}
\documentclass[twocolumn]{svjour3}          

\smartqed  
\usepackage{graphicx}
\usepackage{comment}
\usepackage{natbib}
\usepackage{amsmath}
\usepackage{amssymb}
\usepackage{amsfonts}
\usepackage{algorithm}
\usepackage{algorithmic}
\usepackage{color}
\usepackage{bbm}
\usepackage[caption=false]{subfig}
\newcommand{\argmin}{\mathop{\rm argmin}\limits}
\newcommand{\figurescale}{0.23}
\newcommand{\figurescaleR}{0.28}
\newlength\myindent
\newcommand\bindent{%
  \begingroup
  \setlength{\itemindent}{\myindent}
  \addtolength{\algorithmicindent}{\myindent}
}
\newcommand\eindent{\endgroup}

\begin{document}

\title{Defogging Kinect: Simultaneous Estimation of Object Region and Depth in Foggy Scenes 
}


\author{Yuki Fujimura \and
Motoharu Sonogashira \and
Masaaki Iiyama
}


\institute{Yuki Fujimura \at
            The Graduate School of Informatics, Kyoto University, Japan \\
            \email{fujimura@mm.media.kyoto-u.ac.jp} \and
            Motoharu Sonogashira \at
            Academic Center for Computing and Media Studies, Kyoto University, Japan \\
            \email{sonogashira@mm.media.kyoto-u.ac.jp} \and
            Masaaki Iiyama \at
            Academic Center for Computing and Media Studies, Kyoto University, Japan \\
            \email{iiyama@mm.media.kyoto-u.ac.jp}
}

\date{Received: date / Accepted: date}

\maketitle

\begin{abstract}
Three-dimensional (3D) reconstruction and scene depth estimation from 2-dimensional (2D) images are major tasks in computer vision.
However, using conventional 3D reconstruction techniques gets challenging in participating media such as murky water, fog, or smoke.
We have developed a method that uses a time-of-flight (ToF) camera to estimate an object region and depth in participating media simultaneously.
The scattering component is saturated, so it does not depend on the scene depth, and received signals bouncing off distant points are negligible due to light attenuation in the participating media, so the observation of such a point contains only a scattering component.
These phenomena enable us to estimate the scattering component in an object region from a background that only contains the scattering component.
The problem is formulated as robust estimation where the object region is regarded as outliers, and it enables the simultaneous estimation of an object region and depth on the basis of an iteratively reweighted least squares (IRLS) optimization scheme.
We demonstrate the effectiveness of the proposed method using captured images from a Kinect v2 in real foggy scenes and evaluate the applicability with synthesized data.
\keywords{time-of-flight \and depth estimation \and participating media \and light scattering \and iteratively reweighted least squares}
\end{abstract}

\section{Introduction}\label{sec:introduction}
\begin{figure}[tb]
\centering
  \includegraphics[width=0.45\textwidth]{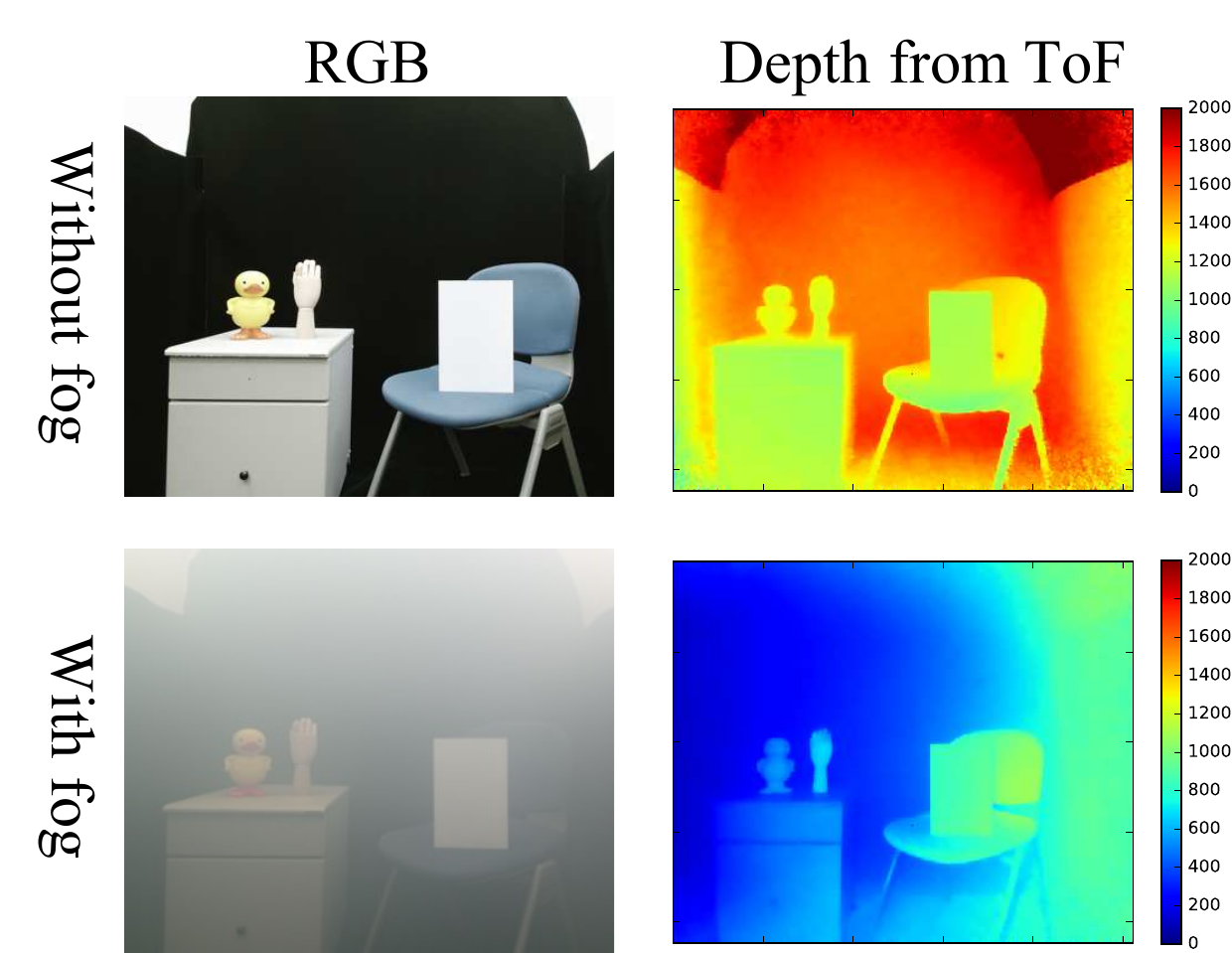}
\caption{Depth error due to light scattering. Depth measurement suffers from scattring effect in participating media such as a foggy scene.}
\label{fig:depth_error}
\end{figure}

Three-dimensional (3D) reconstruction and scene depth estimation from 2-dimensional (2D) images are important tasks in computer vision.
These techniques can be utilized for a variety of applications such as robot vision and self-driving vehicles.
However, using 3D reconstruction techniques gets challenging in participating media such as murky water, fog, or smoke.
As shown at the bottom left of Fig. \ref{fig:depth_error}, the contrast of a captured image in participating media is reduced because light propagating through the media gets attenuated and the received signal contains not only reflected light from the object surface but also scattered light due to suspended particles.
These effects make it difficult to use conventional 3D reconstruction techniques such as structure from motion or shape from shading.

\begin{figure*}[tb]
\centering
  \includegraphics[width=0.7\textwidth]{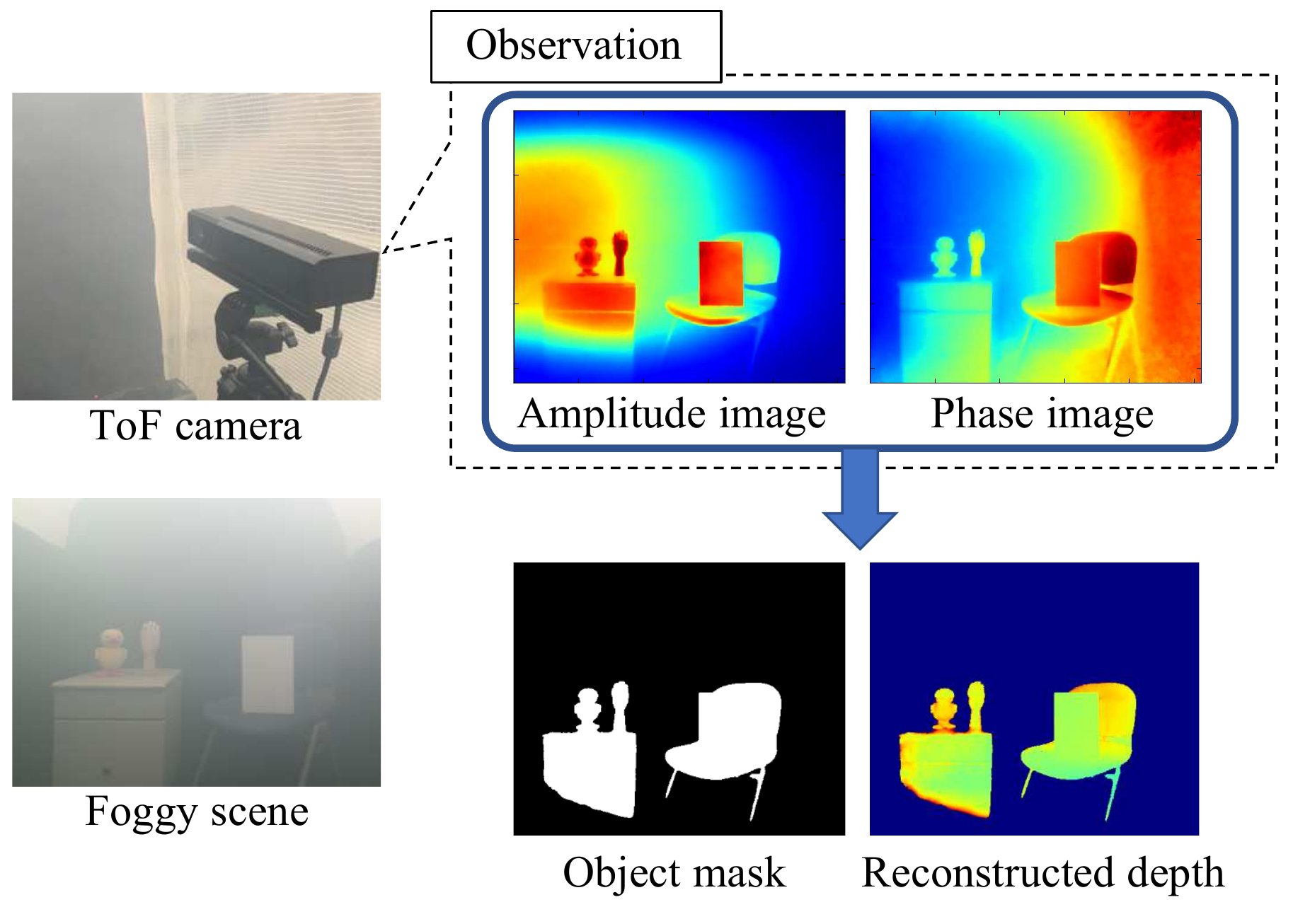}
  \caption{Overview of proposed method. A continuous-wave ToF camera captures an amplitude image and a phase image. From these images captured in participating media, we estimate the object region and recover depth simultaneously.}
\label{fig:overview}
\end{figure*}

In response to the issue above, we have developed a method that enables us to acquire 3D geometry in participating media.
Specifically, we use a time-of-flight (ToF) camera to detect an object and estimate the distance in participating media (see Fig. \ref{fig:overview}).
The proposed method enables simultaneous automatic obstacle detection and depth reconstruction in challenging environments such as disaster sites where light is scattered.

There are several architectures for ToF cameras.
We use Microsoft Kinect v2, a continuous-wave ToF camera that emits a modulated sinusoid signal into a scene and then measures the amplitude of light that bounces off an object surface and the phase shift between the illumination and received signal.
These observations are represented as an amplitude image and a phase image as shown in Fig. \ref{fig:overview}.
Since the phase shift depends on an optical path, we can reconstruct the depth from the phase shift.
In this study, we denote the observation of an object surface by a direct component.

This architecture assumes that each camera pixel observes a single point in a scene.
As mentioned previously, however, the observed signal in participating media includes a scattering component due to light scattering as well as a direct component.
The amplitude and phase shift suffer from the scattering effect, and this causes error of depth measurement (see Fig. \ref{fig:depth_error}).

We aim to recover the direct component, and this is an ill-posed problem because the separation of two components is required.
To deal with this problem, we leverage the saturation of a scattering component and light attenuation in participating media.
Given a near light source in participating media, a scattering component is saturated close to the camera \citep{treibitz09,tsiotsios14}.
This means the scattering component does not depend on an object in the scene if the object is located at a certain distance from the camera.
Moreover, the intensity of light propagating through participating media is attenuated exponentially relative to distance.
Thus, reflected light from a distant point is negligible and the observation of such a point includes only a scattering component.
In this paper, we assume that a target scene consists of an object region and a background that only contains a scattering component.
This scattering component can then be estimated simply by observing the background.
In addition to the above assumption, we introduce two priors to estimate the scattering component: first, a scattering component can be approximated using a quadratic function in a local image patch ({\it local quadratic prior}), and second, a scattering component has a symmetrical characteristic in an overall image ({\it global symmetrical prior}).

The estimation of a scattering component is formulated as robust estimation, where the object region is regarded as outliers because it contains a direct component.
We propose an optimization scheme based on iteratively reweighted least squares (IRLS) \citep{holland77,fox02,chartrand08,wipf10}, which minimizes weighted least squares iteratively.
The object region can then be extracted via the IRLS weight as outliers.

In section \ref{sec:related_work} of this paper, we briefly overview previous studies on computer vision application in participating media.
In section \ref{sec:image_formation}, an image formation model in participating media for ToF measurement is introduced.
In section \ref{sec:simultaneous_estimation_of_object_region_and_depth}, we describe the proposed method for simultaneous estimation of the object region and depth, and in section \ref{sec:experiments}, we present the experimental results.
We conclude the paper in section \ref{sec:conclusion} with a breif summary and mention of future work.

\section{Related work}\label{sec:related_work}
\subsection{Scattering removal}\label{sec:scattering_removal}
In participating media, the contrast of a captured image is reduced due to scattered light and light attenuation.
In computer vision and image processing, several methods have been proposed to recover an original image from a degraded one \citep{he11,nishino12,fattal14,berman16}.
These methods are referred to as defogging or dehazing, and the problem is known to be ill-posed.
Many priors have been introduced to address the ill-posed nature of the problem.
For example, \cite{he11} and \cite{berman16} respectively proposed a dark channel prior and a haze-line prior.
Recently, many learning-based methods using neural networks have also been proposed \citep{cai16,ren16,li17,zhang18,d_yang18}.

\subsection{3D reconstruction in participating media}\label{sec:3d_reconstruction_in_participating_media}
Our goal is to reconstruct a 3D scene in participating media.
There have been several works on the design of 3D reconstruction techniques for participating media.
For example, \cite{li15} proposed simultaneous dehazing and 3D reconstruction using multi-view stereo.
Photometric stereo in participating media has been proposed by \cite{tsiotsios14,murez17,fujimura18}, in which light transport models were built in participating media for photometric stereo application.
Other studies have been based on structured light \citep{narasimhan05,gu13}, light field \citep{tian17}, and the absorption of infrared light in underwater scenes \citep{asano16}.

\subsection{Multipath interference of ToF}\label{sec:multipath_interference_of_tof}
A ToF camera assumes that each camera pixel observes a single point in a scene.
In participating media, however, the measurement also includes scattered light.
This problem is known as multipath interference (MPI).
MPI is caused not just by light scattering in participating media but also by subsurface scattering or interreflection in common scenes.
Thus, many previous studies have tackled MPI compensation \citep{fuchs10, freedman14, naik15, kadambi16, guo18}.

In this paper, we limit our focus to MPI caused by light scattering in participating media.
ToF measurement in participating media has been proposed by \cite{heide14, satat18}.
\cite{heide14} developed a scattering model based on exponentially modified Gaussians for transient imaging using a photonic mixer device (PMD) \citep{heide13}.
\cite{satat18} demonstrated that scattered photons observed with a single photon avalanche diode (SPAD) have gamma distribution and leveraged this observation to separate received photons into a directly reflected component and a scattering component.
Our method differs from these approaches in that we just use an off-the-shelf ToF camera (Kinect v2) with no special hardware modification.

\section{ToF observation in participating media}\label{sec:image_formation}
In this section, we describe our image formation model for a continuous-wave ToF camera in participating media.
As with many previous studies \citep{narasimhan05, narasimhan06, treibitz09, tsiotsios14}, we assume here that forward scattering and multiple scattering are negligible and that the density of a participating medium in a scene is homogeneous.

\begin{figure}[tb]
  \centering
  \includegraphics[width=0.45\textwidth]{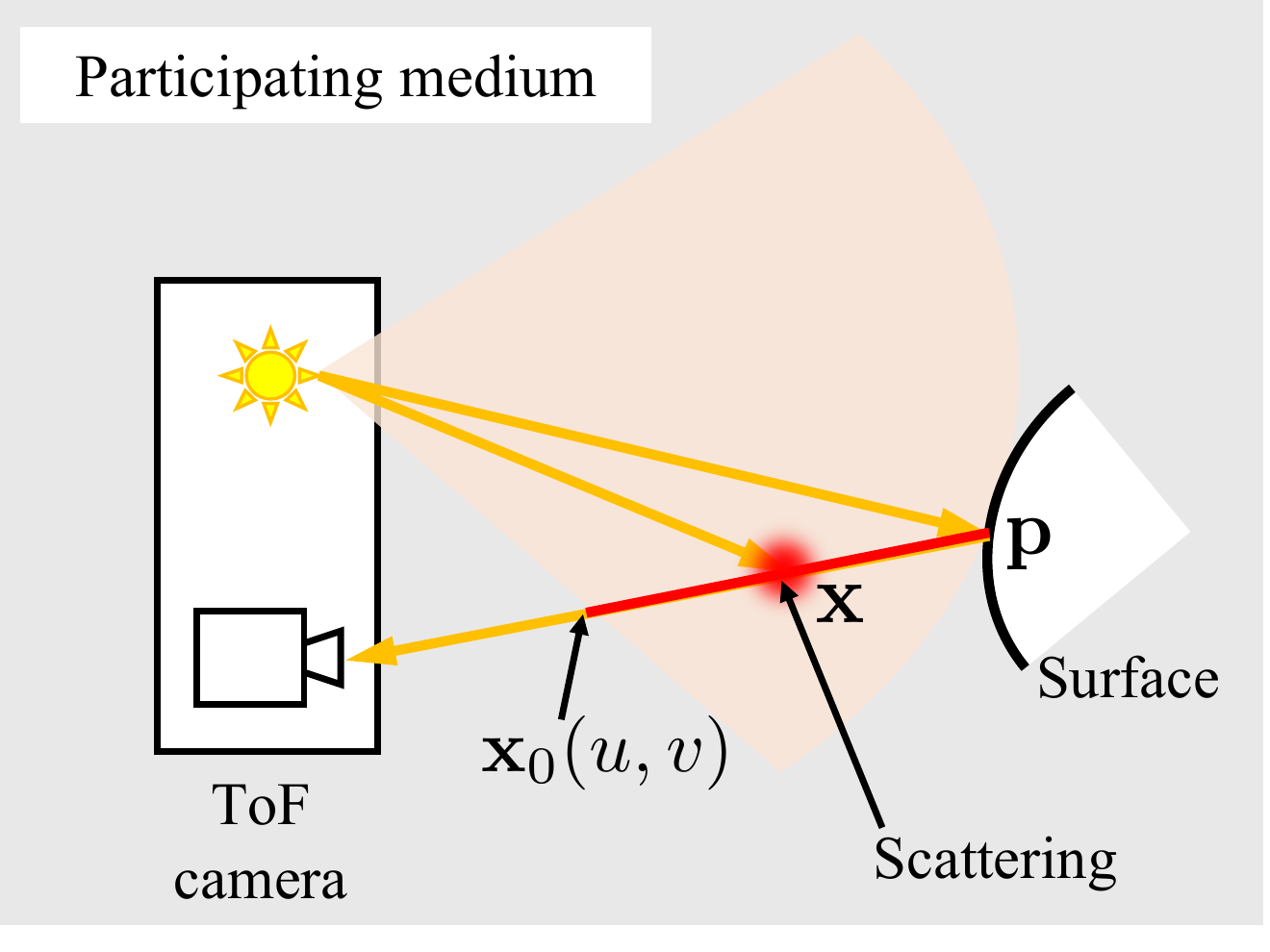}
  \caption{Light scattering in participating media. Light interacts with a participating medium on the line of sight and then arrives at a camera pixel. The total scattering component is the sum of scattered light on the red line in the figure, which depends on the limited beam angle of the light source.}
  \label{fig:image_formation}
\end{figure}

A continuous-wave ToF camera illuminates a scene with amplitude-modulated light and then measures the amplitude of received signal $\alpha$ and phase shift $\varphi$ between the illumination and received signal.
This observation can be described using a phasor \citep{gupta15}, as
\begin{equation}
\alpha e^{j \varphi} \in \mathbb{C}.
\end{equation}
Since the phase shift is proportional to the depth of an object, we can compute the depth as
\begin{equation}
\label{eq:tof}
z = \frac{c \varphi}{4\pi f},
\end{equation}
where $z$ is depth, $c$ is the speed of light, and $f$ is the modulation frequency of the camera.

In participating media, the observation contains scattered light.
As shown in Fig. \ref{fig:image_formation}, light interacts with the medium on the line of sight and then arrives at the camera pixel.
Thus, the observed scattering component is the sum of scattered light on the line of sight.
Now, we consider the 3D coordinate, the origin of which is the camera center.
When the camera observes a surface point $\mathbf{p} \in \mathbb{R}^3$ at a camera pixel $(u,v)$, the total observation $\tilde{\alpha}(u,v;\mathbf{p})e^{j \tilde{\varphi}(u,v;\mathbf{p})}$ can be written as
\begin{eqnarray}
\tilde{\alpha}(u,v;\mathbf{p})e^{j\tilde{\varphi}(u,v;\mathbf{p})} = \alpha_d (u,v;\mathbf{p}) e^{j\varphi_d(u,v;\mathbf{p})} \nonumber \\
+ \int_{\|\mathbf{x}\|=\|\mathbf{x}_0(u,v)\|}^{\|\mathbf{p}\|} \alpha(u,v;\mathbf{x})e^{j\varphi(u,v;\mathbf{x})}d\|\mathbf{x}\| ,
\label{eq:lte}
\end{eqnarray} 
where $\alpha_d(u,v;\mathbf{p})$ and $\varphi_d(u,v;\mathbf{p})$ are the direct components.
$\alpha_d(u,v;\mathbf{p})$ depends on the surface albedo, shading, and attenuation, which is caused by the medium as well as the inverse square law.
$\alpha (u,v;\mathbf{x})e^{j\varphi(u,v;\mathbf{x})}$ is the observation of scattered light at a position $\mathbf{x}$.
Note that although the scattering component can be written using an integral, the domain of the integral (red line in Fig. \ref{fig:image_formation}) depends on the relative position between the light source and camera pixel.
This is because an ideal point light source irradiates a scene with isotropic intensity, while a practical illumination such as a spotlight has a limited beam angle \citep{tsiotsios14}.

Assuming a near light source in participating media, an observed scattering component is saturated close to the camera \citep{treibitz09,tsiotsios14}.
That is, there exists $\mathbf{x}_{saturate}$ for which
\begin{equation}
\|\mathbf{x}\| \geq \|\mathbf{x}_{saturate}\| \Rightarrow \alpha(u,v;\mathbf{x}) = 0.
\end{equation}
Therefore, we can rewrite Eq. (\ref{eq:lte}) as
\begin{eqnarray}
\tilde{\alpha}(u,v;\mathbf{p})e^{j\tilde{\varphi}(u,v;\mathbf{p})} = \alpha_d (u,v;\mathbf{p}) e^{j\varphi_d(u,v;\mathbf{p})}  \nonumber \\
+ \underbrace{\int_{\|\mathbf{x}\|=\|\mathbf{x}_0(u,v)\|}^{\|\mathbf{x}_{saturate}\|} \alpha(u,v;\mathbf{x})e^{j\varphi(u,v;\mathbf{x})}d\|\mathbf{x}\|}_{=\alpha_s(u,v)e^{j\varphi_s(u,v)}},
\label{eq:final_lte}
\end{eqnarray}
where $\alpha_s(u,v)$ and $\varphi_s(u,v)$ are the scattering components, which depend on only the camera pixel $(u,v)$ rather than the object depth.

Although the observation consists of the direct component $\alpha_d(u,v;\mathbf{p})e^{j\varphi_d(u,v;\mathbf{p})}$ and the scattering component $\alpha_s(u,v)e^{j\varphi_s(u,v)}$, the attenuation due to the medium reduces the direct component dramatically.
Thus, if the camera observes a distant point $\mathbf{p}_{far}$, the amplitude of the reflected light fades away, that is,
\begin{equation}
\alpha_d(u,v;\mathbf{p}_{far}) = 0.
\end{equation}
Therefore, the observation of the distant point includes only a scattering component:
\begin{equation}
\tilde{\alpha}(u,v;\mathbf{p}_{far})e^{j\tilde{\varphi}(u,v;\mathbf{p}_{far})} = \alpha_s(u,v)e^{j\varphi_s(u,v)}.
\end{equation} 

\begin{figure}[tb]
\centering
  \subfloat[RGB image]{\includegraphics[width=0.2\textwidth]{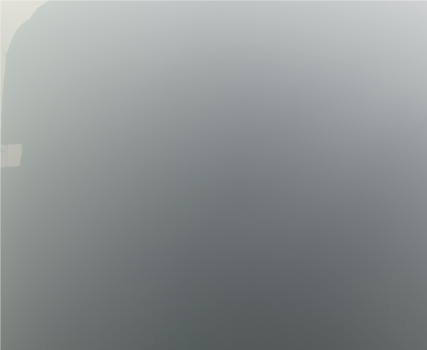}} \\
  \subfloat[Amplitude image]{\includegraphics[width=0.2\textwidth]{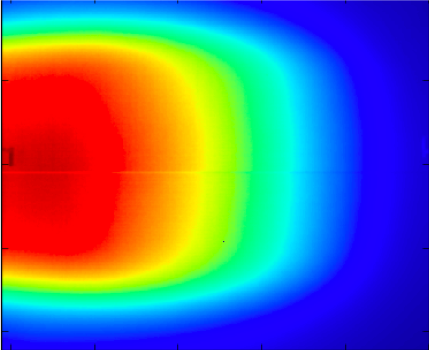}} \\
  \subfloat[Phase image]{\includegraphics[width=0.2\textwidth]{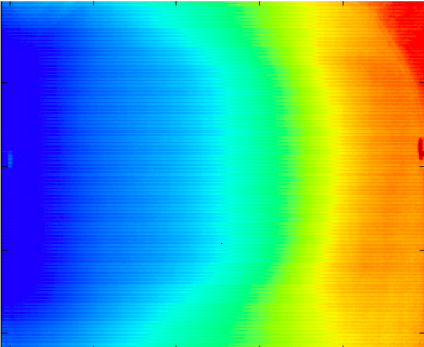}}
\caption{Observation of a black surface in a foggy scene. The black surface approximates a distant observation where only a scattering component can be observed because reflected light from the scene gets attenuated. Note that the observed scattering component is inhomogeneous due to the limited beam angle of the illumination.}
\label{fig:void_observation}
\end{figure}

Figure \ref{fig:void_observation} shows amplitude and phase images when the camera observes a black surface in a foggy scene.
The intensity of reflected light from the black surface is very small, so this approximates a distant observation where only a scattering component can be observed.
As discussed above, in both the amplitude and phase images, the scattering component is inhomogeneous because the illumination has a limited beam angle.

The measurement range of our method is between a saturation point and a background point that has no direct component.
More details about the saturation of the scattering component and the measurement range are provided in section \ref{sec:discussion_of_measurement_range}.

\section{Simultaneous estimation of object region and depth}\label{sec:simultaneous_estimation_of_object_region_and_depth}
As explained in the previous section, a scattering component depends on the position of a camera pixel rather than a target object.
In addition, only the scattering component is observed in the background where an object is farther away.
Thus, our goal is to estimate the scattering component in an object region from the observation of the background.
After estimating scattering components $\alpha_s(u,v)$ and $\varphi_s(u,v)$ at each pixel, we compute the amplitude and phase shift of a direct component from Eq. (\ref{eq:final_lte}):
\begin{flalign}
&\alpha_d = \sqrt{ (\tilde{\alpha}\cos \tilde{\varphi} - \alpha_s\cos \varphi_s)^2 + (\tilde{\alpha}\sin \tilde{\varphi} - \alpha_s\sin \varphi_s)^2 },& \label{eq:amp_recon}\\
&\varphi_d = {\rm arg}\left( (\tilde{\alpha}\cos \tilde{\varphi} - \alpha_s\cos \varphi_s) + j (\tilde{\alpha}\sin \tilde{\varphi} - \alpha_s\sin \varphi_s)\right),&
\end{flalign}
where an operator ${\rm arg}$ returns the argument of a complex number.
Then, depth is recovered substituting the phase into Eq. (\ref{eq:tof}).

In this section, we describe how our method divides camera pixels into an object region and a background, and simultaneously estimates the scattering component in the object region.
First, we introduce two priors to estimate the scattering component, and then the problem is formulated as robust estimation, which allows us to extract the object region as outliers.
In the following, with a slight alteration of notation, we refer to both an amplitude image and a phase image as an image, since we process both images in the same manner.

\subsection{Prior of scattering component} \label{sec:prior_of_scattering_component}
We can estimate the scattering component of an object region from a background because the component does not depend on the object.
\cite{tsiotsios14} approximated backscatter as a quadratic function in a captured image.
Similarly to their work, we also introduce priors, {\it local quadratic prior} and {\it global symmetrical prior}, that allow us to estimate the scattering component. 

\begin{figure}[tb]
  \centering
    \includegraphics[width=0.5\textwidth]{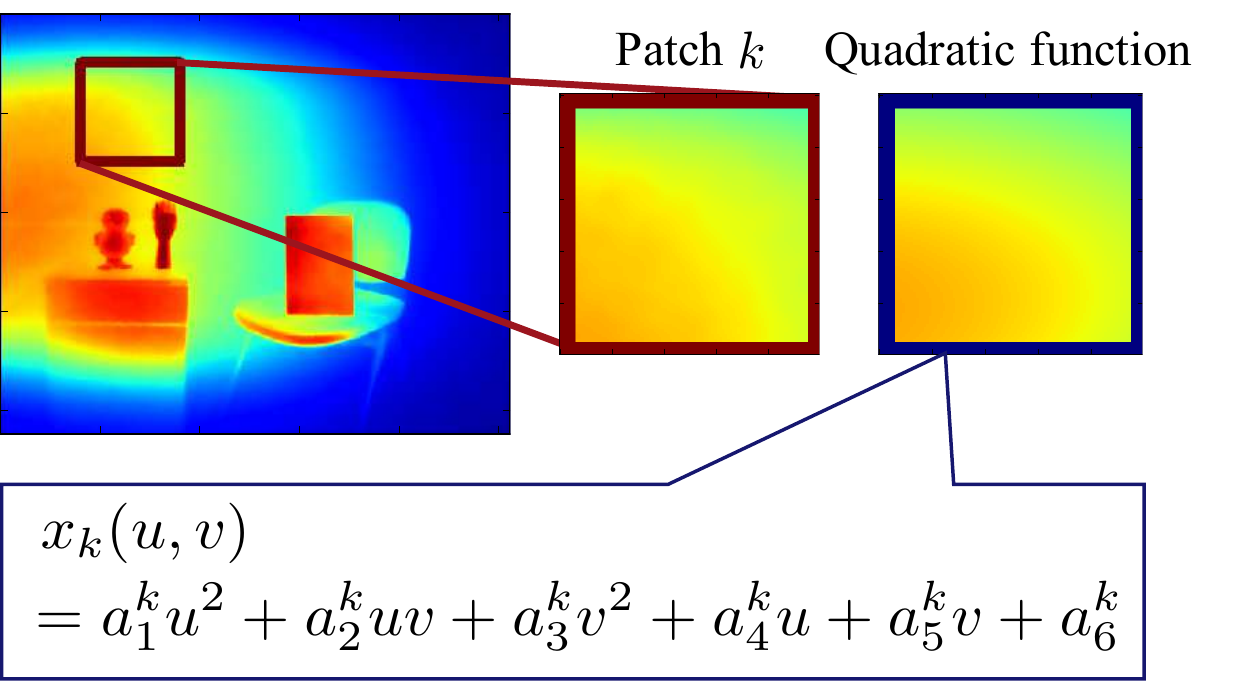}
    \caption{Local quadratic prior. We assume that a scattering component can be represented with a quadratic function in a local image patch.}
    \label{fig:fitting}
\end{figure}

\paragraph{Local quadratic prior.} In our ToF setting, we found that a scattering component cannot be approximated globally with a simple function.
Thus, as shown in Fig. \ref{fig:fitting}, we assume that a scattering component can be represented with a quadratic function in a local image patch, that is,
\begin{eqnarray}
x_k(u,v) &=& a_1^k u^2 + a_2^k uv + a_3^k v^2 + a_4^k u + a_5^k v + a_6^k \nonumber \\ 
&=& \mathbf{a}_k^\top \mathbf{u},
\end{eqnarray}
where $x_k(u,v)$ is the value at a pixel $(u,v)$ in a local image patch $k$.
$\mathbf{u} = [u^2\;\;uv\;\;v^2\;\;u\;\;v\;\;1]^\top$ is a 6-dimensional vector and $\mathbf{a}_k=[a_1^k\;\;a_2^k\;\;a_3^k\;\;a_4^k\;\;a_5^k\;\;a_6^k]^\top$ denotes the coefficients of the quadratic function in patch $k$.

\paragraph{Global symmetrical prior.} However, this local prior is not useful when there exists a large object region and a quadratic function is also fitted into the values in that region.
To address this problem, we introduce a global prior to the scattering component.

\begin{figure}[tb]
  \centering
    \includegraphics[width=0.5\textwidth]{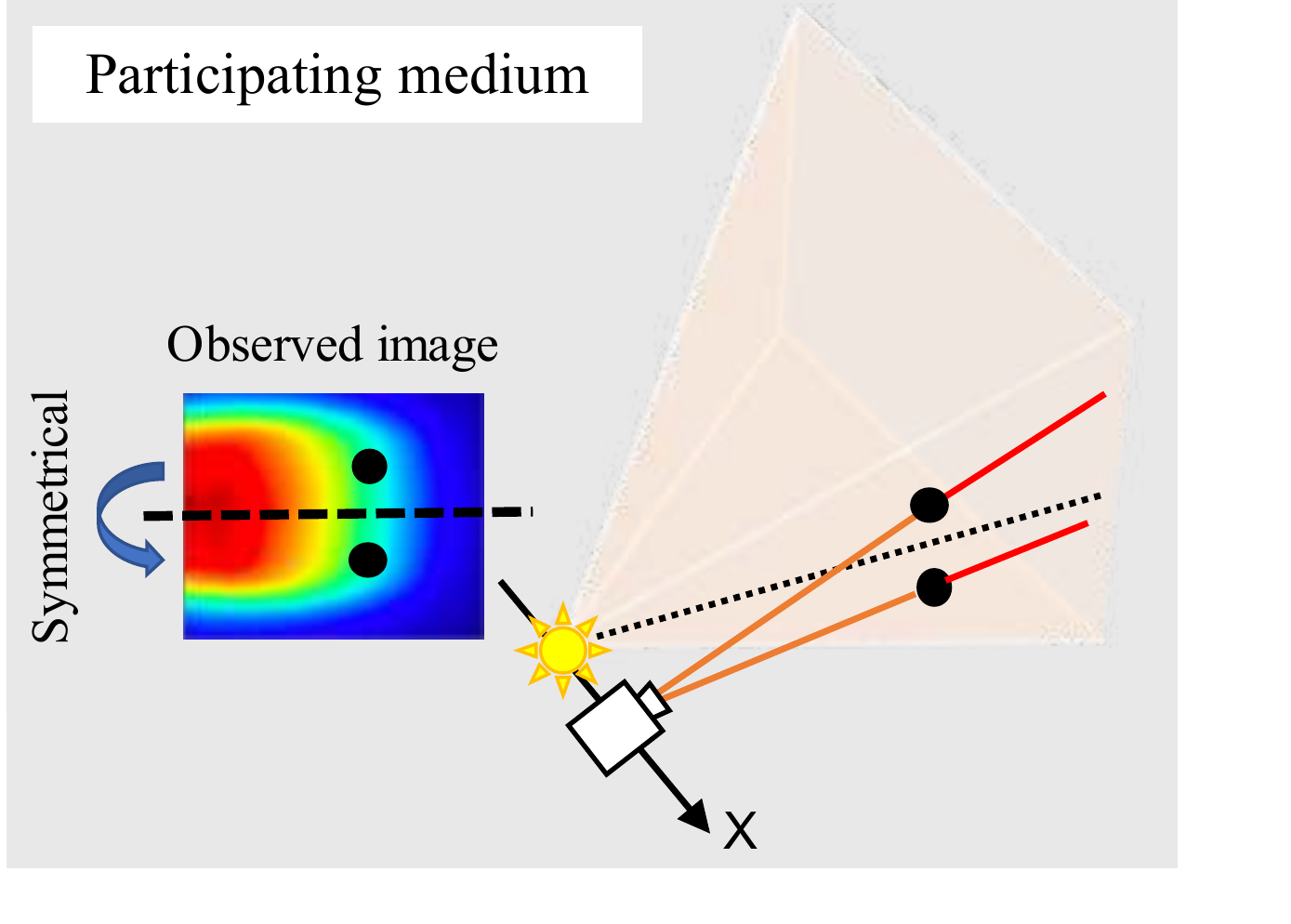}
    \caption{Global symmetrical prior. When the camera and light source are collocated on a line that is parallel to the horizontal axis of the image, the observed scattering component has symmetry because the integral domain of a pixel is consistent with that of the symmetrical pixel with respect to the central axis of the image.}
    \label{fig:symmetrical}
\end{figure}

As discussed in section \ref{sec:image_formation}, a scattering component depends on the relative position between a camera pixel and a light source.
This is because the individual starting points of the integral in Eq. (\ref{eq:lte}) differ from each other.
Meanwhile, as shown in Fig. \ref{fig:symmetrical}, we assume that the camera and light source are collocated on the line that is parallel to the horizontal axis of the image. 
Kinect v2 has this setting, and other devices can easily be built on the basis of this setting.
In this case, the integral domain of a pixel is consistent with that of the symmetrical pixel with respect to the central axis of the image.
Thus, the observed scattering component also has symmetry, and we leverage this symmetry as a global prior.

\subsection{Formulation as robust estimation}
We formulate the scattering component estimation problem as robust estimation.
Specifically, we solve the following optimization problem:
\begin{flalign}\label{eq:objective}
\min_{\mathbf{x}, \mathbf{a}_1,\cdots,\mathbf{a}_K} &\sum_{i=1}^{N} \rho \left( \frac{x_i-\tilde{x}_i}{\sigma_1} \right)& \nonumber\\
&+ \gamma_1 \sum_{k=1}^{K} \| \mathbf{U}\mathbf{a}_k - \mathbf{x}_k\|^2& \nonumber\\
&+ \gamma_2 \| \mathbf{F}\mathbf{x} - \mathbf{x} \|^2& \nonumber\\
&+ \gamma_3 \| \nabla \mathbf{x} \|^2.&
\end{flalign}
The first term of Eq. (\ref{eq:objective}) is a data term where $\tilde{\mathbf{x}}=[\tilde{x}_i\; \cdots\; \tilde{x}_N]^\top$ and $\mathbf{x} = [x_i\;\cdots \; x_N]^\top$ are a captured image and a scattering component, respectively.
$N$ is the number of camera pixels, and $\sigma_1$ is a scale parameter.
We use a nonlinear differentiable function $\rho(x)$ rather than square error $x^2$, which allows us to make the estimation robust against outliers.
In this study, we simply use the residual of the observation and the scattering component as the data term, i.e., pixels that contain a direct component are regarded as outliers.

We use three additional regularization terms.
The second term represents the local prior.
$K$ is the number of patches for local quadratic function fitting.
$\mathbf{U}$ is an $N_k \times 6$ matrix where $N_k$ is the number of pixels in patch $k$ and each row of $\mathbf{U}$ is a vector $\mathbf{u}$ that corresponds to each pixel coordinate.
In this study, these patches do not overlap each other.
The third term represents the global prior where $\mathbf{F} \in \mathbb{R}^{N\times N}$ is a matrix that flips an image vertically.
The last term is a smoothing term where $\nabla$ denotes a gradient operator. 
This smoothing accelerates the optimization.
Hyperparameters $\gamma_1,\gamma_2,\gamma_3$ control the contribution of each term.

\subsection{IRLS and object region estimation}
We minimize Eq. (\ref{eq:objective}) with respect to a scattering component $\mathbf{x}$ and the coefficients of quadratic functions $\mathbf{a}_1,\cdots,\mathbf{a}_K$.
However, the nonlinearity of $\rho (x)$ makes it difficult to obtain a closed-form solution.
For efficient computation, the IRLS optimization was developed in the literature \citep{holland77, fox02}.
IRLS minimizes weighted least squares iteratively and the weight is updated using the current estimate in each iteration.
The objective function in Eq. (\ref{eq:objective}) is transformed into weighted least squares as follows:
\begin{flalign}\label{eq:irls}
\min_{\mathbf{x}, \mathbf{a}_1,\cdots,\mathbf{a}_K} & (\mathbf{x} - \tilde{\mathbf{x}})^\top \mathbf{W} (\mathbf{x} - \tilde{\mathbf{x}})& \nonumber\\
&+ \gamma'_1 \sum_{k=1}^{K} \| \mathbf{U}\mathbf{a}_k - \mathbf{x}_k\|^2& \nonumber\\
&+ \gamma'_2 \| \mathbf{F}\mathbf{x} - \mathbf{x} \|^2& \nonumber\\
&+ \gamma'_3 \| \nabla \mathbf{x} \|^2,&
\end{flalign}
where $\mathbf{W} = {\rm diag}(\mathbf{w})$ is an $N\times N$ matrix and $\mathbf{w} = [w_1,\cdots,w_N]^\top$ is the weight for each error $x_i - \tilde{x}_i$.
Hyperparameters are given as $\gamma'_* = 2\sigma_1^2 \gamma_*$.
Equation (\ref{eq:irls}) is quadratic with respect to the scattering component $\mathbf{x}$, and thus is easy to optimize.
In each iteration, we solve Eq. (\ref{eq:irls}) for $\mathbf{x}$ and $\mathbf{a}_1,\cdots,\mathbf{a}_K$, and the weight can be updated using the current estimate as
\begin{equation}
\label{eq:weight_pixel_wise}
w_i = \frac{\rho'\left((x_i - \tilde{x}_i)/\sigma_1\right)}{(x_i - \tilde{x}_i)/\sigma_1}.
\end{equation}

\begin{algorithm}[tb]                      
\caption{Simultaneous estimation of scattering component and object region}         
\label{alg1}               
\begin{algorithmic}[0]
\REQUIRE Image $\tilde{\mathbf{x}}$
\ENSURE Scattering component $\mathbf{x}$ and object mask $\mathbf{w}$
\STATE Coarse level optimization (Eq. (\ref{eq:patch_regression})):
\bindent
\STATE $\mathbf{W} \leftarrow \mathbf{I}$, $\mathbf{a}_k \leftarrow \argmin_{\mathbf{a}_k}\| \mathbf{U}\mathbf{a}_k - \tilde{\mathbf{x}}_k\|^2$
\REPEAT
\STATE Solve Eq. (\ref{eq:irls}) for $\mathbf{x}$
\STATE Solve Eq. (\ref{eq:irls}) for $\mathbf{a}_1,\cdots,\mathbf{a}_K$
\IF{first iteration}
\STATE Compute $\sigma_2$ using Eq. (\ref{eq:sigma2})
\ENDIF
\STATE Update $\mathbf{w}$ using Eq. (\ref{eq:weight_patch_wise})
\UNTIL{converged}
\eindent
\STATE Fine level optimization (Eq. (\ref{eq:objective})):
\bindent
\STATE Initialize $\mathbf{w}$ and $\mathbf{a}_1,\cdots,\mathbf{a}_K$ with the output of the coarse level
\REPEAT
\STATE Solve Eq. (\ref{eq:irls}) for $\mathbf{x}$
\STATE Solve Eq. (\ref{eq:irls}) for $\mathbf{a}_1,\cdots,\mathbf{a}_K$
\IF{first iteration}
\STATE Compute $\sigma_1$ using Eq. (\ref{eq:sigma1})
\ENDIF
\STATE Update $\mathbf{w}$ using Eq. (\ref{eq:weight_pixel_wise})
\UNTIL{converged}
\eindent
\STATE Binarize $\mathbf{w}$
\end{algorithmic}
\end{algorithm}

The specific update rule of the weight depends on the nonlinear function $\rho(x)$.
In this study, we use the following function as $\rho(x)$:
\begin{equation}\label{eq:tukey_loss}
\rho(x) = \left\{ 
\begin{array}{ll}
\frac{c^2}{6} \left[ 1 - \left\{  1 - \left( \frac{x}{c}  \right)^2 \right\}^3  \right] &   if \;\;|x| \leq c \\
\frac{c^2}{6} & otherwise.
\end{array}
\right. 
\end{equation}
This function yields the following update:
\begin{equation}\label{eq:tukey_weight}
w_i = \left\{ 
\begin{array}{ll}
\left\{ 1 - \left( \frac{r_i}{c} \right)^2 \right\}^2 & if\;\;|r_i| \leq c \\
0 & otherwise,
\end{array}
\right.
\end{equation}
where $r_i = (x_i - \tilde{x}_i)/\sigma_1$, and $c$ is a tuning parameter.
This update is referred to as Tukey's biweight \citep{beaton74, fox02}, where $0 \leq w_i \leq 1$.

The weight controls the robust estimation, that is, a large error term reduces the corresponding weight.
In this study, we consider the object region as outliers, and thus the weight in the object region should be small.
Therefore, we can leverage the IRLS weight to extract the object region from the image.

\subsection{Coarse-to-fine optimization}
The accurate object region extraction is critical for the effectiveness of the scattering component estimation.
In section \ref{sec:prior_of_scattering_component}, we introduced the local and global priors of the scattering component to deal with a large object region.
To make the region extraction more robust, we developped a coarse-to-fine optimization scheme.
Before solving Eq. (\ref{eq:objective}), we optimize the following objective function:
\begin{flalign}\label{eq:patch_regression}
\min_{\mathbf{x}, \mathbf{a}_1,\cdots,\mathbf{a}_K} &\sum_{k=1}^{K} \rho \left( \frac{\| \mathbf{x}_k - \tilde{\mathbf{x}}_k \|}{\sigma_2} \right)& \nonumber\\
&+ \gamma_1 \sum_{k=1}^{K} \| \mathbf{U}\mathbf{a}_k - \mathbf{x}_k\|^2& \nonumber\\
&+ \gamma_2 \| \mathbf{F}\mathbf{x} - \mathbf{x} \|^2& \nonumber\\
&+ \gamma_3 \| \nabla \mathbf{x} \|^2.&
\end{flalign}
This is similar to the patch-based robust regression proposed by \cite{chaudhury13}.
The difference from Eq. (\ref{eq:objective}) is that the data term consists of patch-wise errors.
Equation (\ref{eq:patch_regression}) can be transformed into IRLS as well as Eq. (\ref{eq:objective}), and the IRLS weight is updated patch-wise rather than pixel-wise.
Differentiating the first term of Eq. (\ref{eq:patch_regression}) with respect to $\mathbf{x}$, we can obtain
\begin{flalign}
\frac{\partial}{\partial \mathbf{x}}\sum_{k=1}^{K}\rho\left( \frac{\| \mathbf{x}_k - \tilde{\mathbf{x}}_k\|}{\sigma_2} \right) &= \sum_{k=1}^{K}\mathcal{F}_k\left( \frac{\partial}{\partial \mathbf{x}_k}\rho\left( \frac{\| \mathbf{x}_k - \tilde{\mathbf{x}}_k\|}{\sigma_2} \right) \right)& \nonumber \\
&= \sum_{k=1}^{K}\mathcal{F}_k\left( \frac{1}{\sigma_2^2} w_k (\mathbf{x}_k - \tilde{\mathbf{x}}_k) \right)&  \nonumber \\
&= \frac{1}{\sigma_2^2} \mathbf{W}(\mathbf{x}-\tilde{\mathbf{x}}),&
\end{flalign}
where an operator $\mathcal{F}_k: \mathbb{R}^{N_k}\to \mathbb{R}^N$ embeds an input patch into an overall image with zero padding and returns its vectorized form.
The weight matrix is $\mathbf{W} = {\rm diags}(\mathbf{w}) = {\rm diags}([ w_1 \mathbbm{1}^\top_1 \cdots w_K \mathbbm{1}^\top_K ]^\top)$ where $\mathbbm{1}_k \in \mathbb{R}^{N_k}$.
The weight $w_k$ for each patch $k$ is given as
\begin{equation}\label{eq:weight_patch_wise}
w_k = \frac{\rho'(\| \mathbf{x}_k - \tilde{\mathbf{x}}_k\| / \sigma_2)}{\| \mathbf{x}_k - \tilde{\mathbf{x}}_k\| / \sigma_2}.
\end{equation}
Therefore, we can obtain the same objective function as Eq. ($\ref{eq:irls}$) where $\gamma'_* = 2\sigma_2^2 \gamma_2$.

\begin{figure*}[tb]
  \centering
    \includegraphics[scale=0.8]{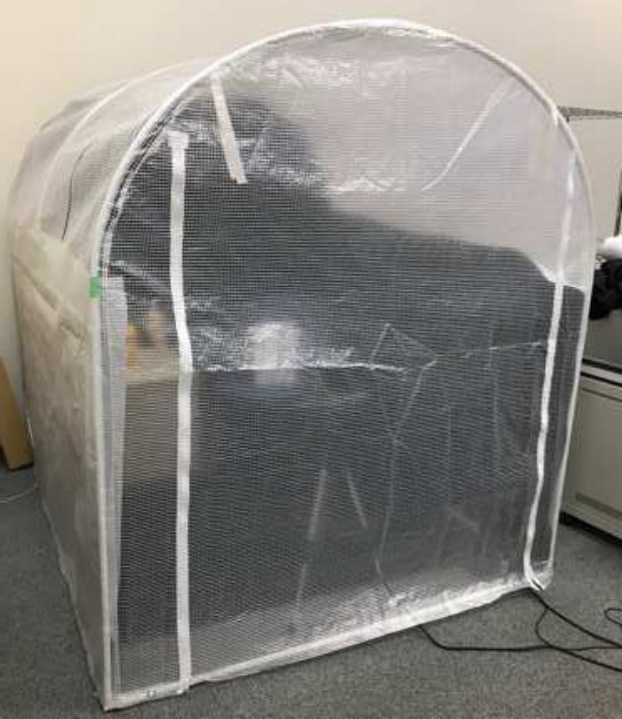}\quad
    \includegraphics[scale=0.8]{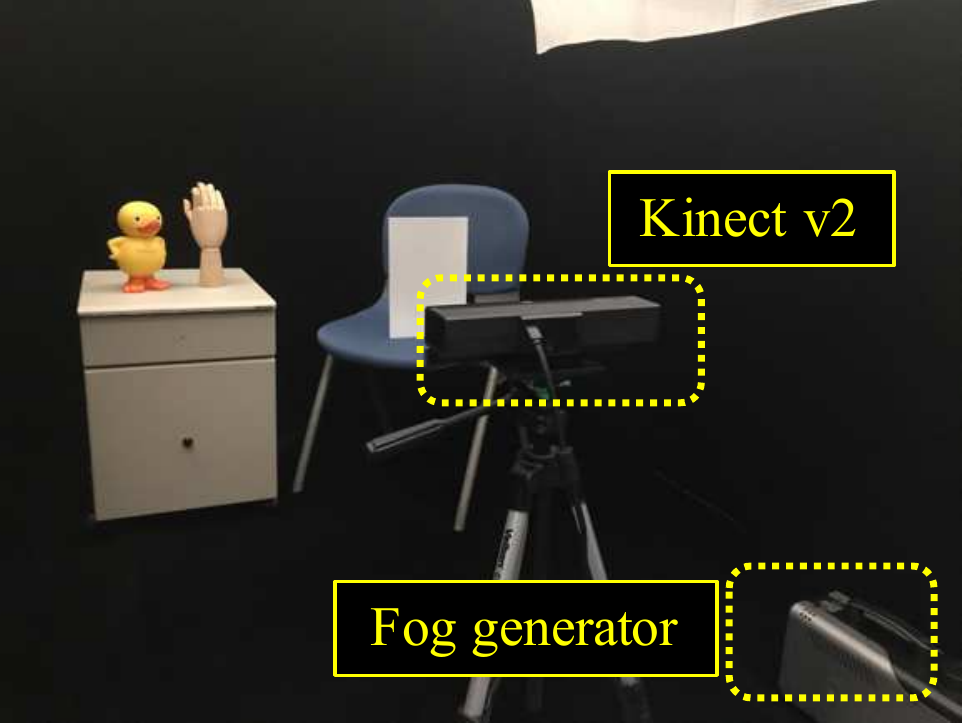}
    \caption{Experimental environment.}
    \label{fig:environment}
\end{figure*}

\begin{figure*}[tb]
\centering
  \subfloat[Scene]{
    \includegraphics[scale=\figurescaleR]{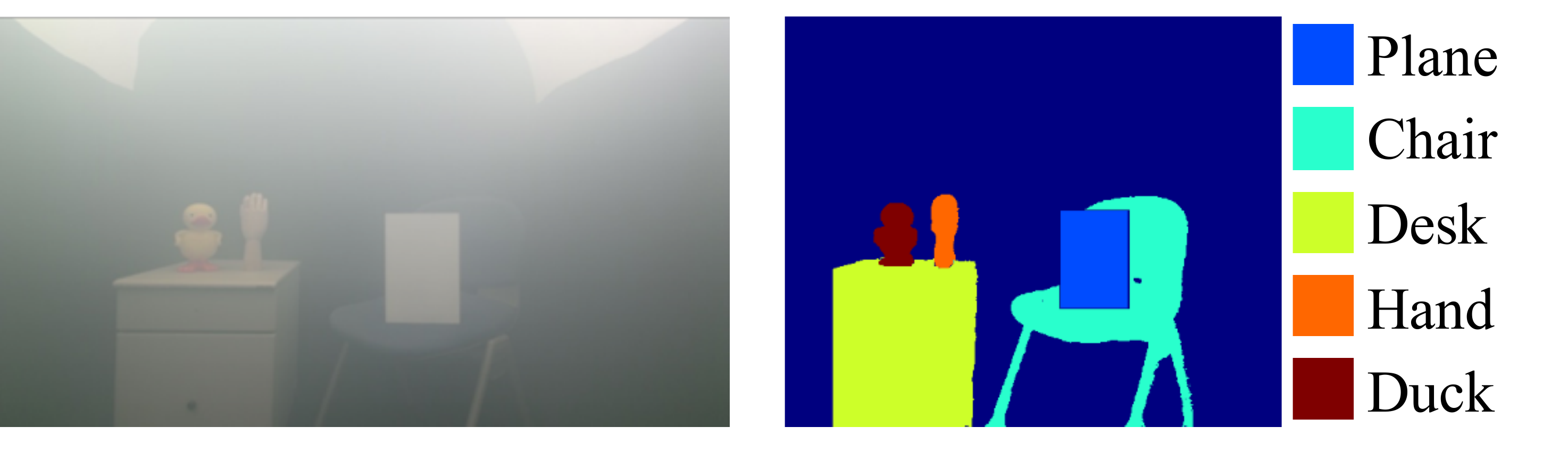}}\\
  \subfloat[Amplitude]{
    \includegraphics[scale=\figurescaleR]{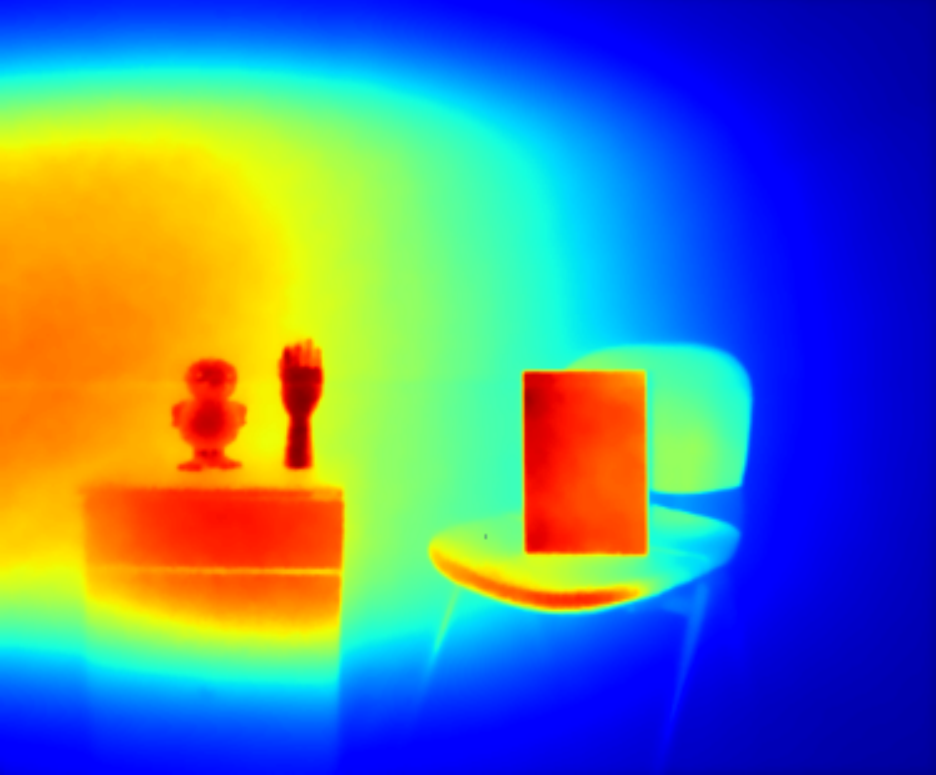}
    \includegraphics[scale=\figurescaleR]{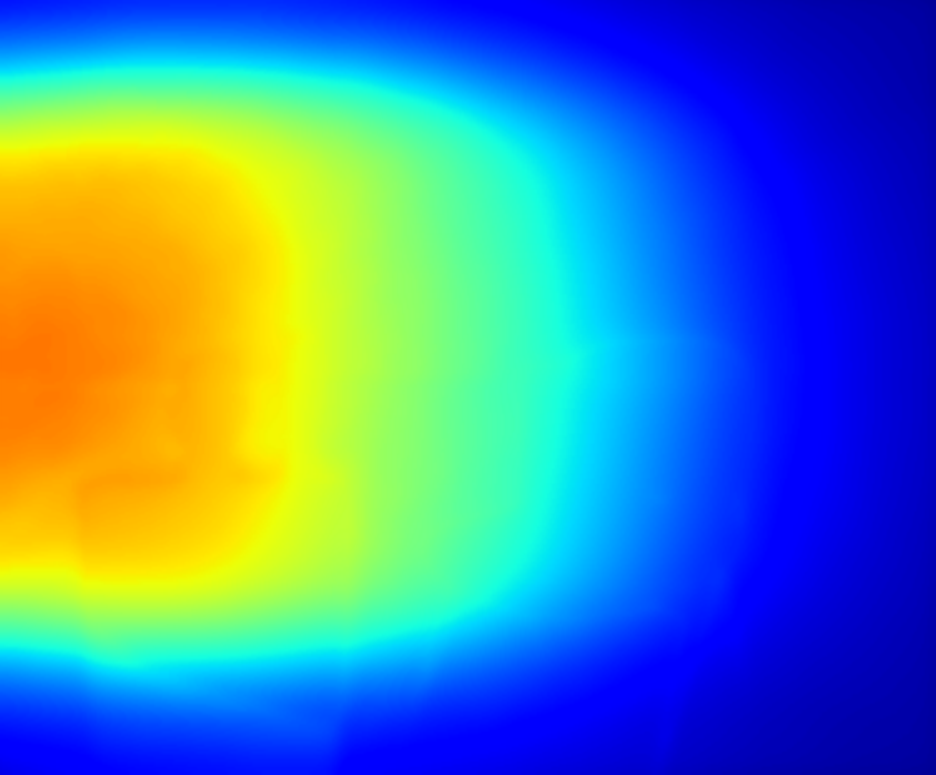}
    \includegraphics[scale=\figurescaleR]{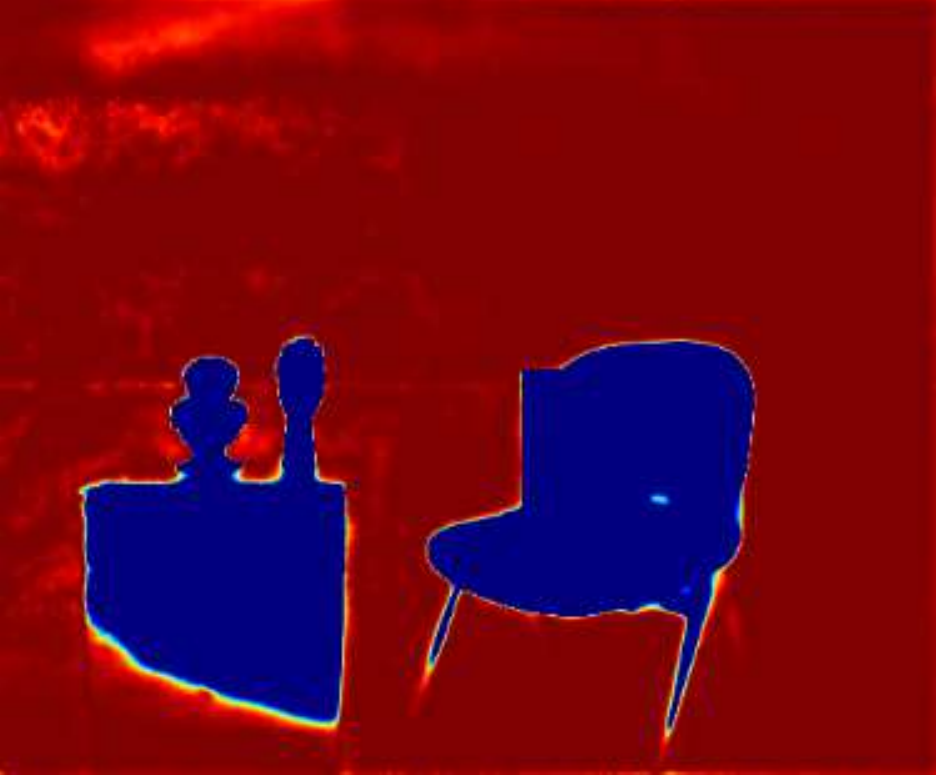}}\quad
  \subfloat[Phase]{
    \includegraphics[scale=\figurescaleR]{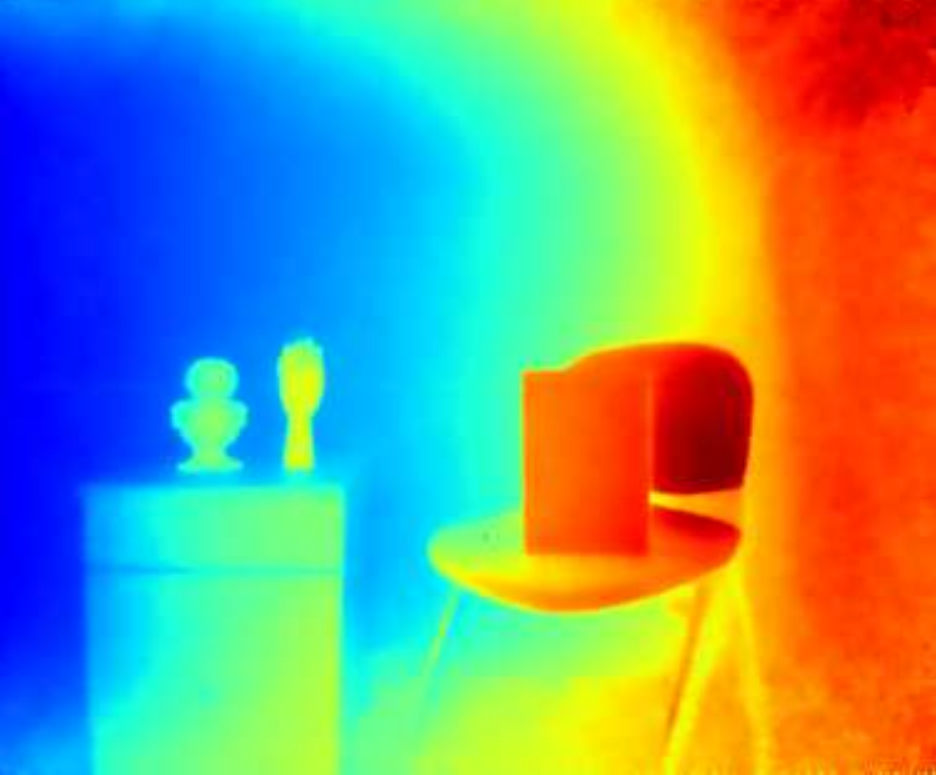}
    \includegraphics[scale=\figurescaleR]{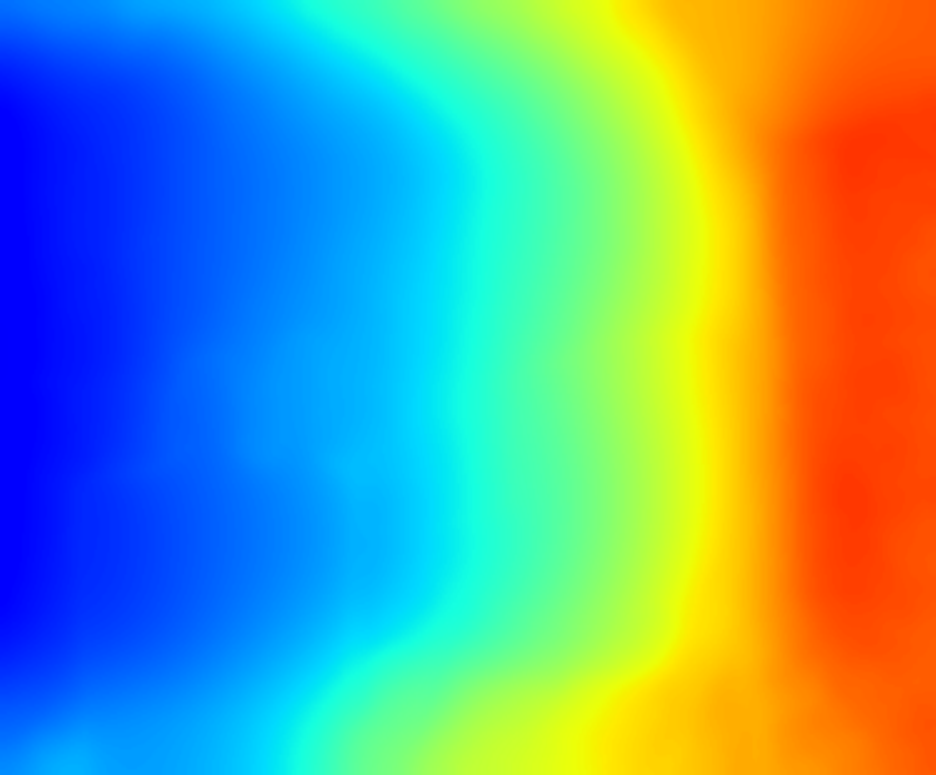}
    \includegraphics[scale=\figurescaleR]{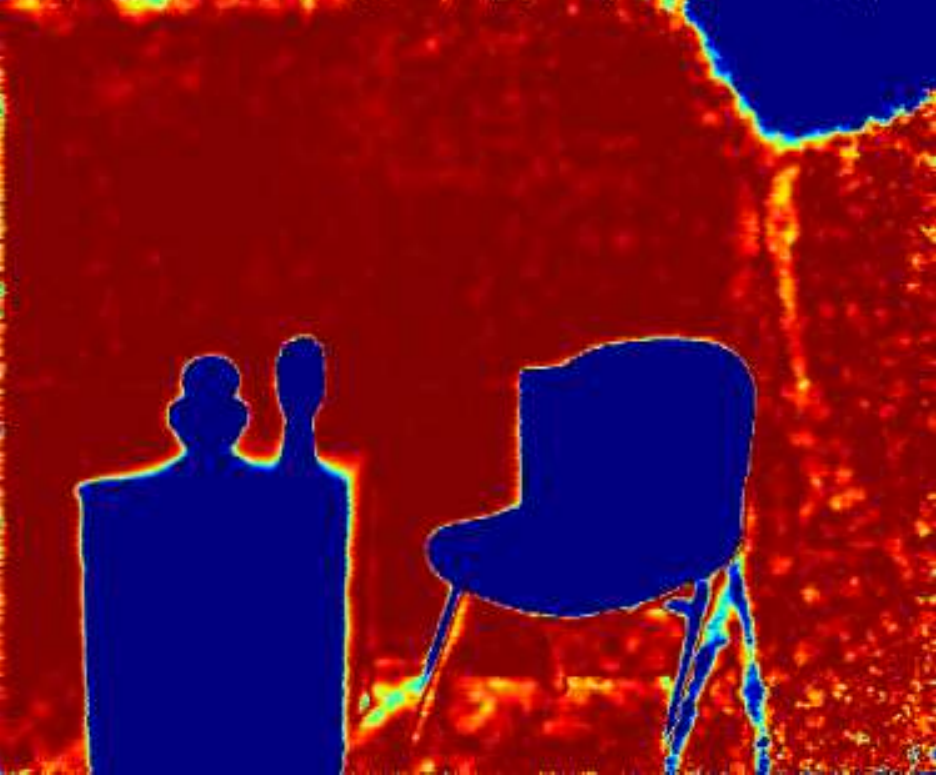}}\\
  \subfloat[Depth reconstruction]{
    \includegraphics[scale=\figurescaleR]{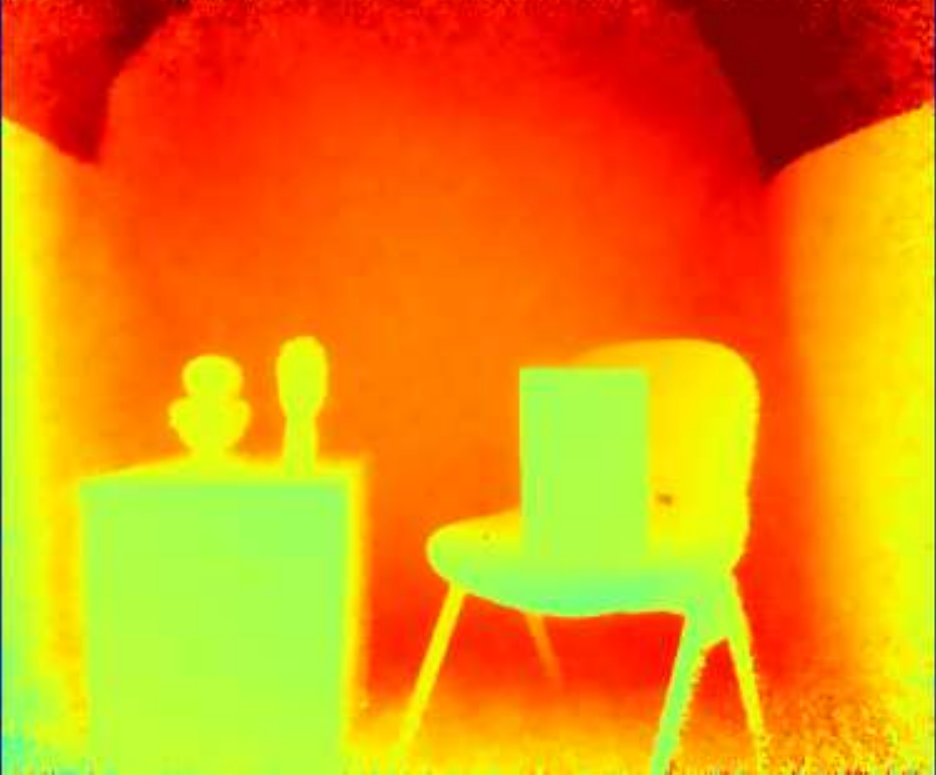}\quad 
    \includegraphics[scale=\figurescaleR]{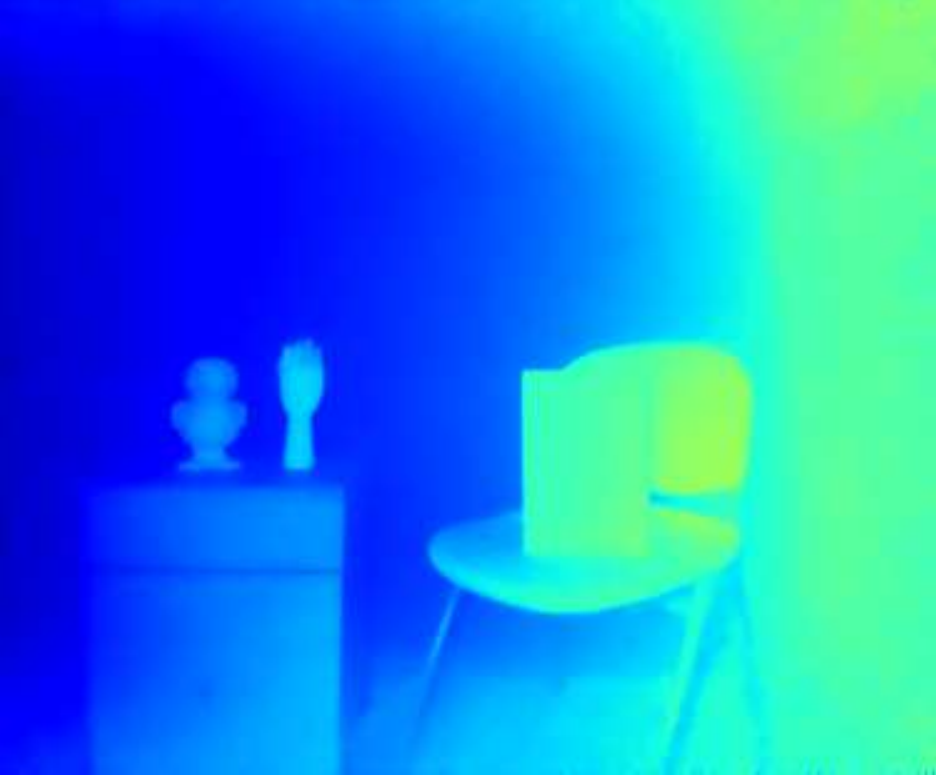}\quad
    \includegraphics[scale=\figurescaleR]{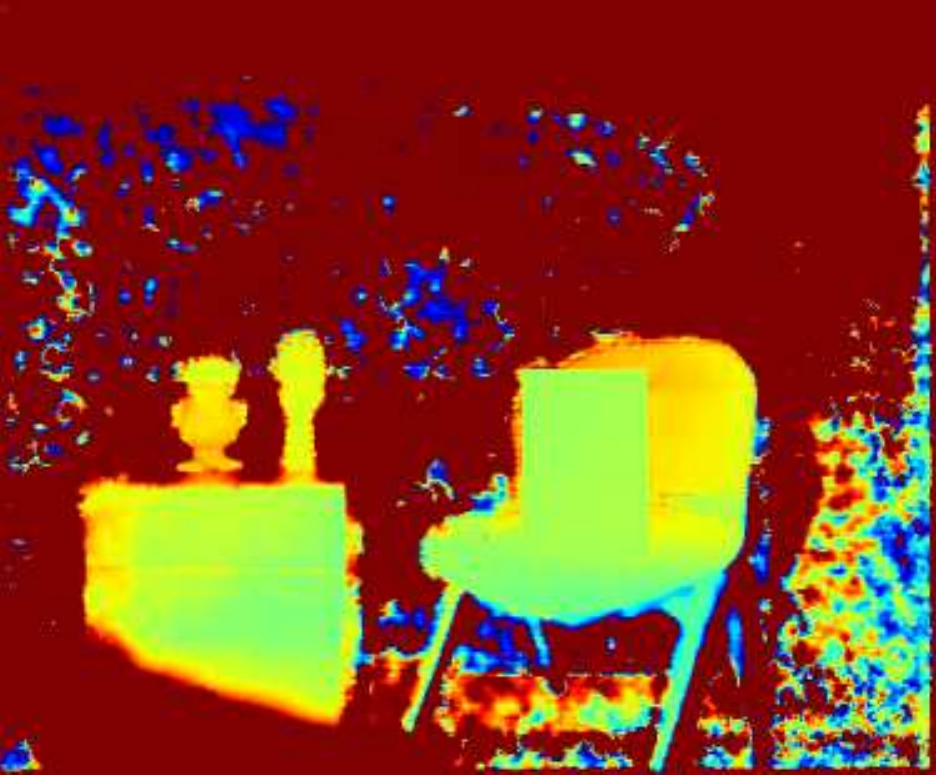}\quad
    \includegraphics[scale=\figurescaleR]{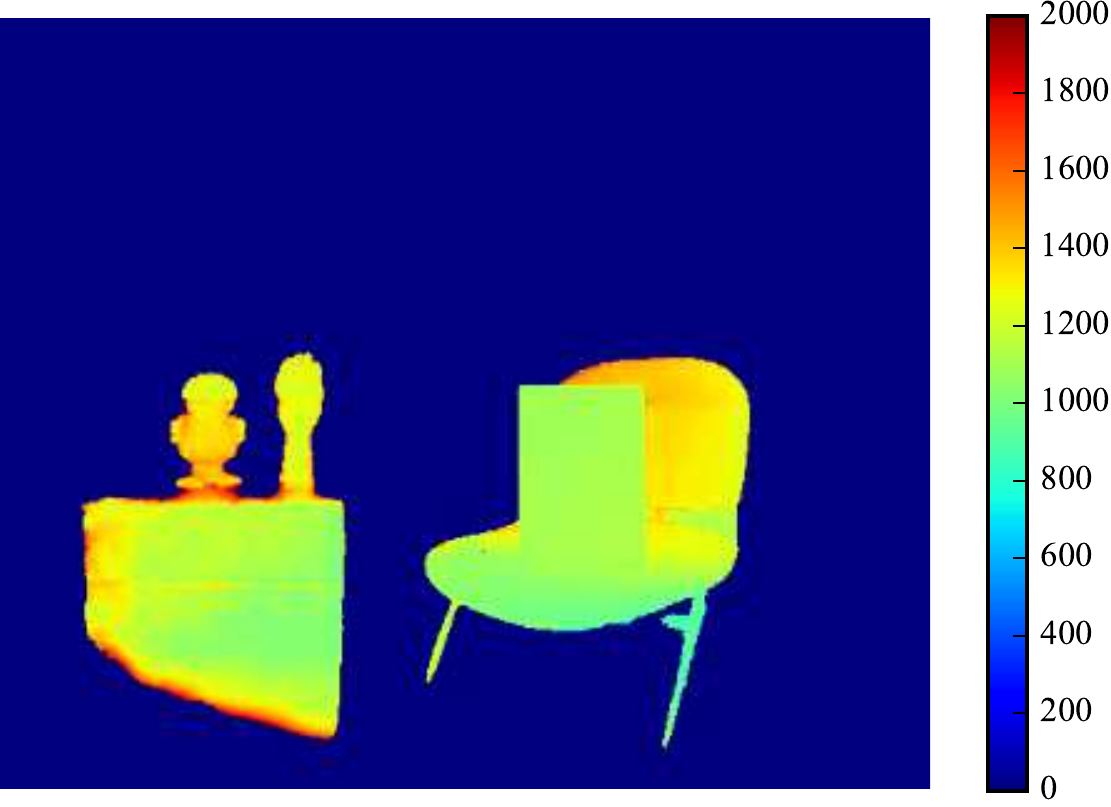} \quad
    \includegraphics[scale=\figurescaleR]{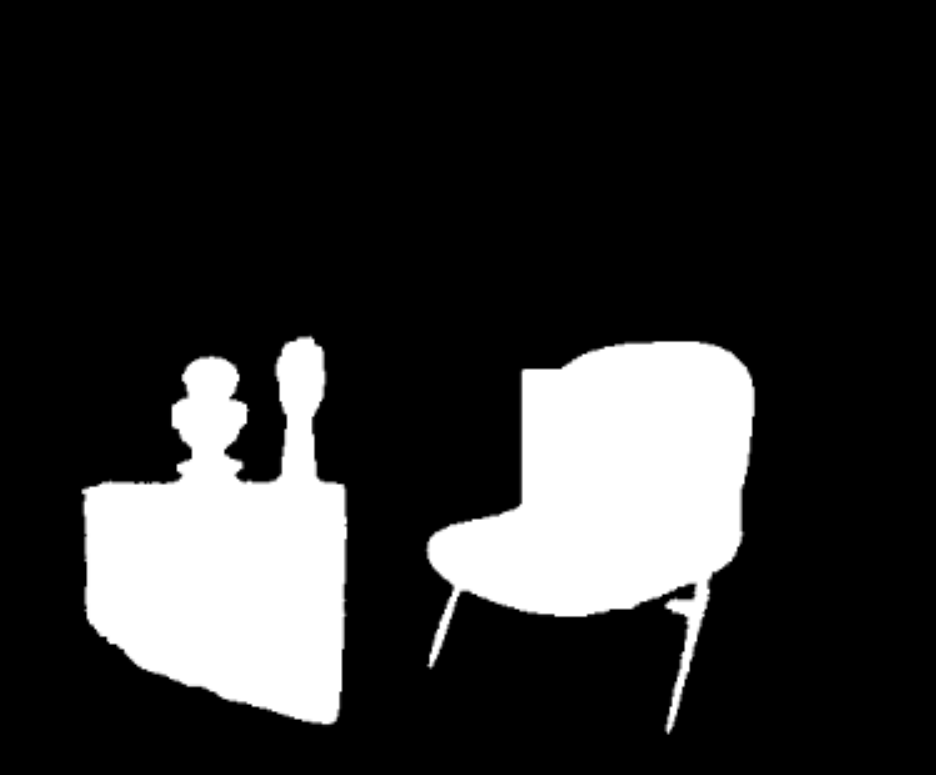}}
\caption{Results of proposed method. (a) Target foggy scene. (b)(c) Left to right: input image, estimated scattering component, and IRLS weight for the amplitude and phase image, respectively. (d) Left to right: depth without fog, depth with fog, reconstructed depth, masked depth, and estimated object mask.}
\label{fig:medium_result}
\end{figure*}

Algorithm \ref{alg1} shows the overall procedure of the simultaneous estimation of a scattering component and an object region.
We first solve Eq. (\ref{eq:patch_regression}) for the weight in a patch level and then solve (\ref{eq:objective}) in a pixel level.
Each scale parameter is computed only at the first iteration and is fixed during subsequent iterations.
We compute the scale parameters using a median absolute deviation, which is the robust measure of a deviation, as follows \citep{fox02}: 
\begin{flalign}
\sigma_1 &= \frac{{\rm median}\{|x_1 - \tilde{x}_1|,\cdots,|x_N - \tilde{x}_N|\}}{0.6745},& \label{eq:sigma1}\\
\sigma_2 &= \frac{{\rm median}\{\|\mathbf{x}_1 - \tilde{\mathbf{x}}_1\|,\cdots,\|\mathbf{x}_K - \tilde{\mathbf{x}}_K\|\} }{0.6745}. \label{eq:sigma2}&
\end{flalign}
At the end of the algorithm, we binarize the IRLS weight to generate an object mask.
This procedure is applied to an amplitude and a phase image in the same manner, and thus we can obtain the object mask in each domain.
In this study, we determine a final object mask as their intersection.

\begin{figure*}[tb]
\centering
  \subfloat[Result under thin fog]{
    \begin{tabular}{c}
      \begin{minipage}{1.0\hsize}
        \centering
              \includegraphics[scale=\figurescale]{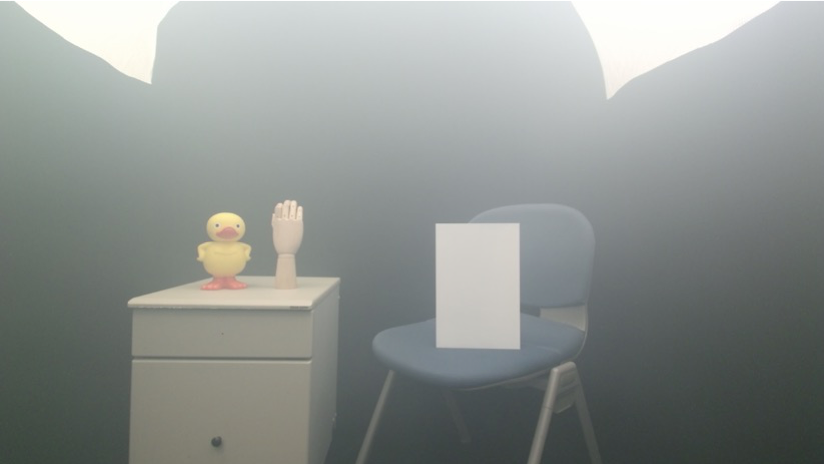}\quad
              \includegraphics[scale=\figurescale]{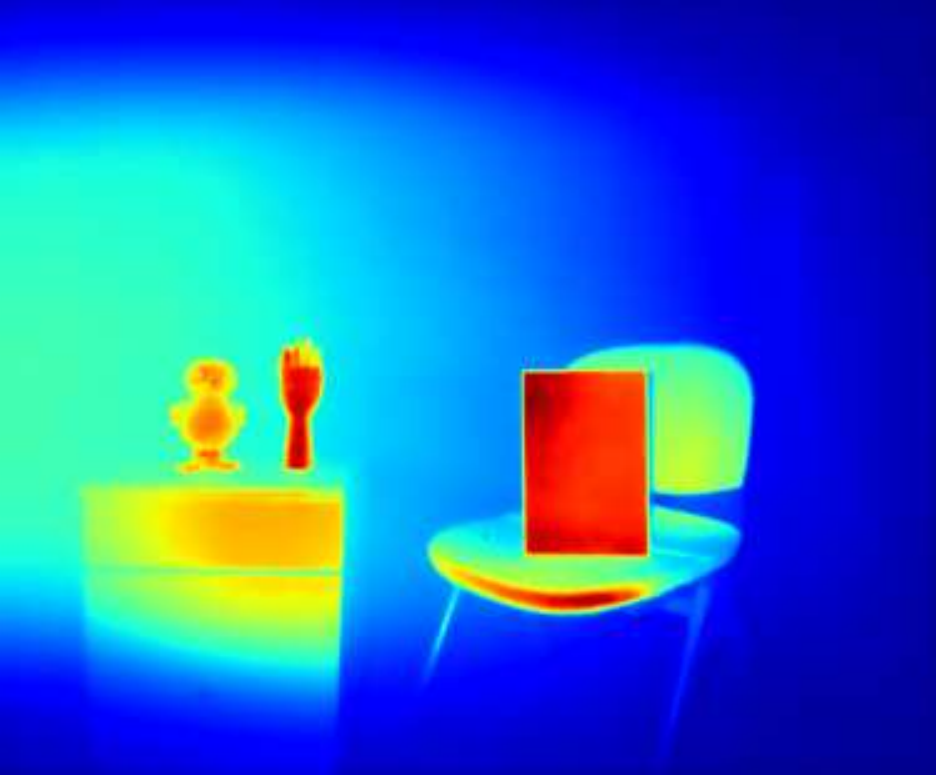}
              \includegraphics[scale=\figurescale]{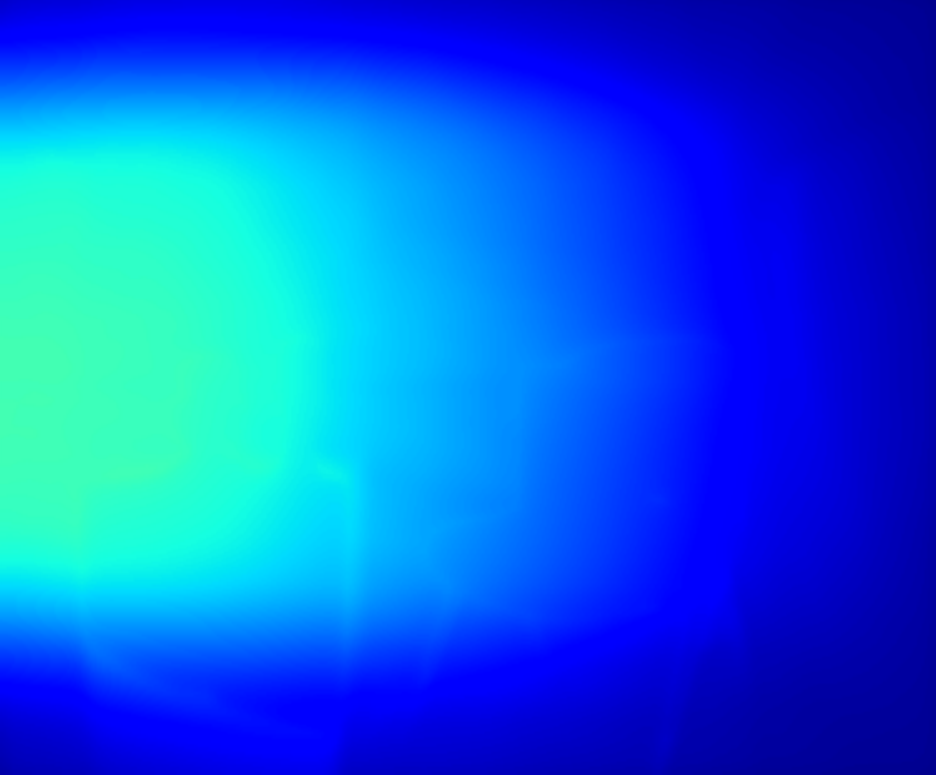}
              \includegraphics[scale=\figurescale]{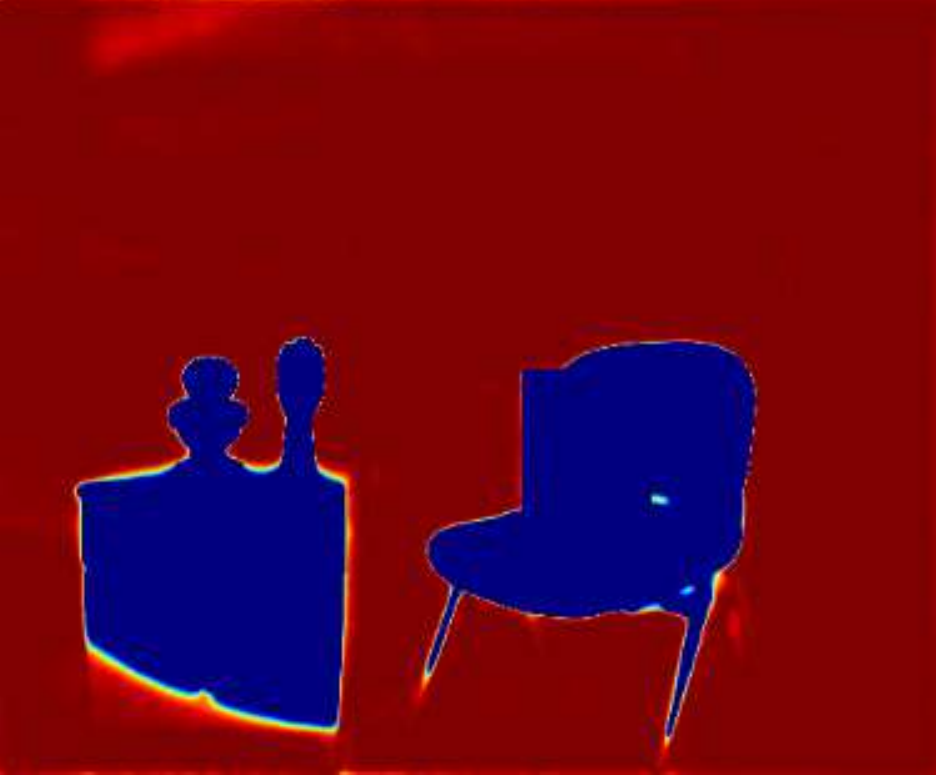}\quad
              \includegraphics[scale=\figurescale]{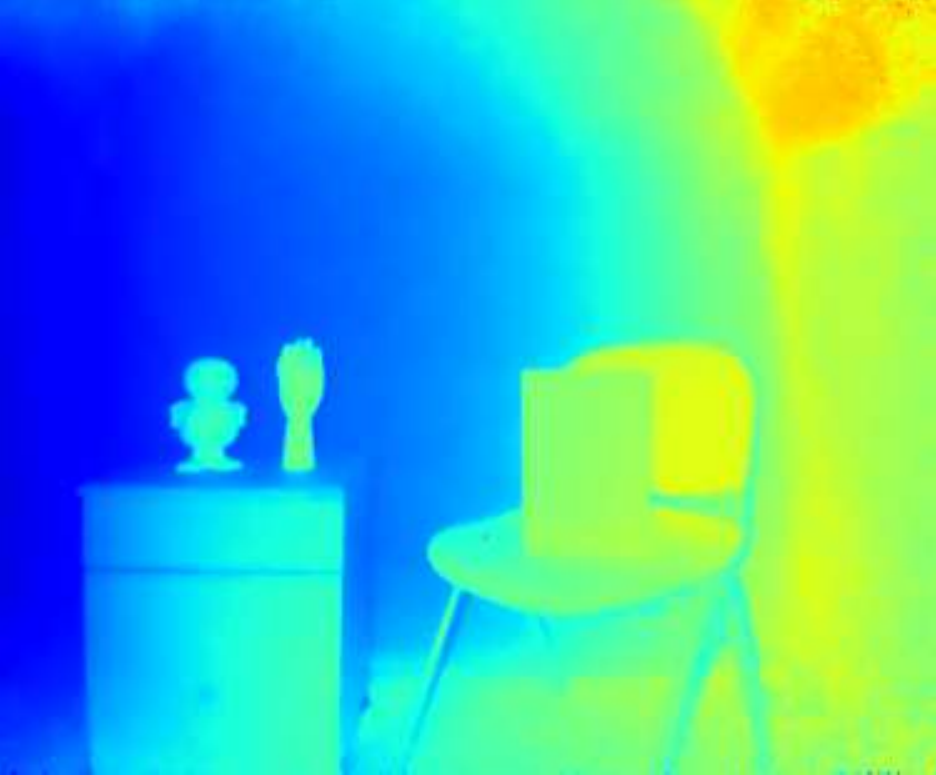}
              \includegraphics[scale=\figurescale]{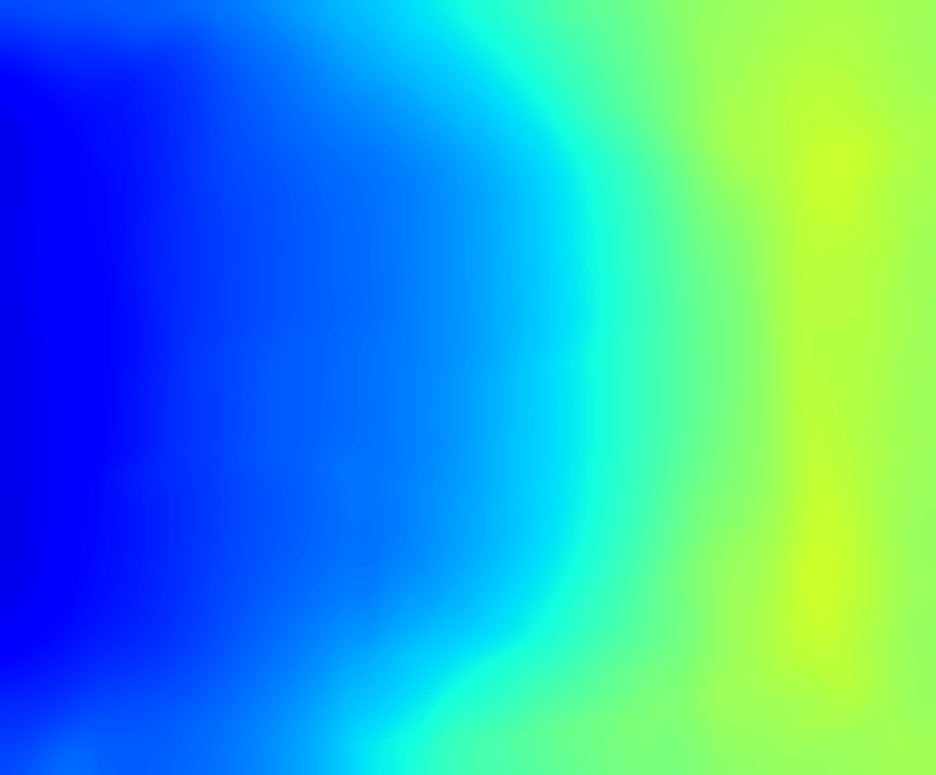}
              \includegraphics[scale=\figurescale]{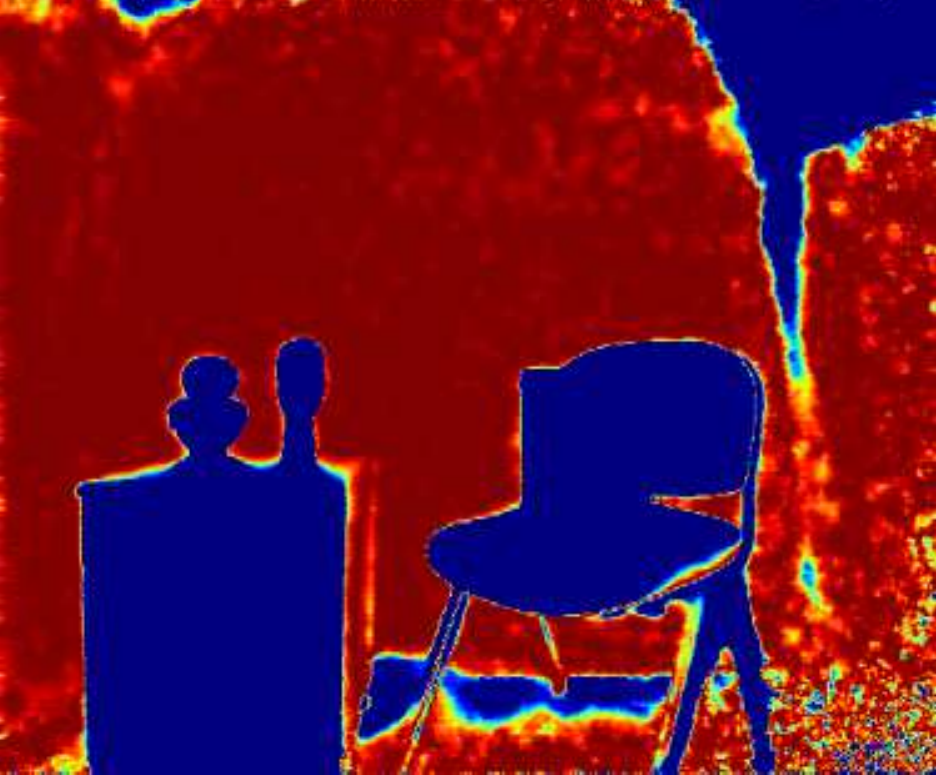}
      \end{minipage}\\ \\
      \begin{minipage}{1.0\hsize}
        \centering
              \includegraphics[scale=\figurescale]{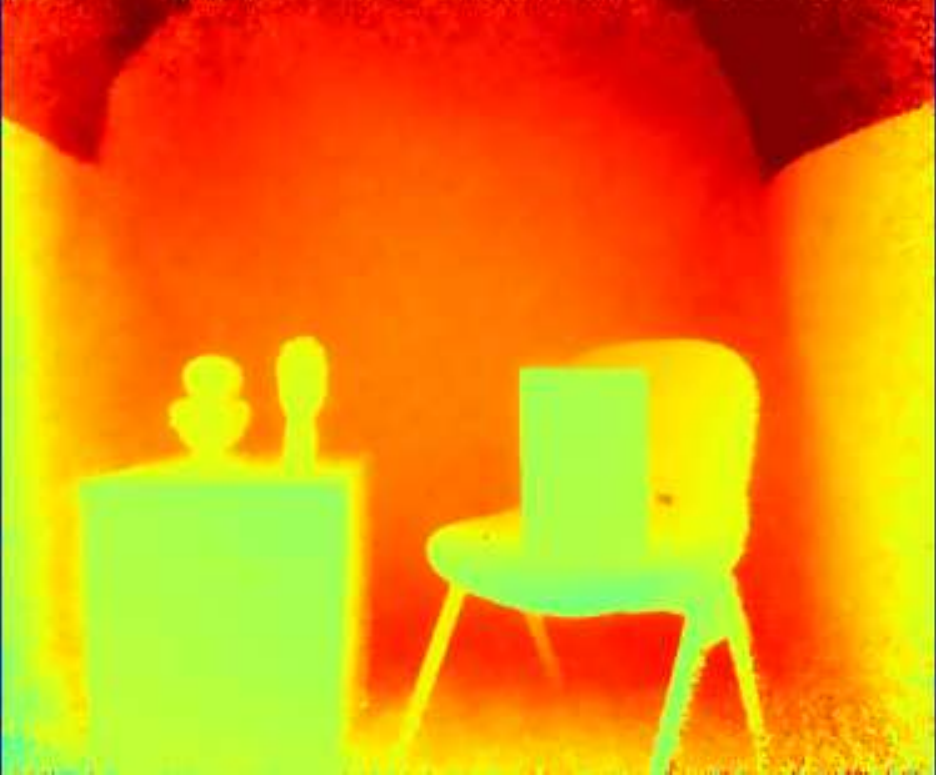}\quad 
              \includegraphics[scale=\figurescale]{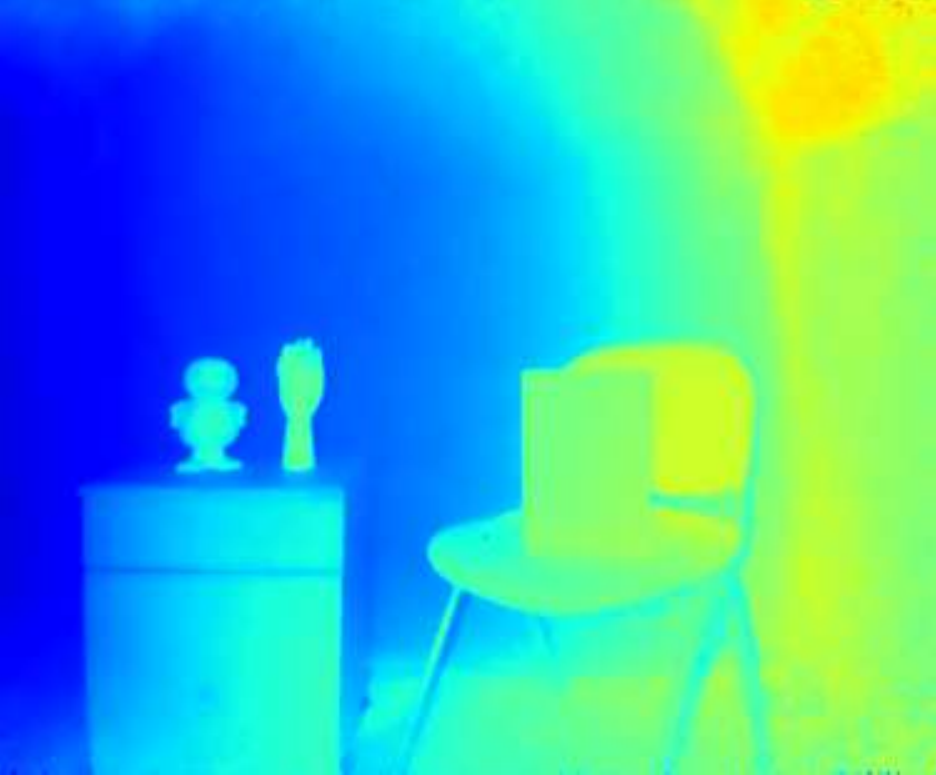}\quad
              \includegraphics[scale=\figurescale]{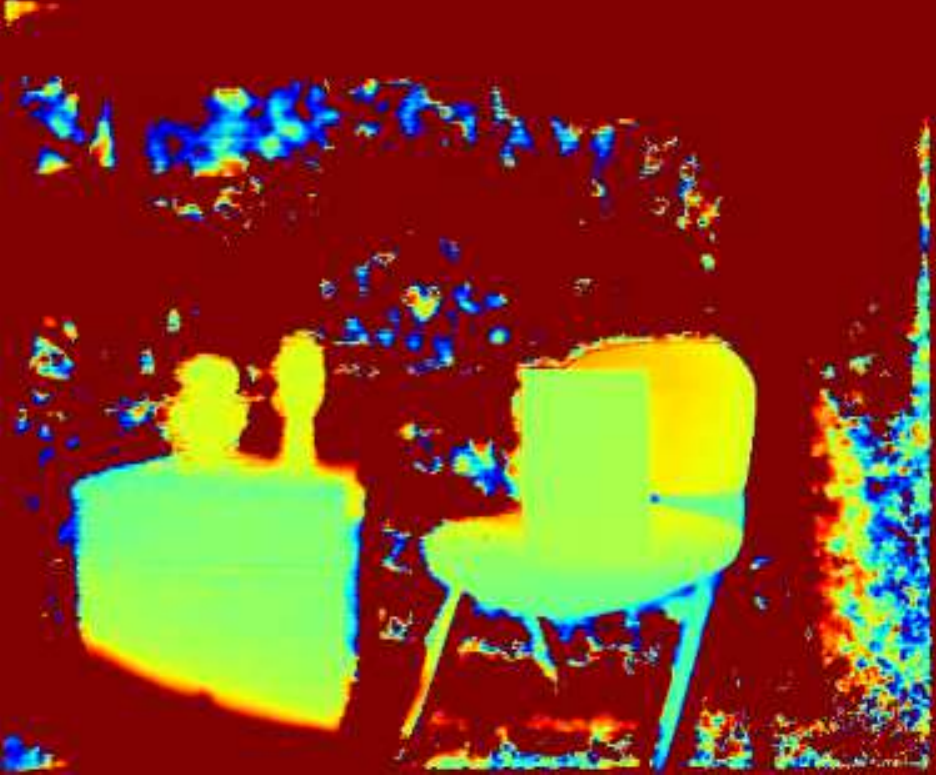}\quad
              \includegraphics[scale=\figurescale]{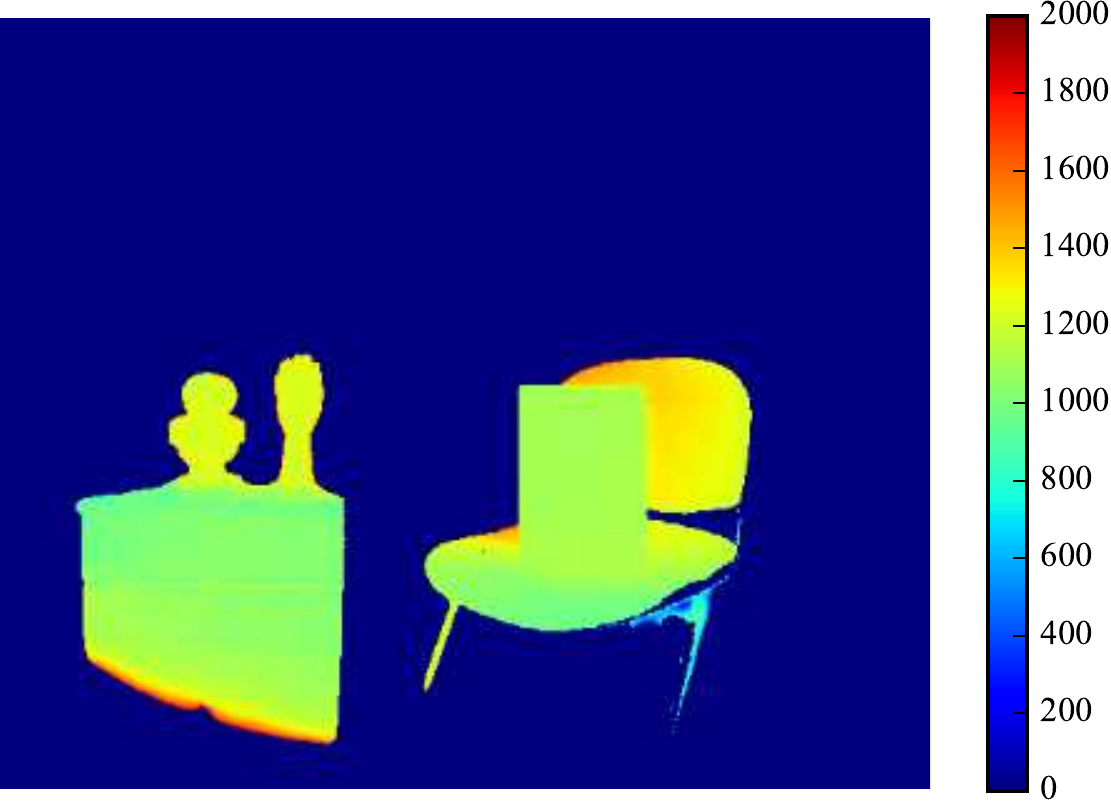}\quad
              \includegraphics[scale=\figurescale]{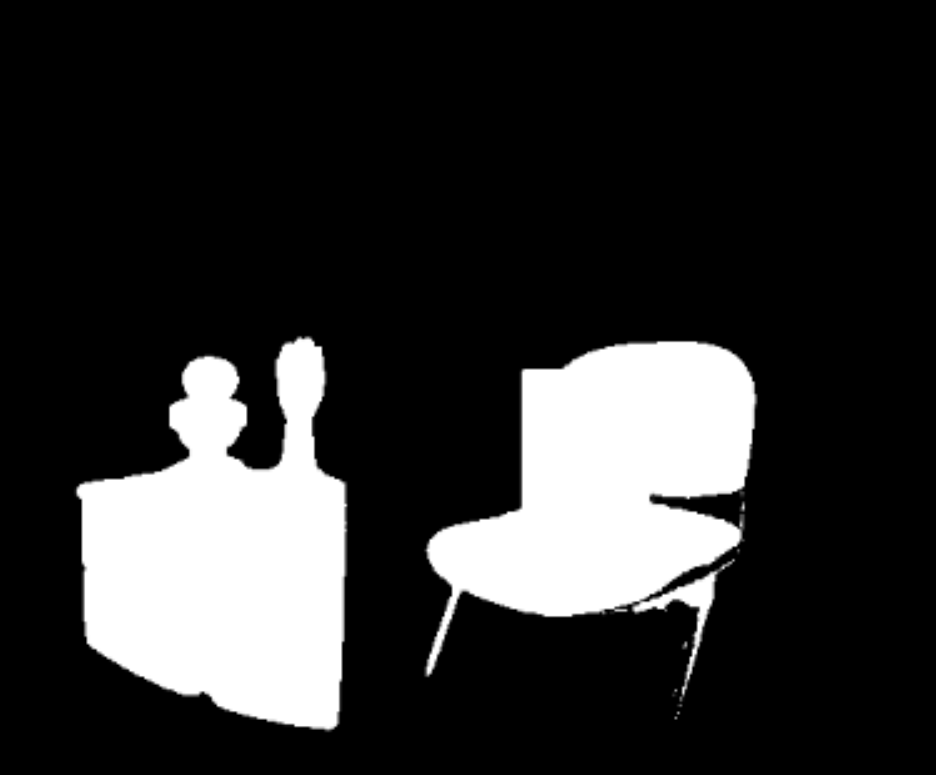}
      \end{minipage}
    \end{tabular}
    }\\
  \subfloat[Result under thick fog]{
    \begin{tabular}{c}
      \begin{minipage}{1.0\hsize}
        \centering
              \includegraphics[scale=\figurescale]{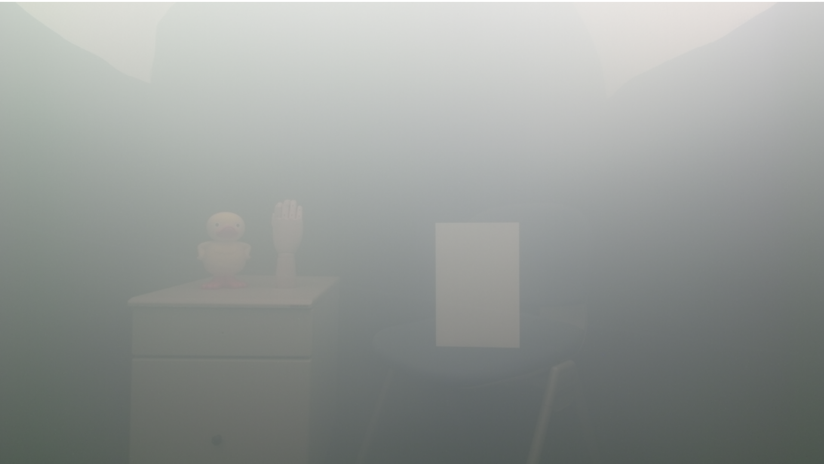}\quad
              \includegraphics[scale=\figurescale]{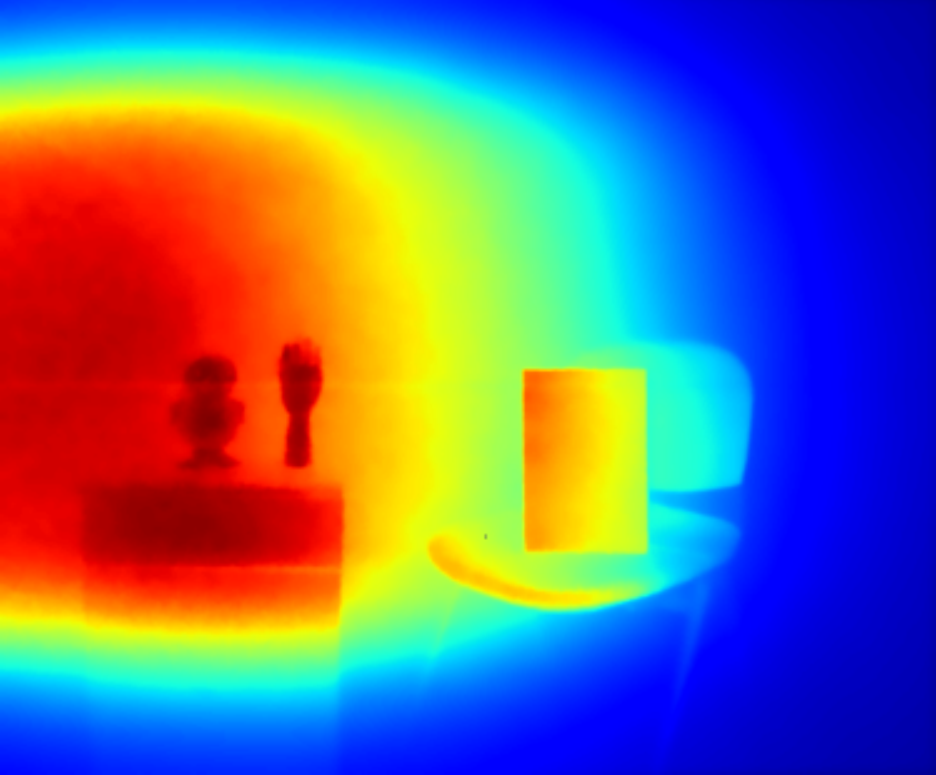}
              \includegraphics[scale=\figurescale]{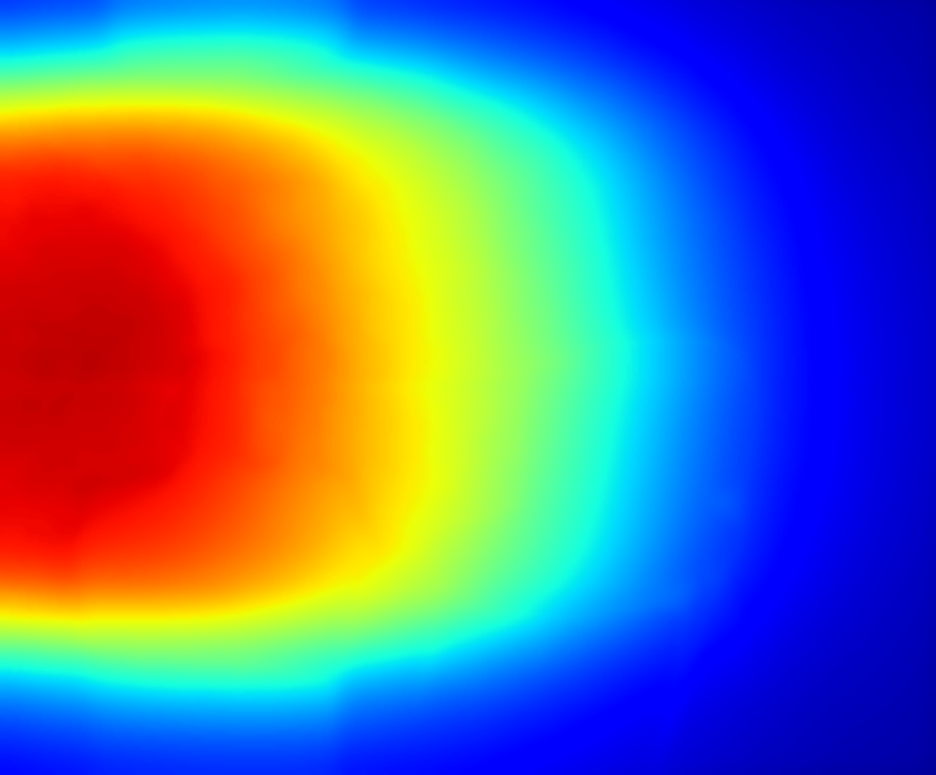}
              \includegraphics[scale=\figurescale]{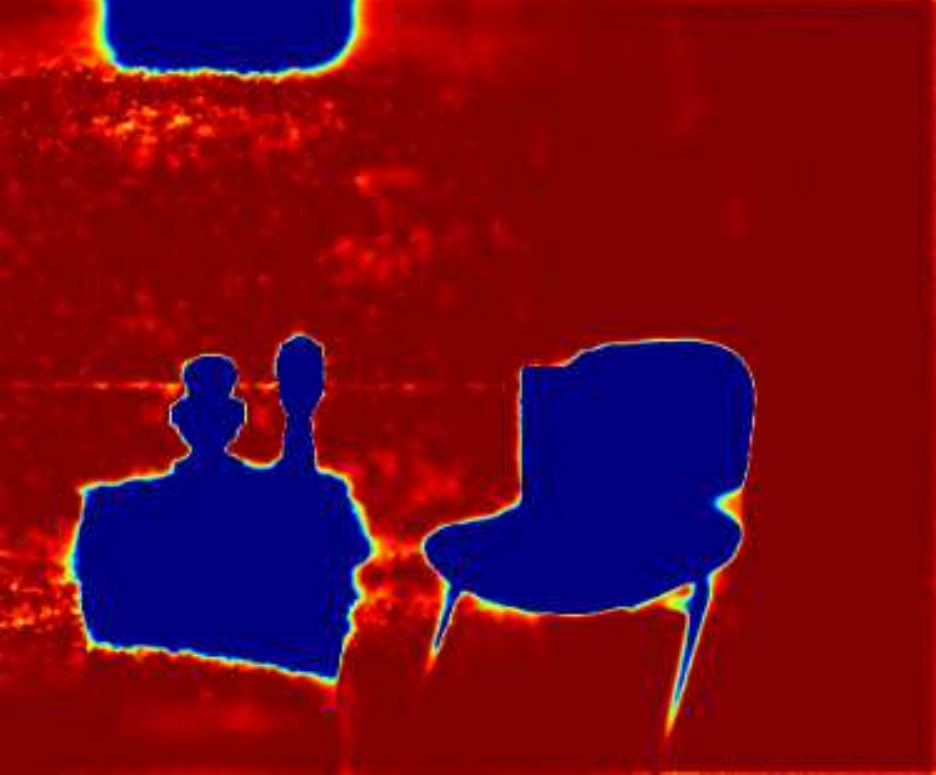}\quad
              \includegraphics[scale=\figurescale]{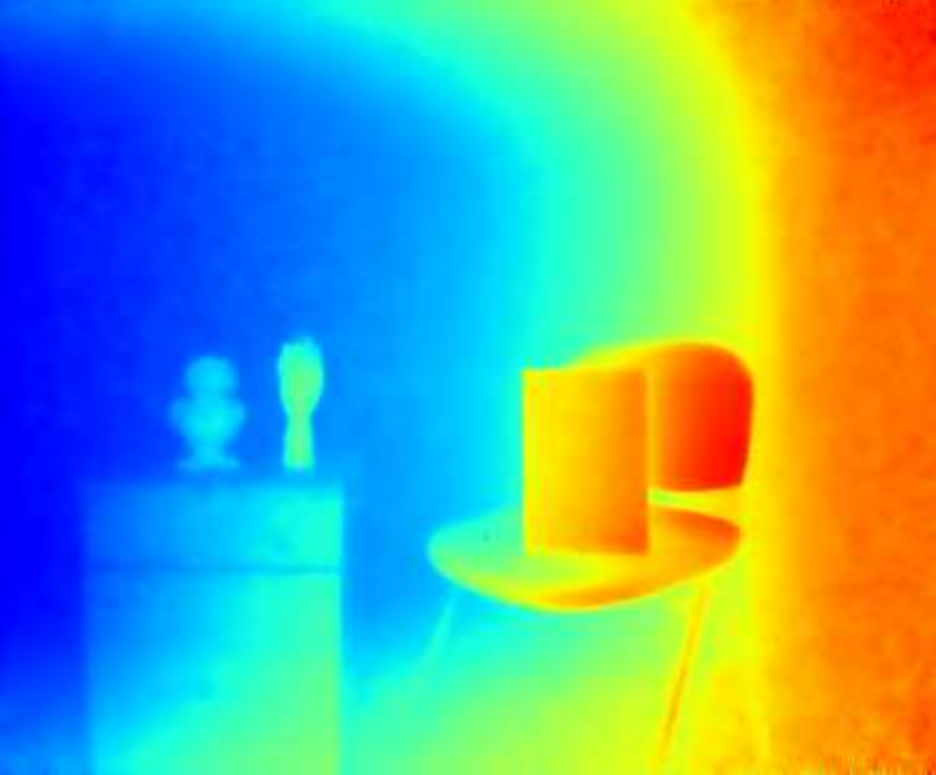}
              \includegraphics[scale=\figurescale]{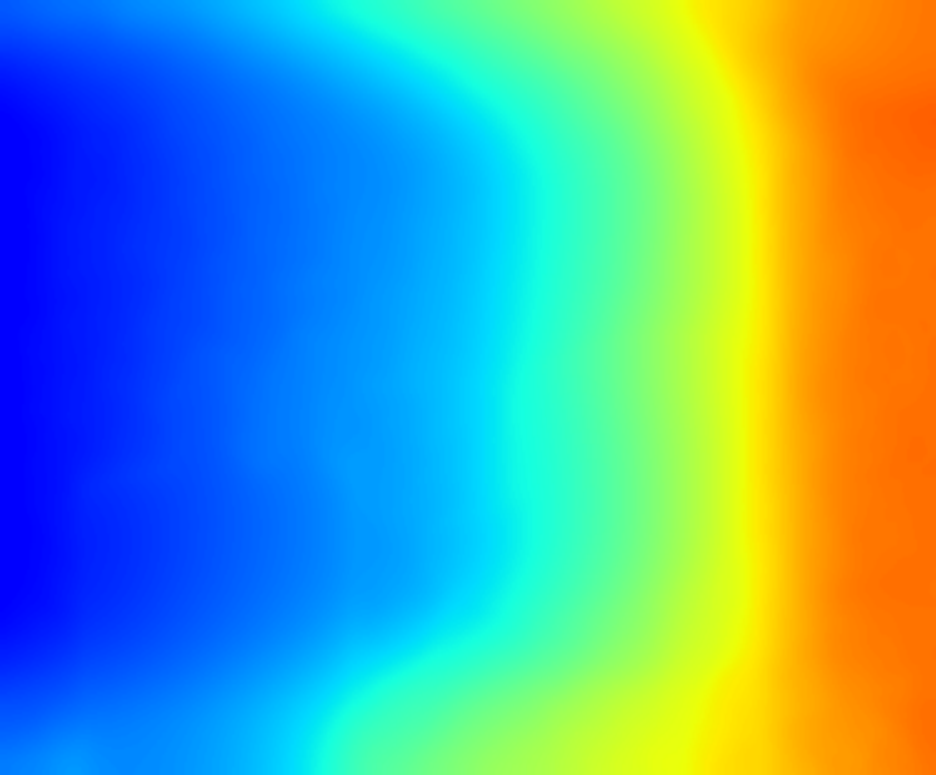}
              \includegraphics[scale=\figurescale]{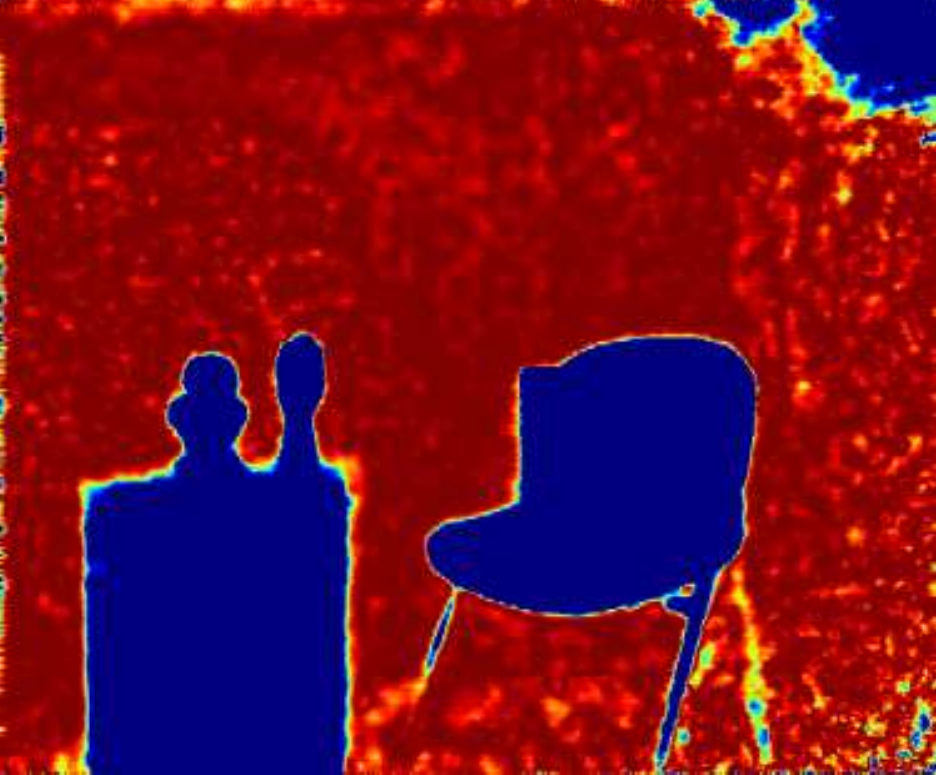}
      \end{minipage}\\ \\
      \begin{minipage}{1.0\hsize}
        \centering
              \includegraphics[scale=\figurescale]{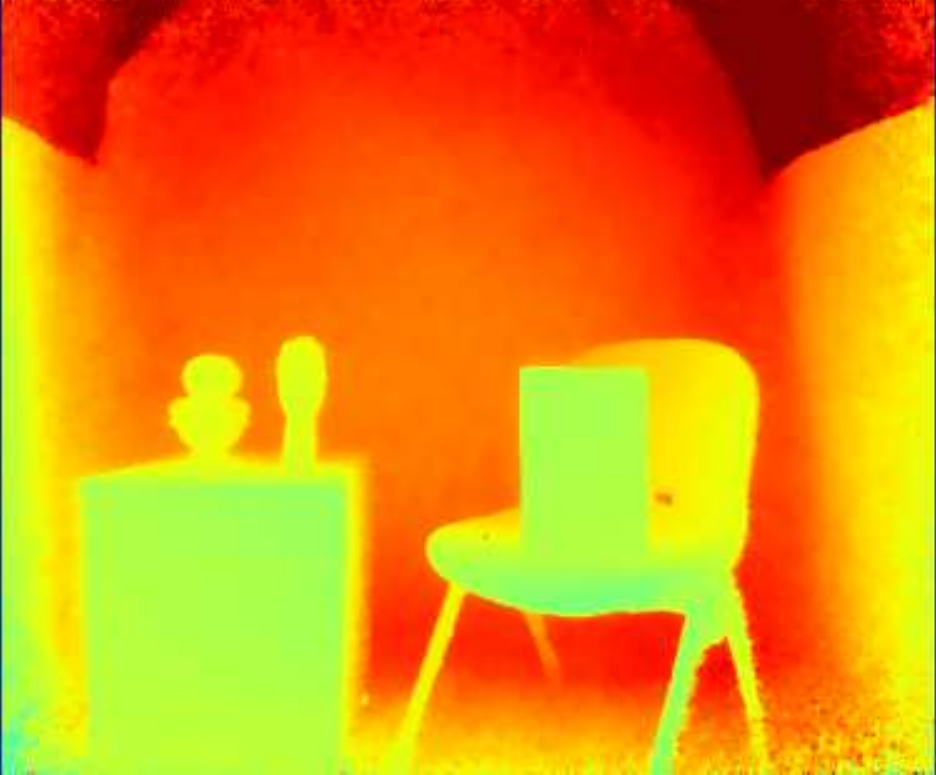}\quad 
              \includegraphics[scale=\figurescale]{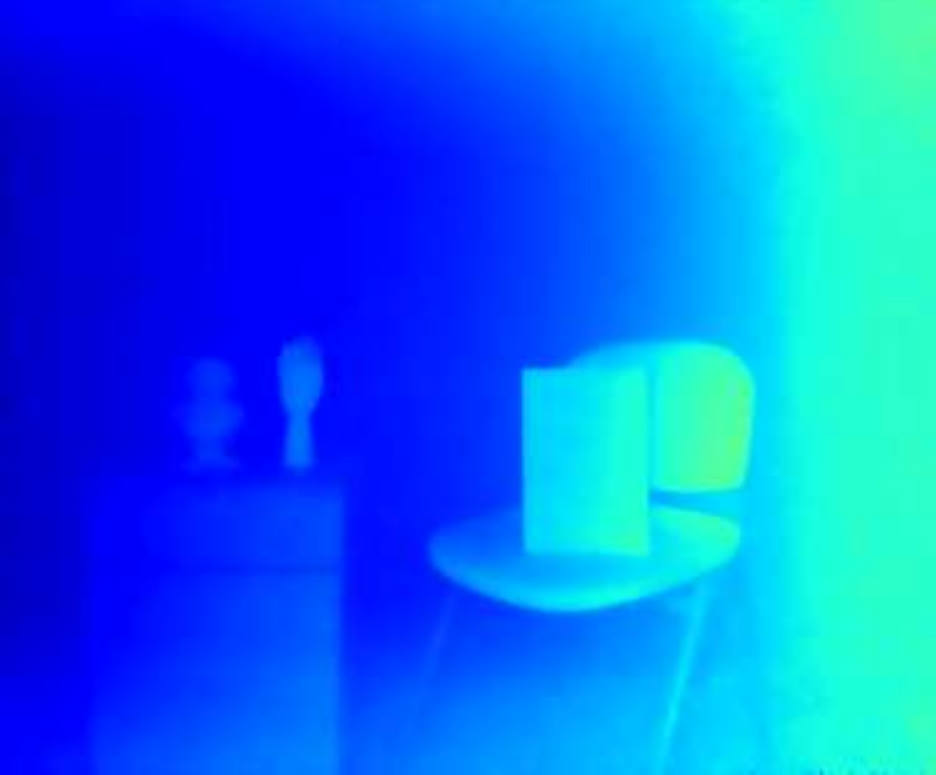}\quad
              \includegraphics[scale=\figurescale]{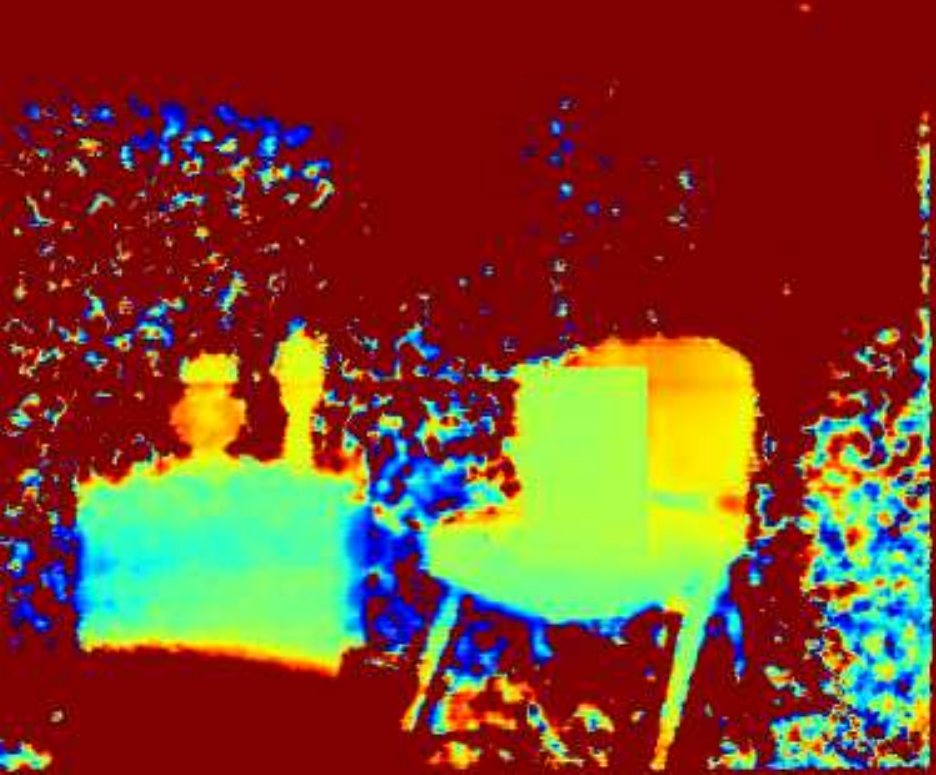}\quad
              \includegraphics[scale=\figurescale]{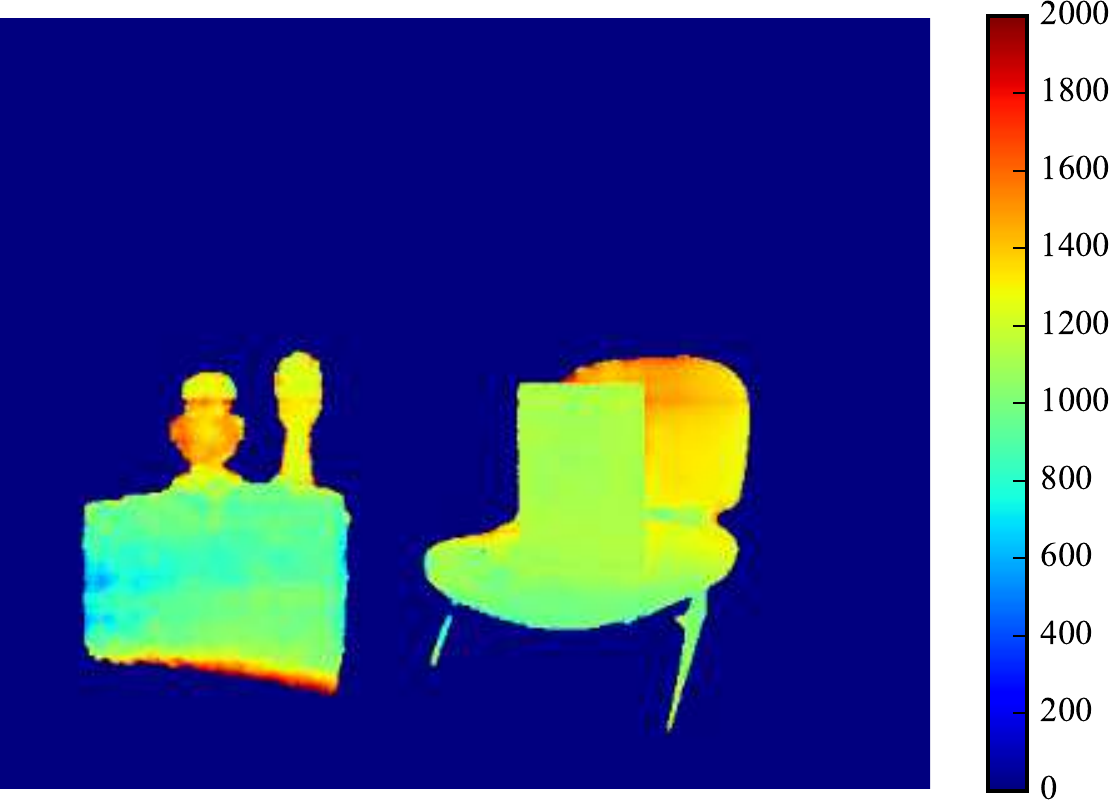}\quad
              \includegraphics[scale=\figurescale]{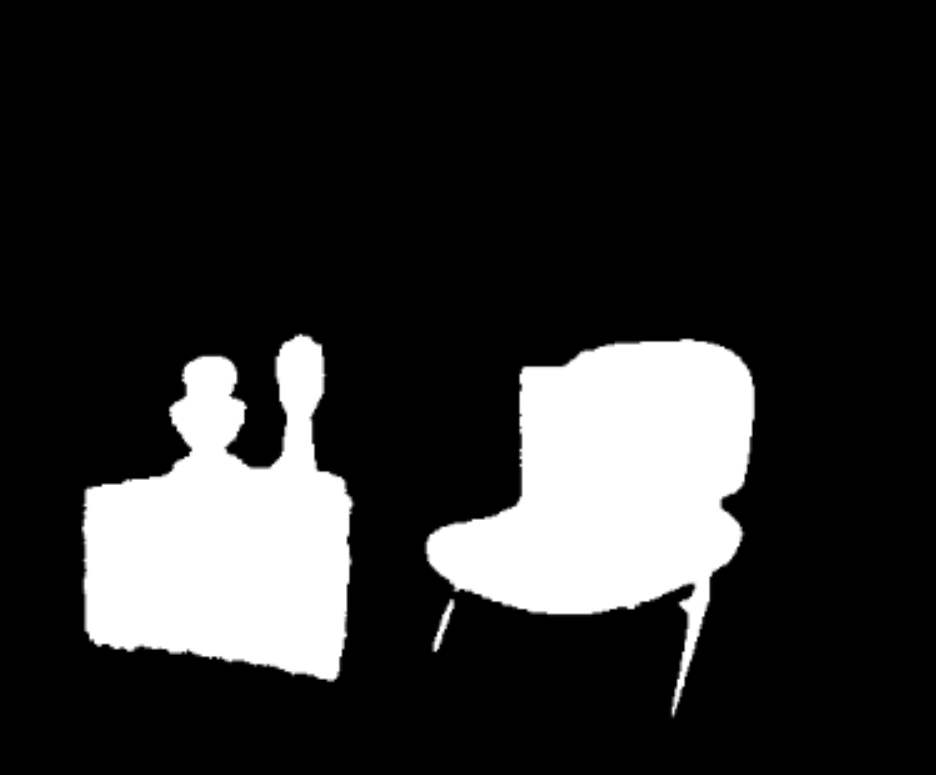}
      \end{minipage}
    \end{tabular}
    }
\caption{Results under different density conditions: (a) thin fog and (b) thick fog. In such a highly foggy scene, the accuracy of depth reconstruction is reduced where a scattering component has a large effect, but the proposed method can estimate an object region and improve a depth measurement regardless of medium density.}
\label{fig:various_density_result}
\end{figure*}

\begin{table*}[tb]
\centering
    \caption{Mean depth error on each object under different density conditions. w/o method denotes error before applying the proposed method.}
    \label{tab:mean_error}
    \begin{tabular}{lr|ccccc} \hline
      & & Plane & Chair & Desk & Hand & Duck \\ \hline
      Figure \ref{fig:various_density_result}(a) & w/o method & 117.26 mm & 198.79 mm & 459.73 mm & 425.21 mm & 574.28 mm \\
      (thin) & proposed & 17.97 mm & 65.01 mm & 83.40 mm & 32.04 mm & 45.08 mm \\ \hline
      Figure \ref{fig:medium_result} & w/o method & 253.38 mm & 372.02 mm & 656.11 mm & 669.64 mm & 798.17 mm \\
      (medium) & proposed & 20.67 mm & 60.27 mm & 106.68 mm & 50.63 mm & 118.21 mm \\ \hline
      Figure \ref{fig:various_density_result}(b) & w/o method & 421.31 mm & 531.48 mm & 798.85 mm & 844.32 mm & 953.56 mm \\
      (thick) & proposed & 19.95 mm & 70.63 mm & 202.36 mm & 75.79 mm & 143.20 mm \\ \hline
    \end{tabular}
\end{table*} 

\begin{figure*}[tb]
\centering
  \subfloat[Scene]{
    \includegraphics[scale=\figurescaleR]{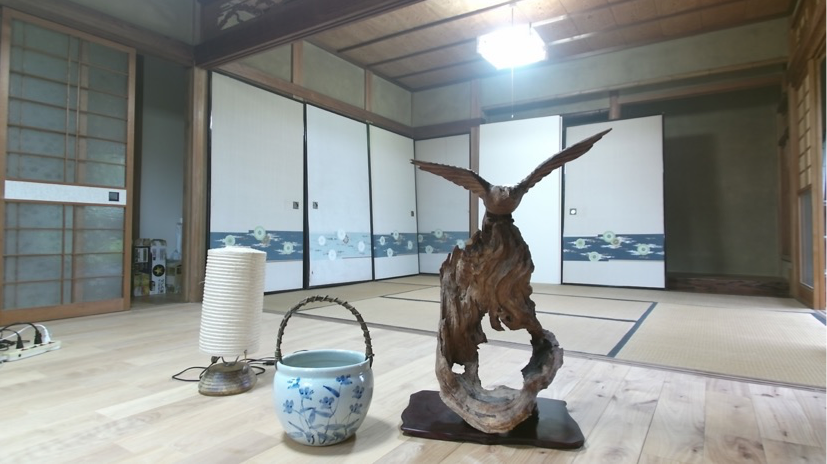}\quad
    \includegraphics[scale=\figurescaleR]{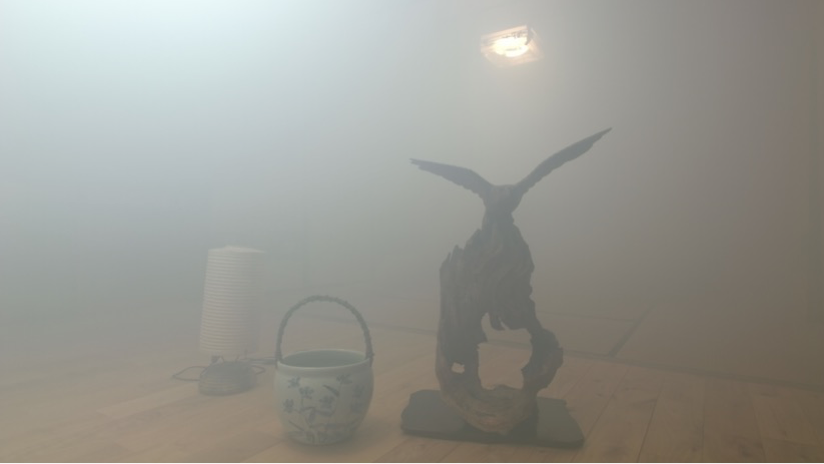}}\\
  \subfloat[Amplitude]{
    \includegraphics[scale=\figurescaleR]{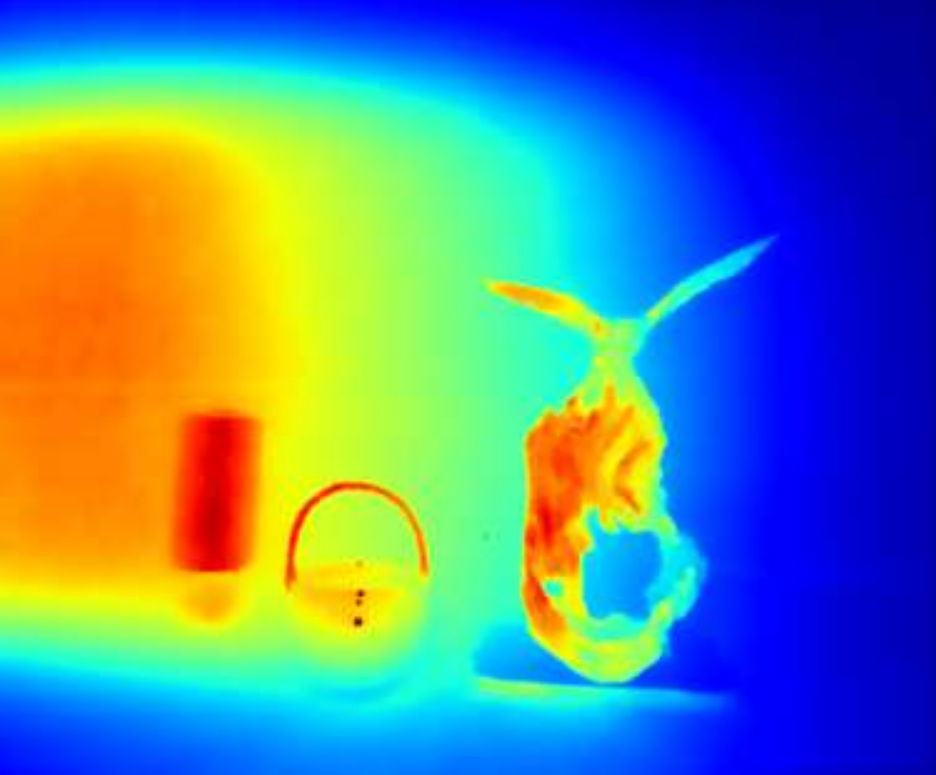}
    \includegraphics[scale=\figurescaleR]{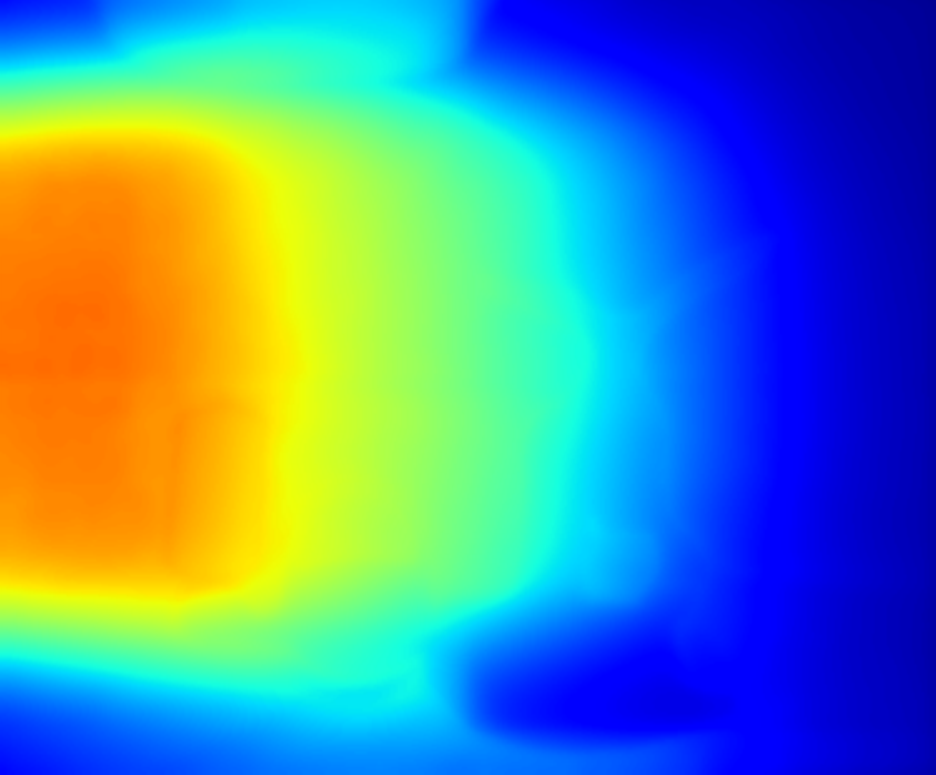}
    \includegraphics[scale=\figurescaleR]{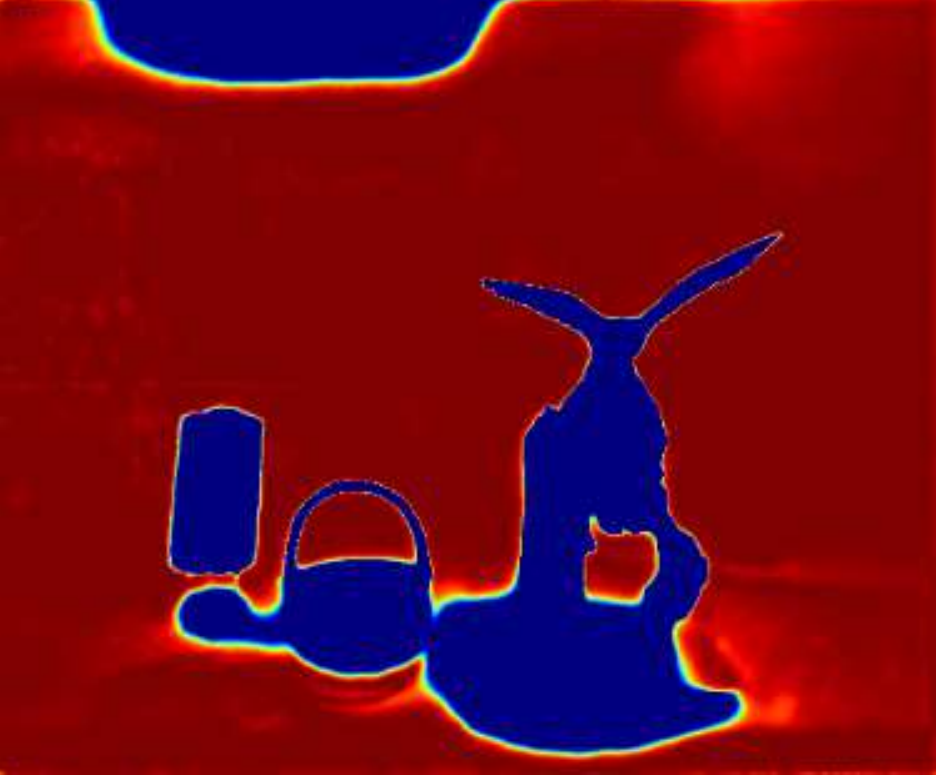}}\,
  \subfloat[Phase]{
    \includegraphics[scale=\figurescaleR]{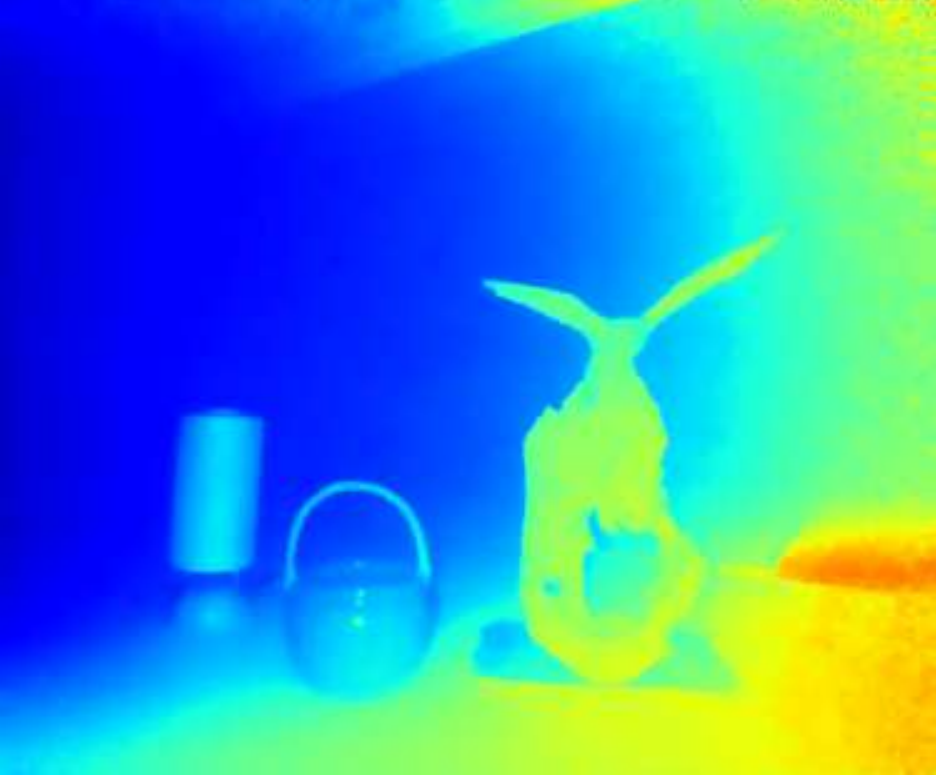}
    \includegraphics[scale=\figurescaleR]{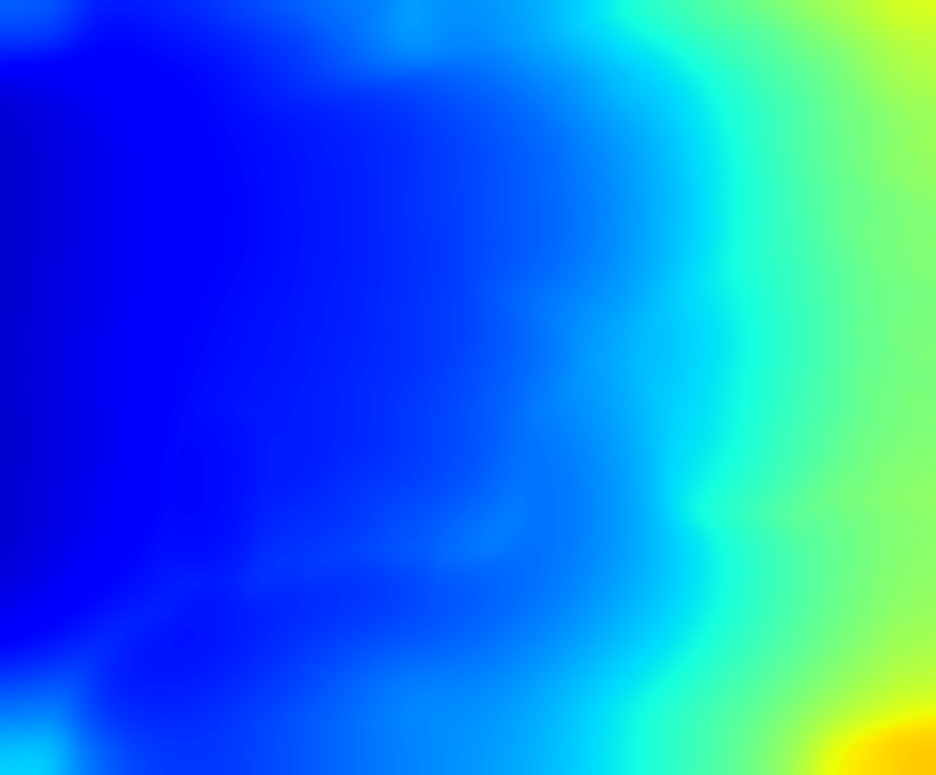}
    \includegraphics[scale=\figurescaleR]{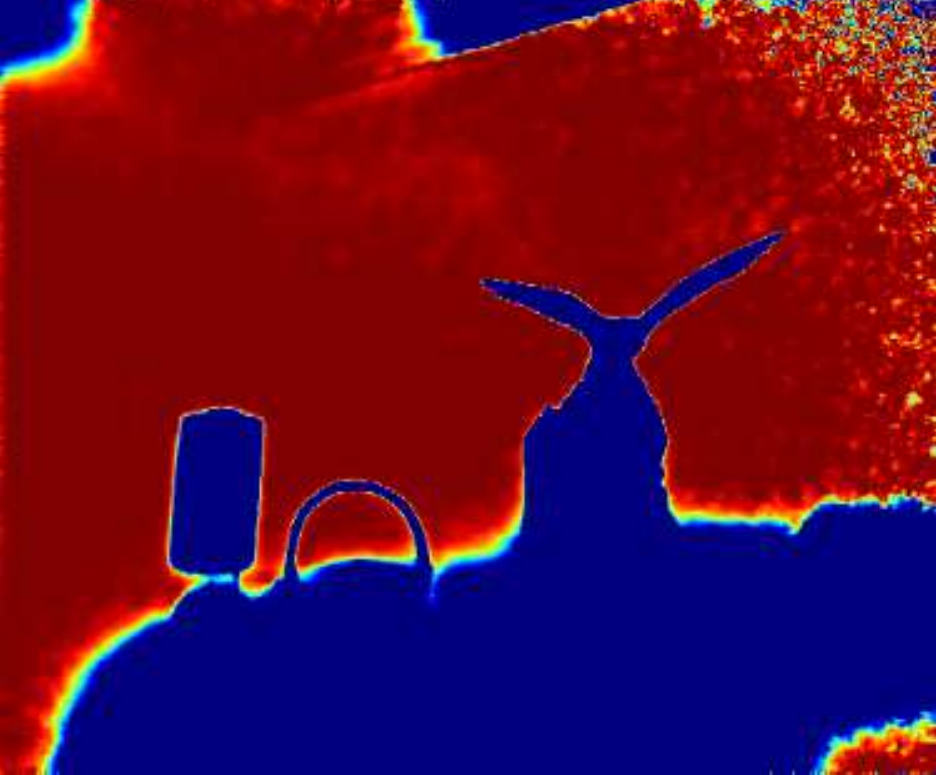}}\\
  \subfloat[Depth reconstruction]{
    \includegraphics[scale=\figurescaleR]{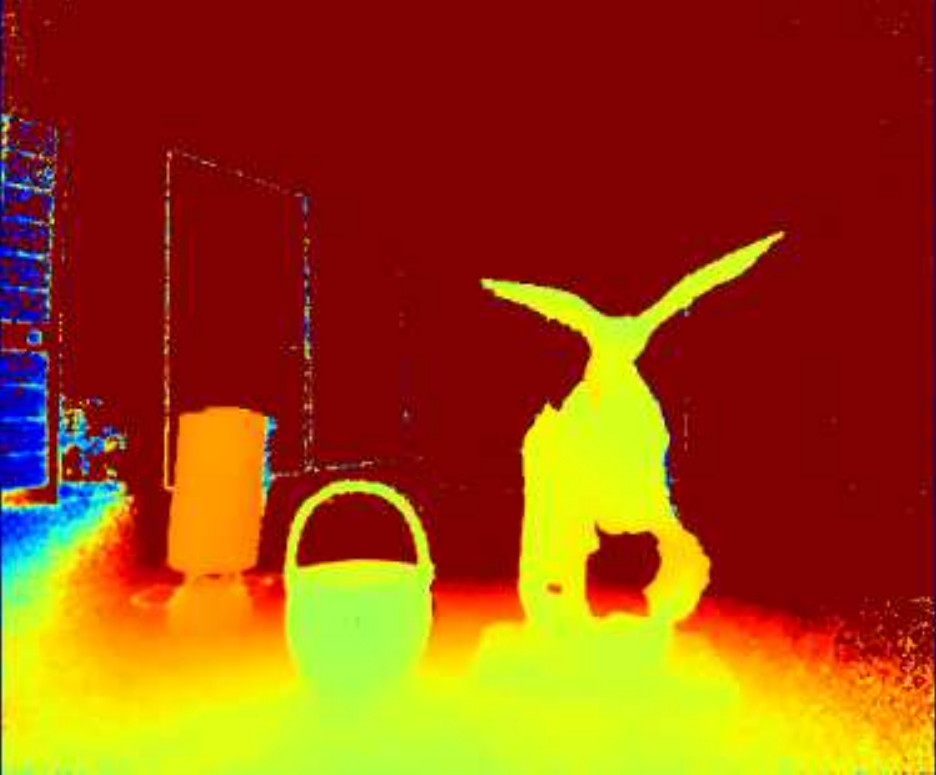}\quad 
    \includegraphics[scale=\figurescaleR]{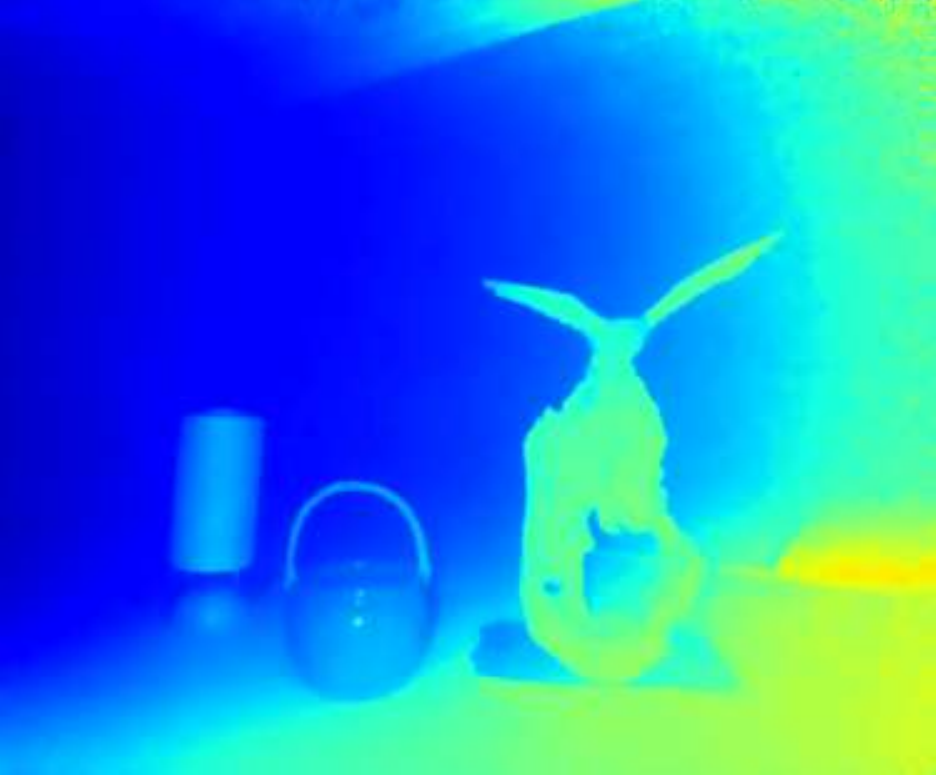}\quad
    \includegraphics[scale=\figurescaleR]{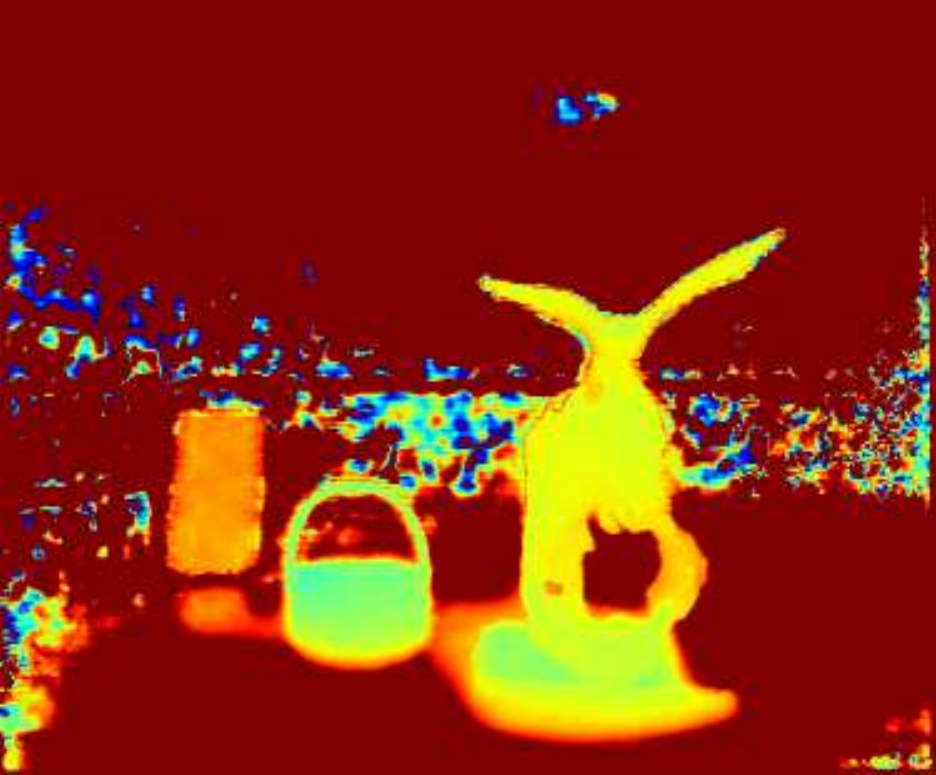}\quad
    \includegraphics[scale=\figurescaleR]{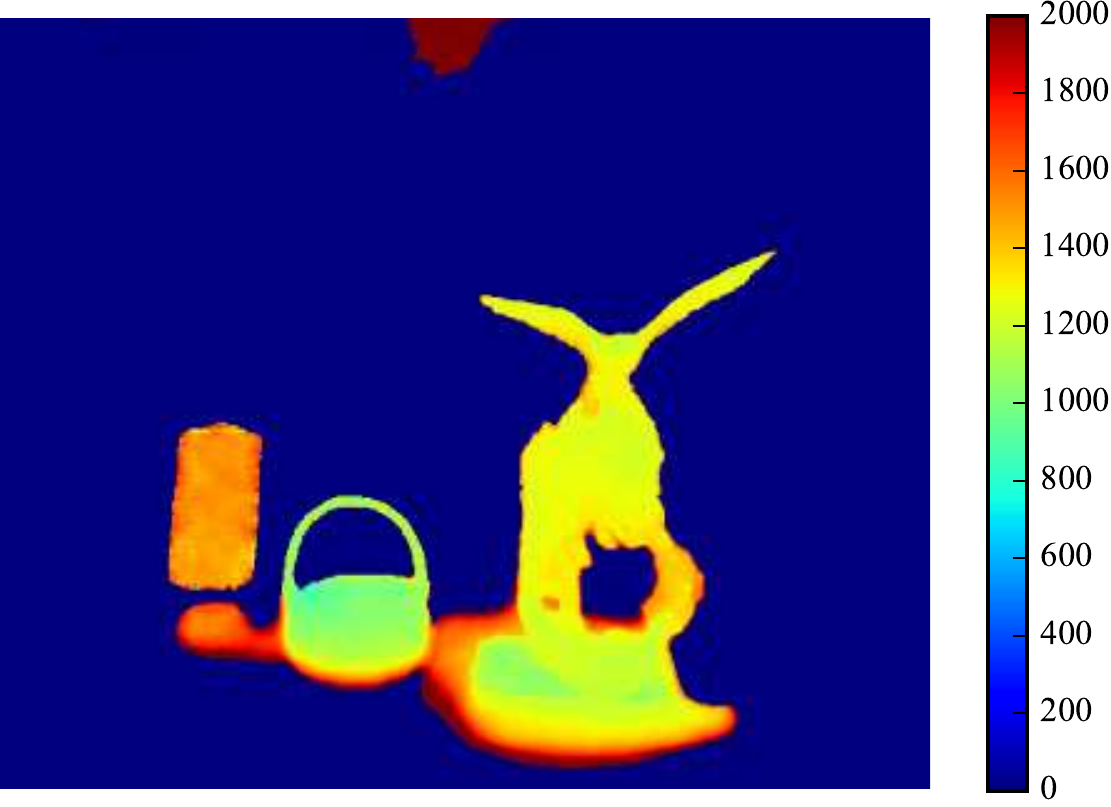}\quad
    \includegraphics[scale=\figurescaleR]{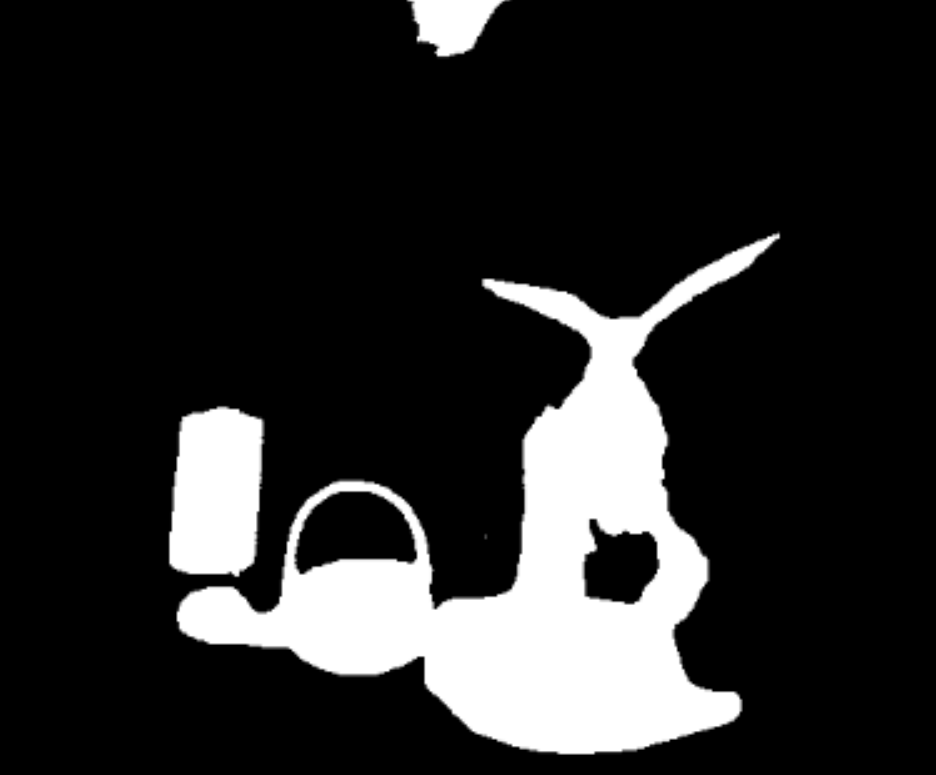}}
\caption{Results for a more realistic scene. (a) Target scene without and with fog. (b)(c) Left to right: input image, estimated scattering component, and IRLS weight for the amplitude and phase image, respectively. (d) Left to right: depth without fog, depth with fog, reconstructed depth, masked depth, and estimated object mask.}
\label{fig:realistic_result}
\end{figure*}

\begin{figure*}[tb]
  \centering
    \includegraphics[width=0.9\textwidth]{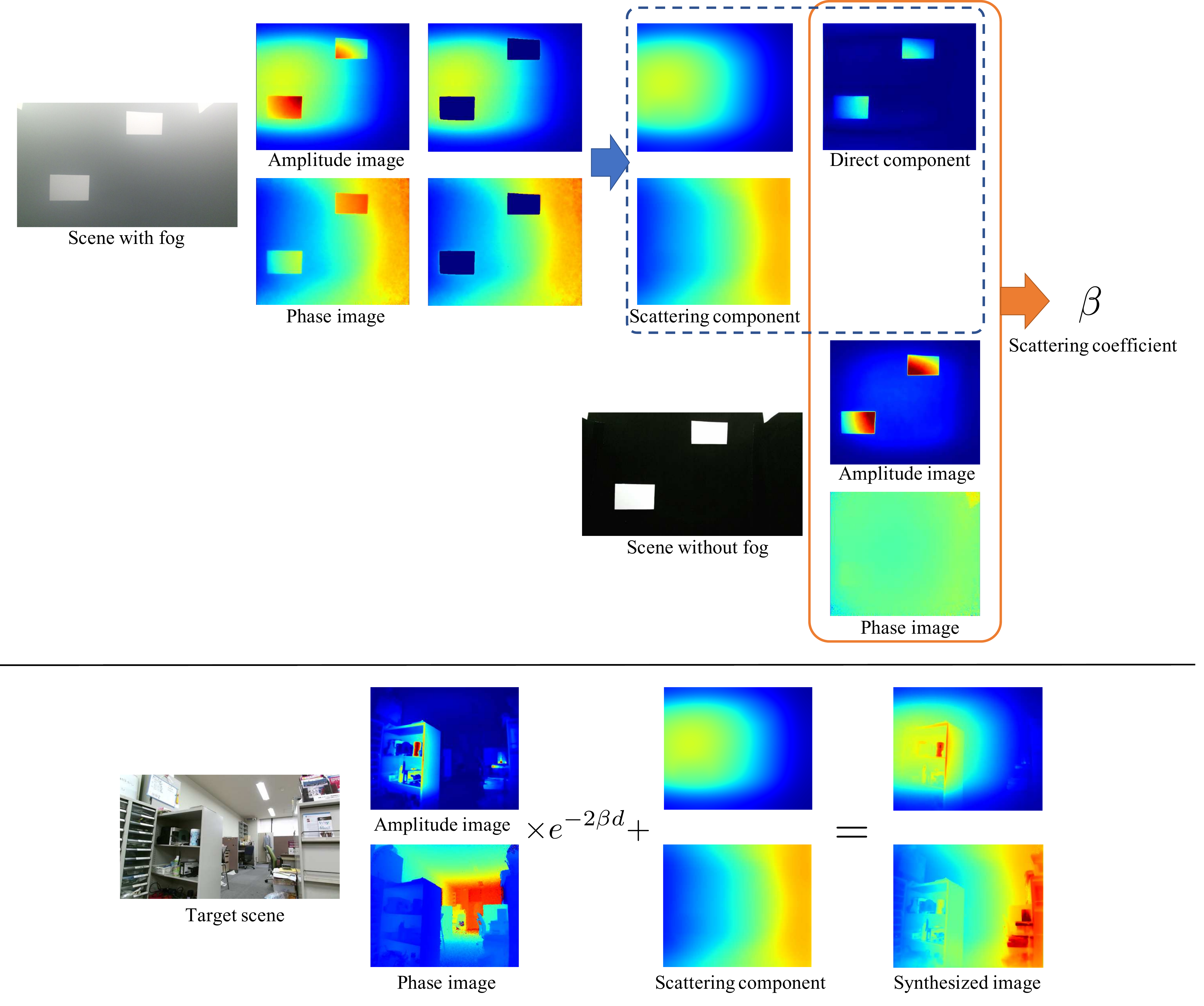}
    \caption{Procedure of synthesizing images. First, we captured a scene that has calibration objects in a foggy scene  and masked the region of the calibration objects manually. After that, we compensated for the defective region to estimate the scattering component. Using the observation without fog, the scattering coefficient can be computed. The images of a target scene without fog were captured separately, and the attenuated direct component and the estimated scattering component were combined into synthesized images.}
    \label{fig:synthesized_overview}
\end{figure*}

\begin{figure*}[p]
\centering
    \begin{tabular}{c}
      \begin{minipage}{0.16\hsize}
        \centering
          \subfloat[Scene]{\includegraphics[scale=\figurescale]{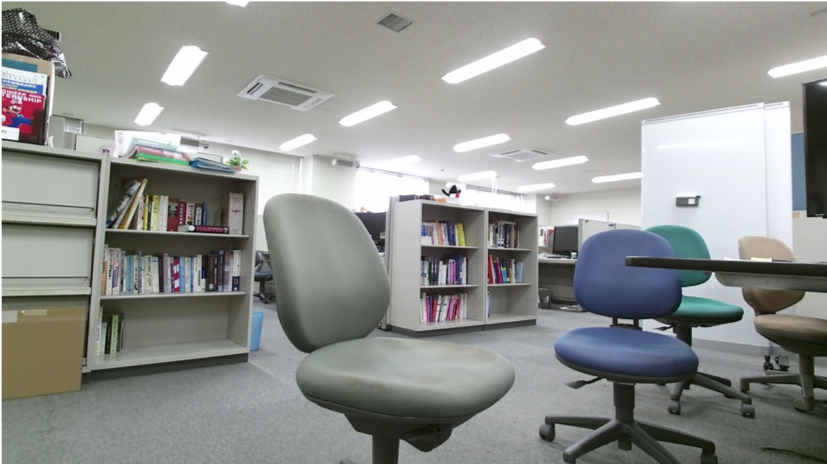}}
      \end{minipage}
      \begin{minipage}{0.8\hsize}
        \centering
          \subfloat[Amplitude]{\includegraphics[scale=\figurescale]{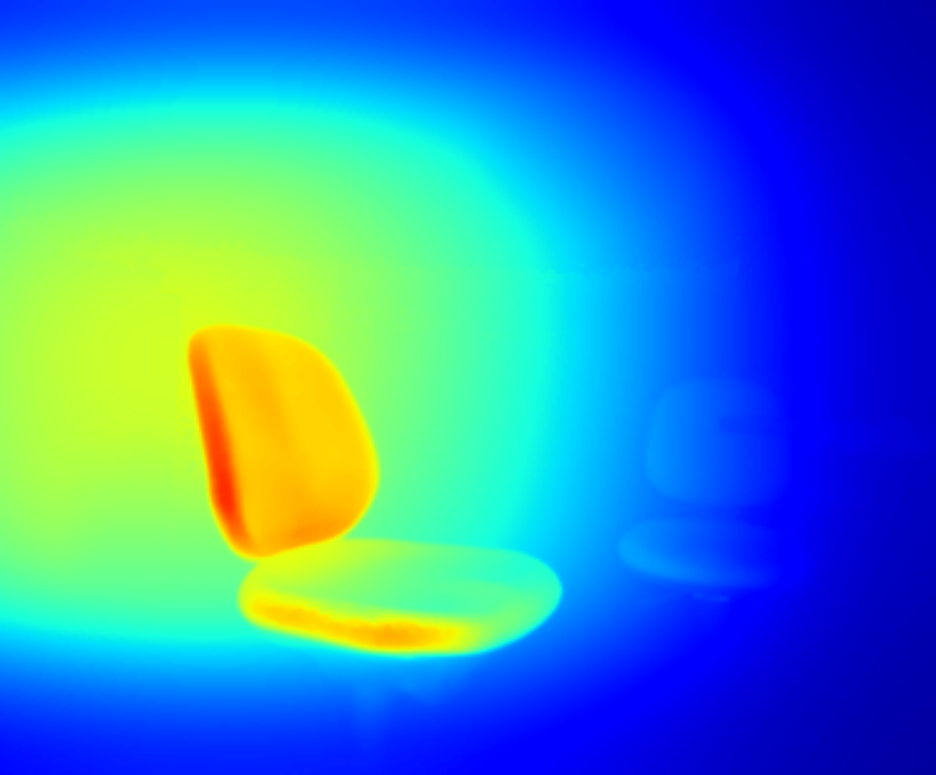}\quad
                          \includegraphics[scale=\figurescale]{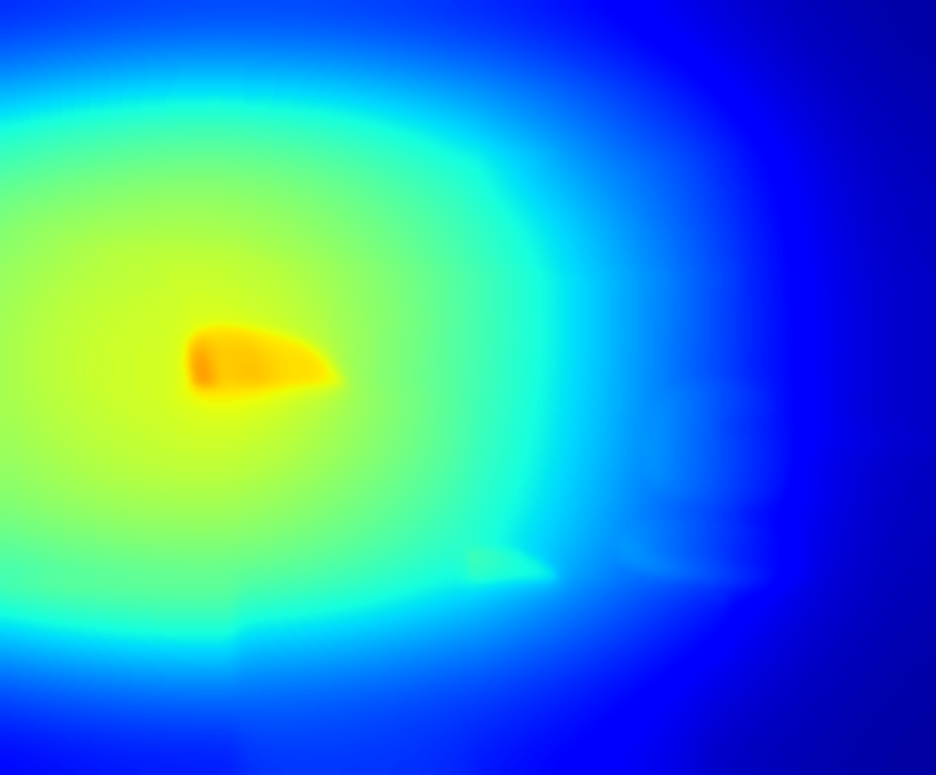}\quad
                          \includegraphics[scale=\figurescale]{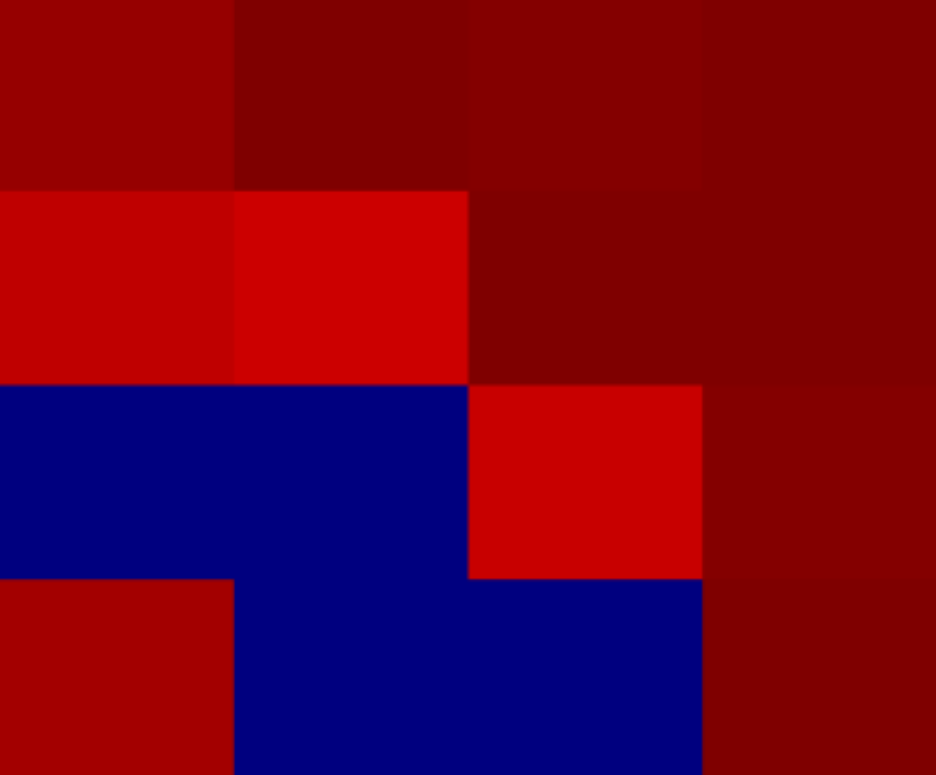}\quad
                          \includegraphics[scale=\figurescale]{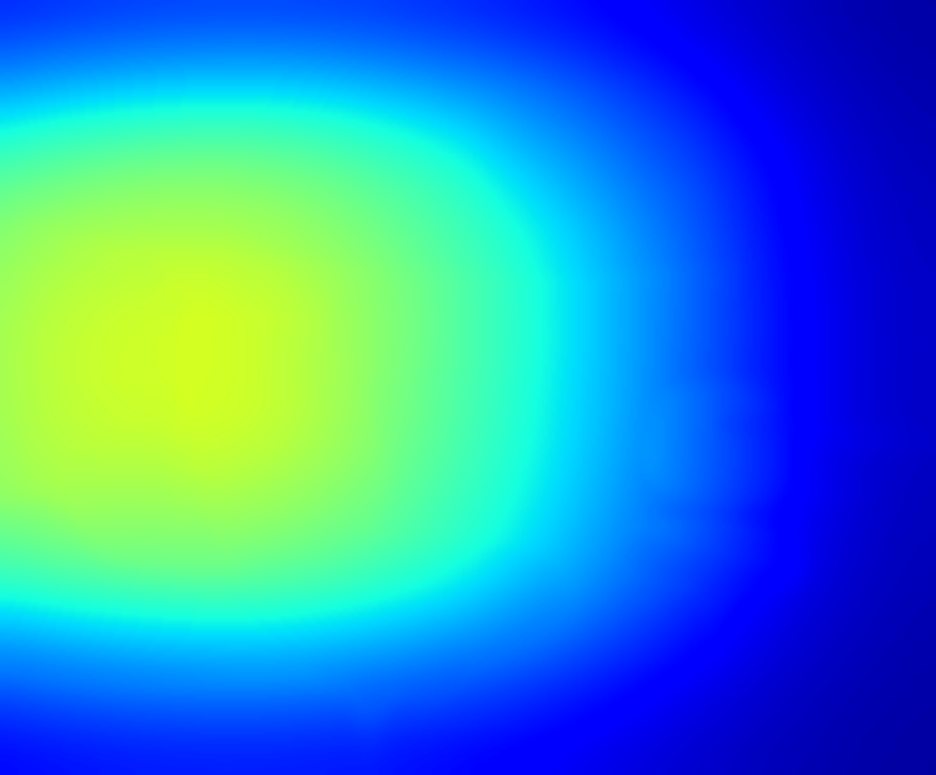}\quad
                          \includegraphics[scale=\figurescale]{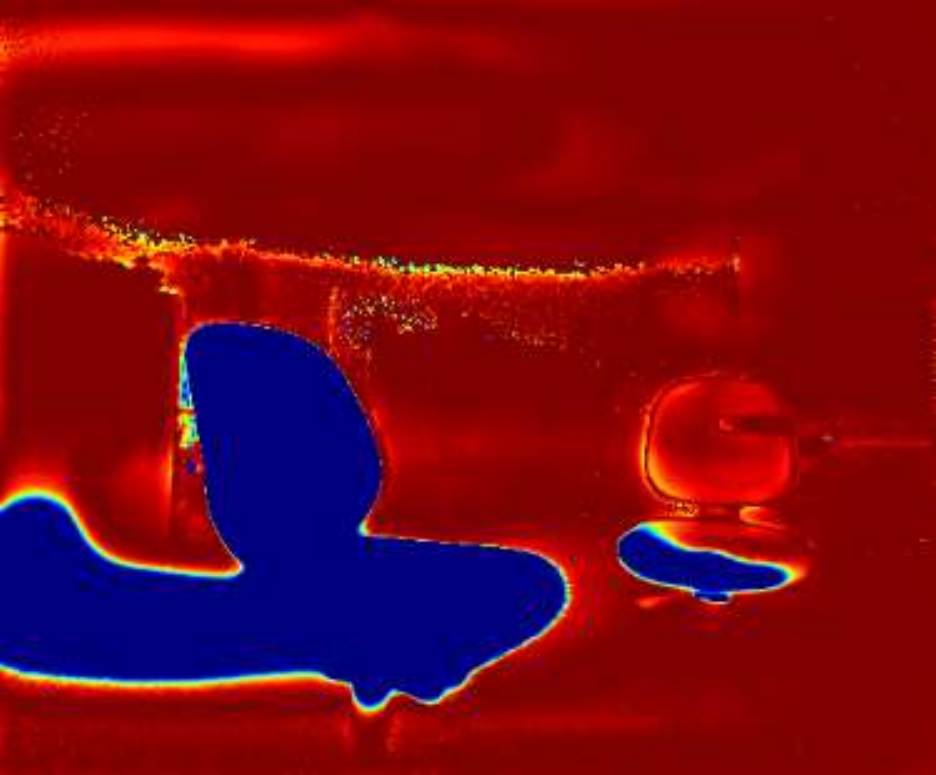}}\\ [-0.1ex]
          \subfloat[Phase]{\includegraphics[scale=\figurescale]{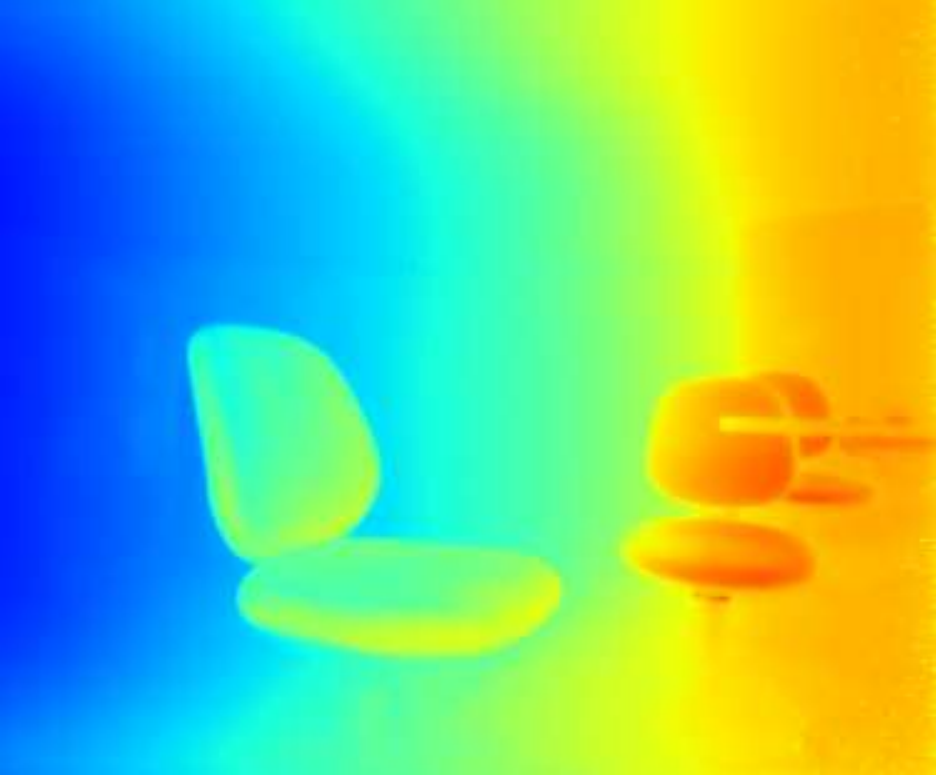}\quad
                          \includegraphics[scale=\figurescale]{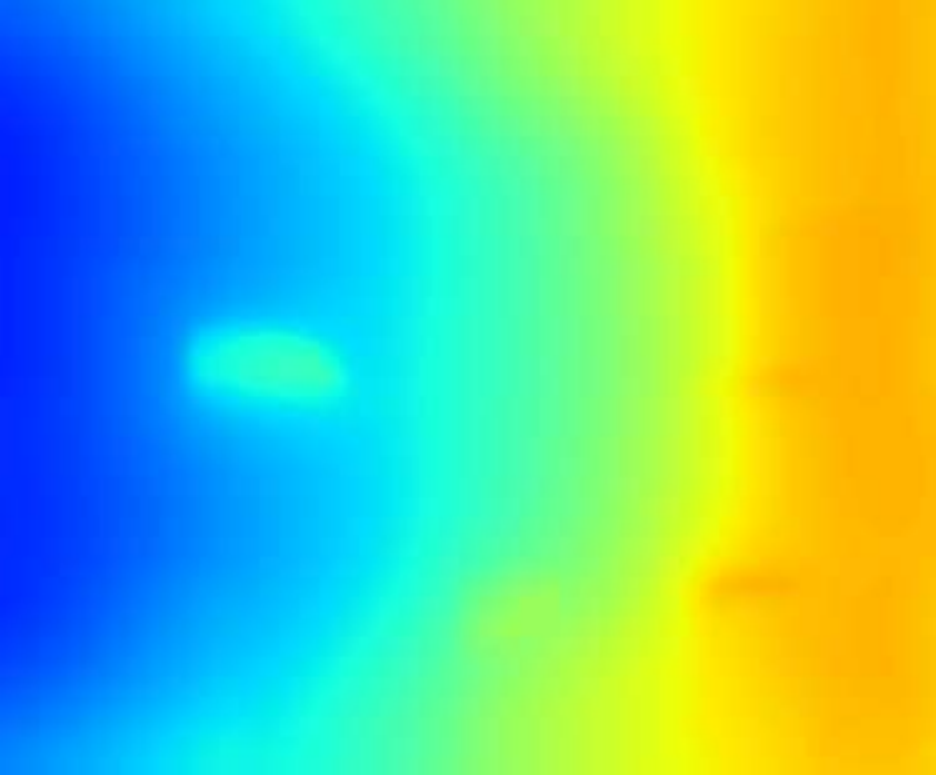}\quad
                          \includegraphics[scale=\figurescale]{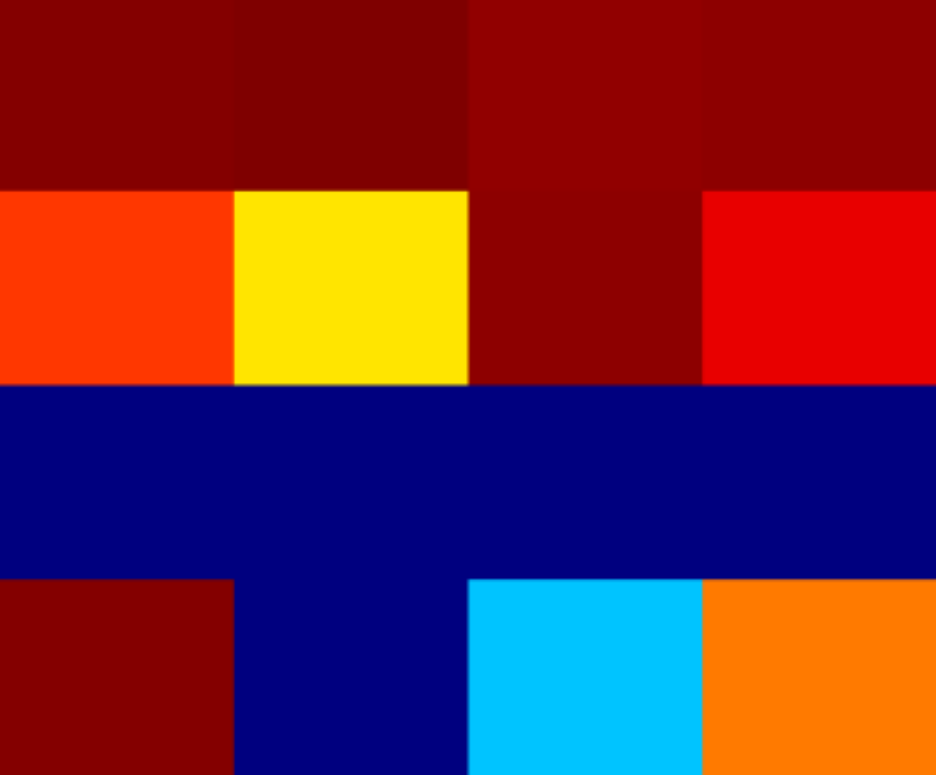}\quad
                          \includegraphics[scale=\figurescale]{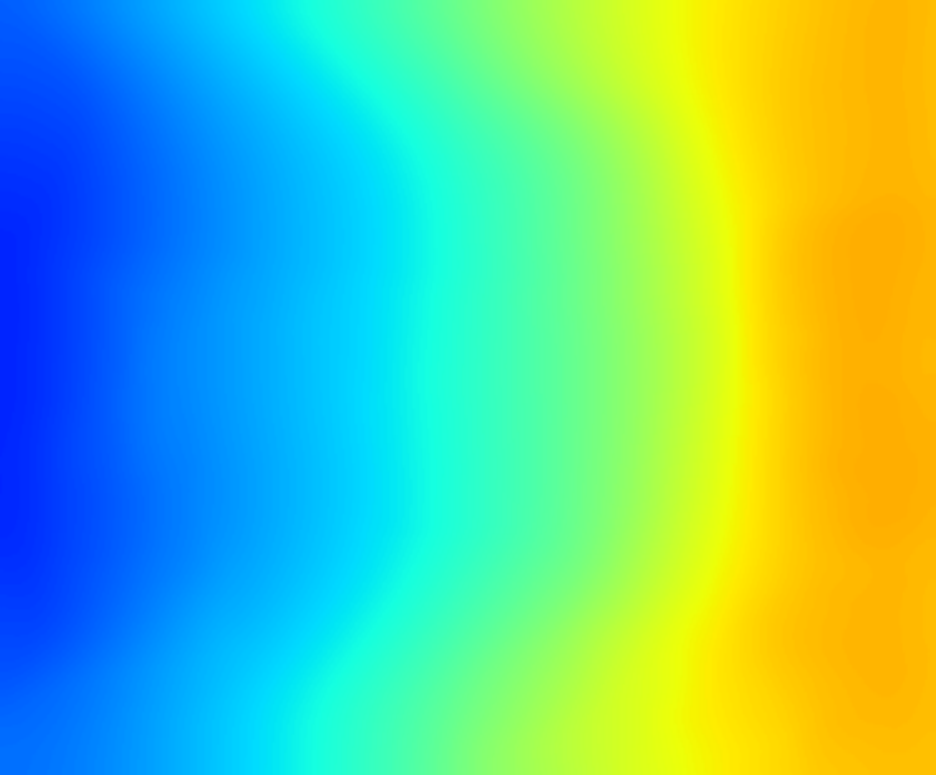}\quad
                          \includegraphics[scale=\figurescale]{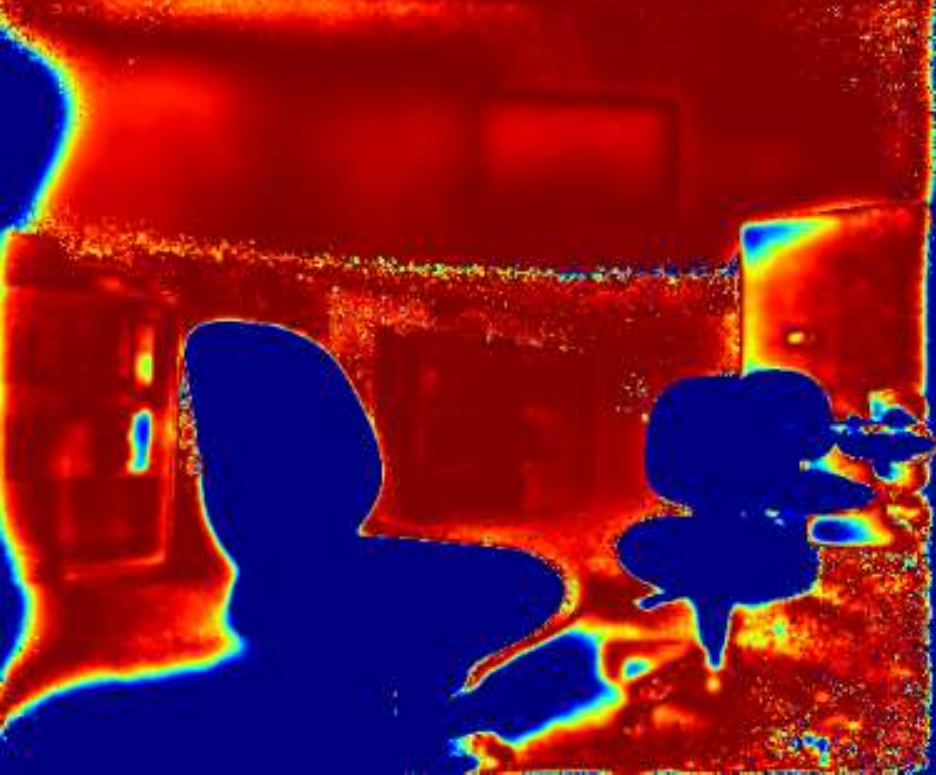}}\\[-0.1ex]
          \subfloat[Depth reconstruction]{\includegraphics[scale=\figurescale]{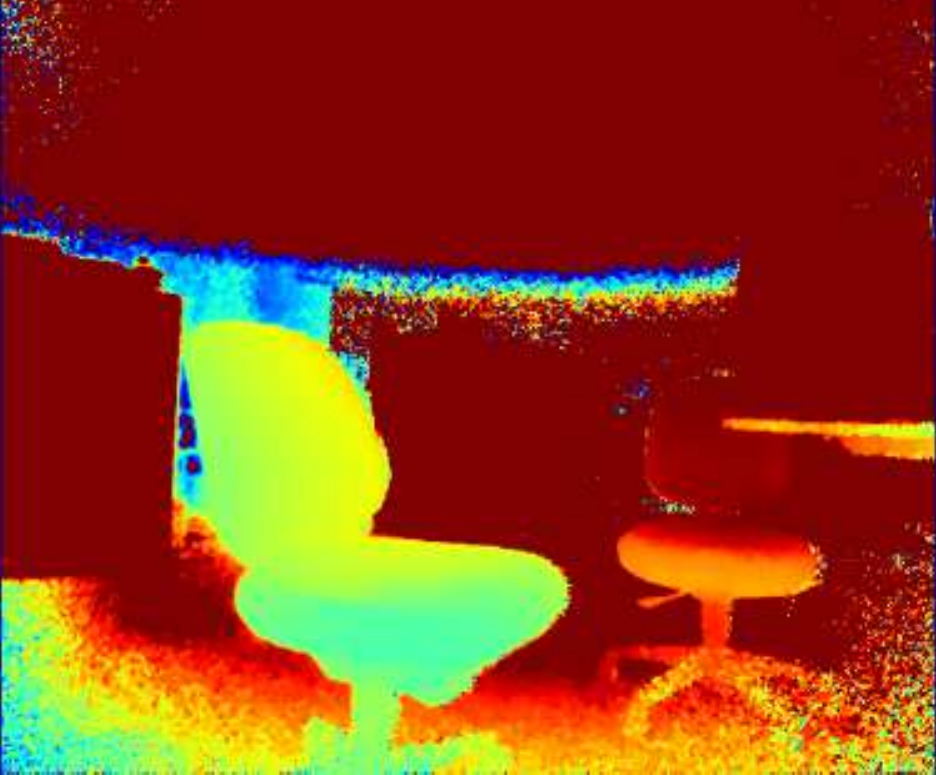}\quad
                          \includegraphics[scale=\figurescale]{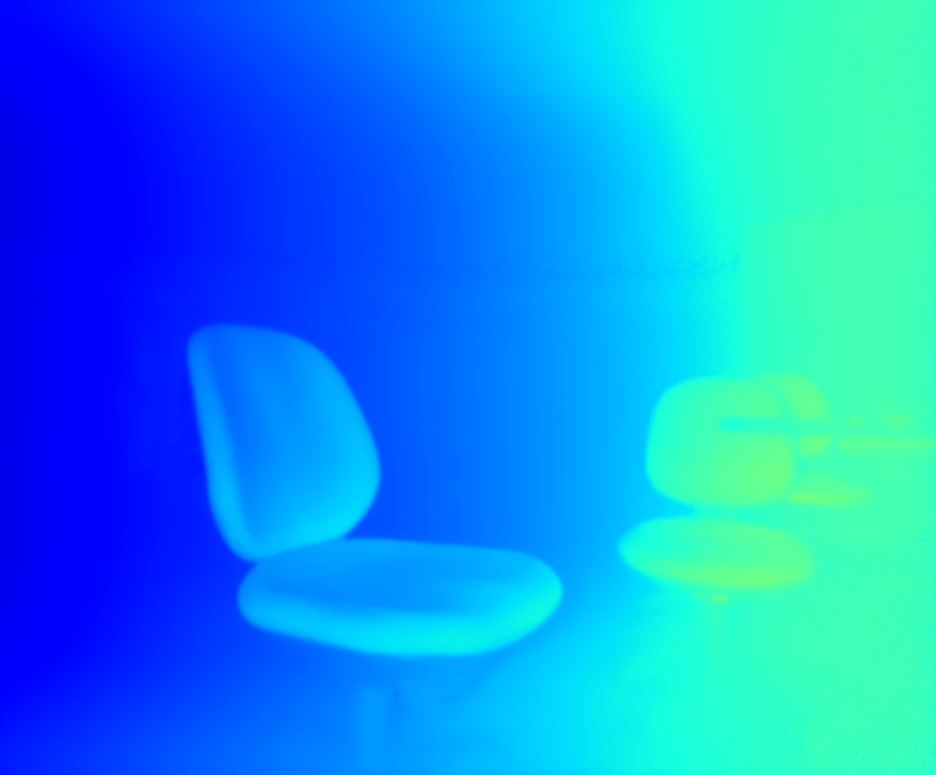}\quad
                          \includegraphics[scale=\figurescale]{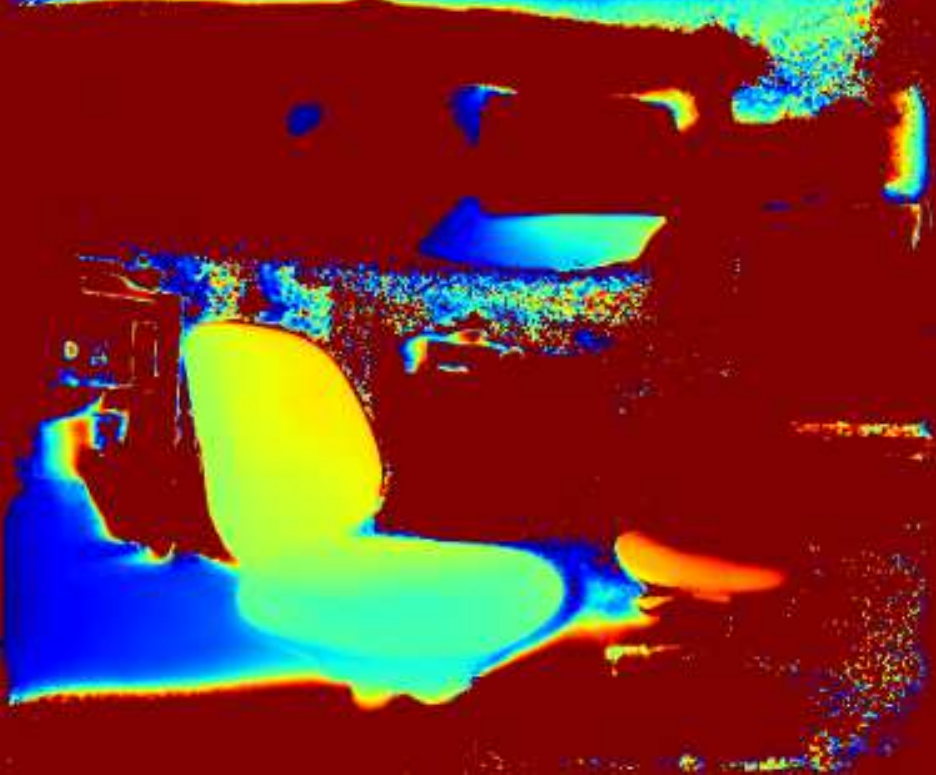}\quad
                          \includegraphics[scale=\figurescale]{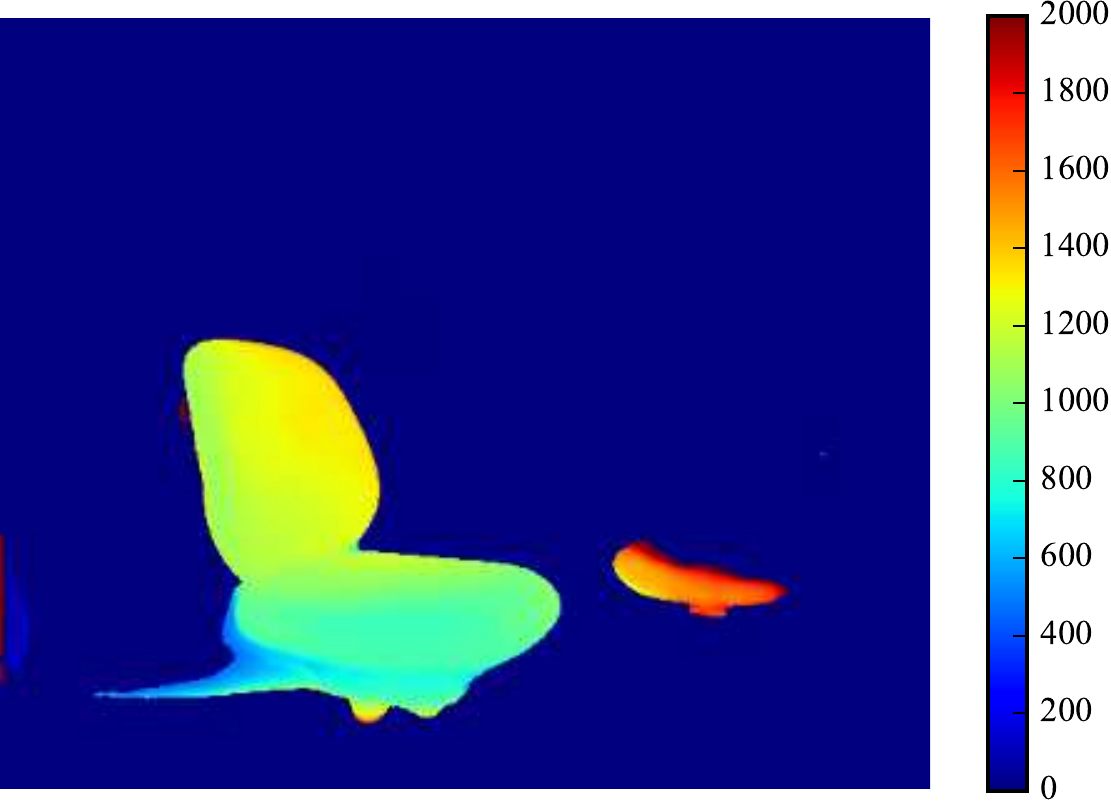}\quad
                          \includegraphics[scale=\figurescale]{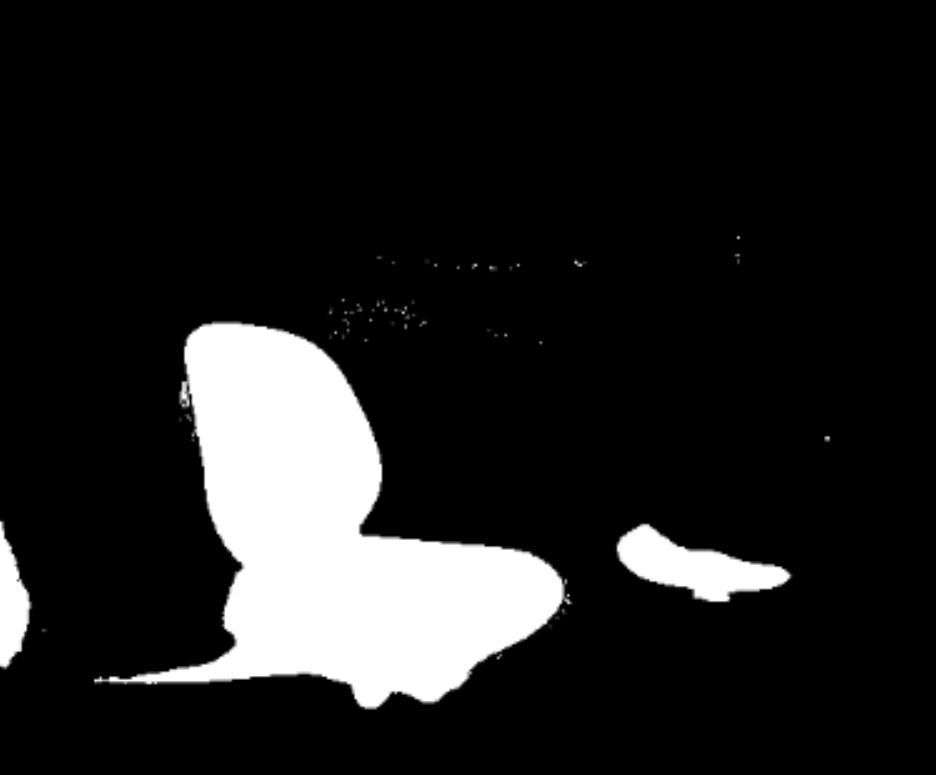}}\\[-0.1ex]
          \subfloat[Only fine optimization]{\includegraphics[scale=\figurescale]{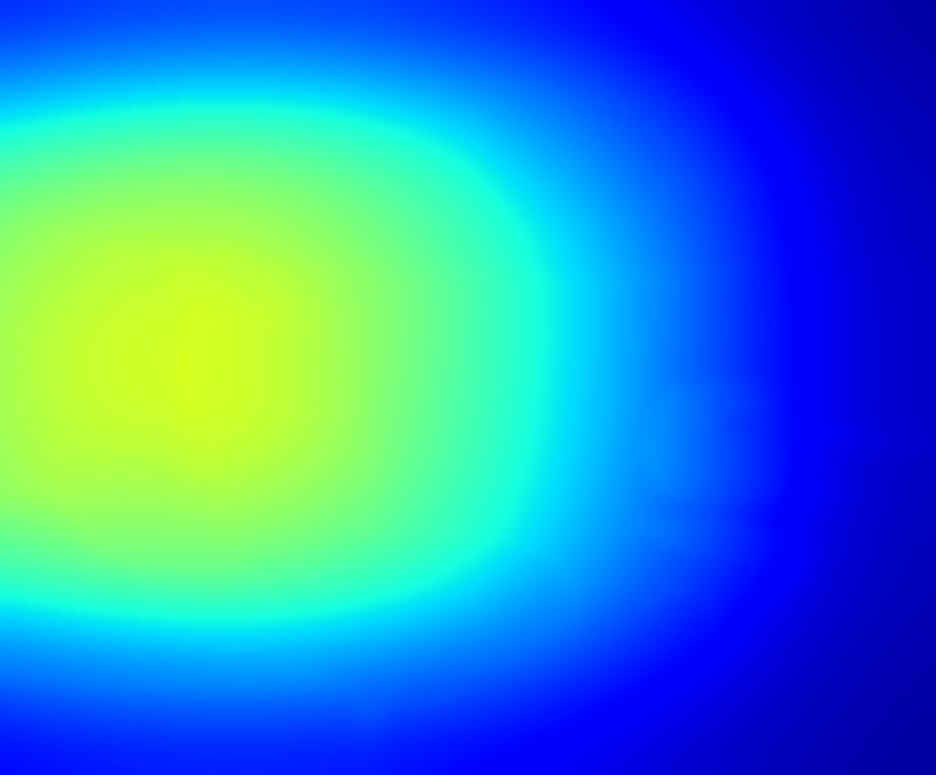}\quad
                          \includegraphics[scale=\figurescale]{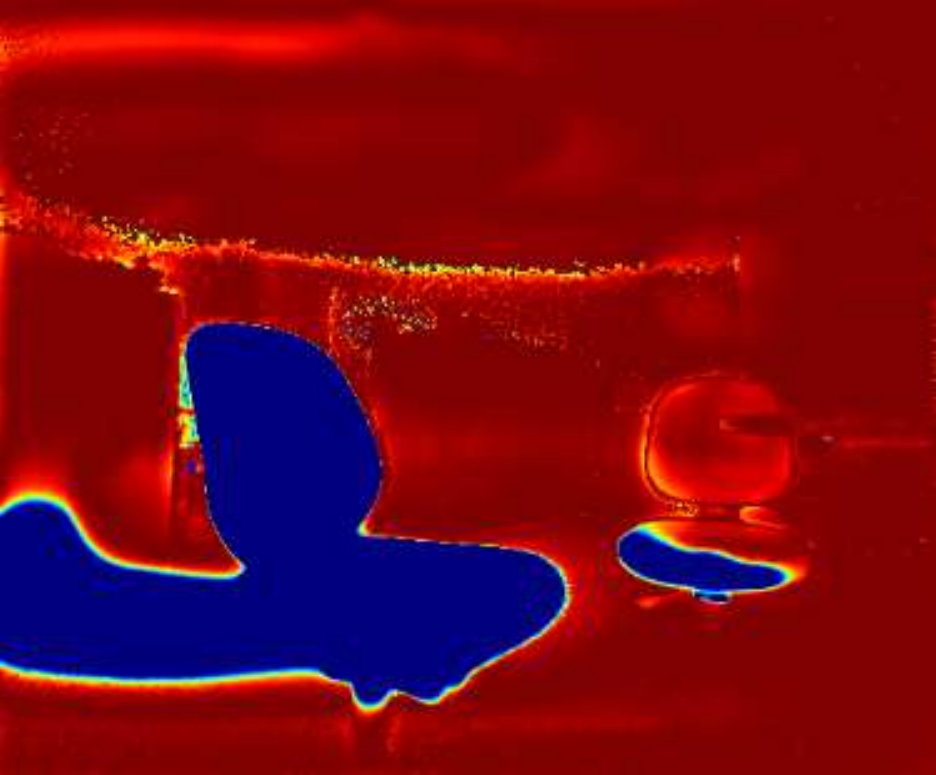}\quad
                          \includegraphics[scale=\figurescale]{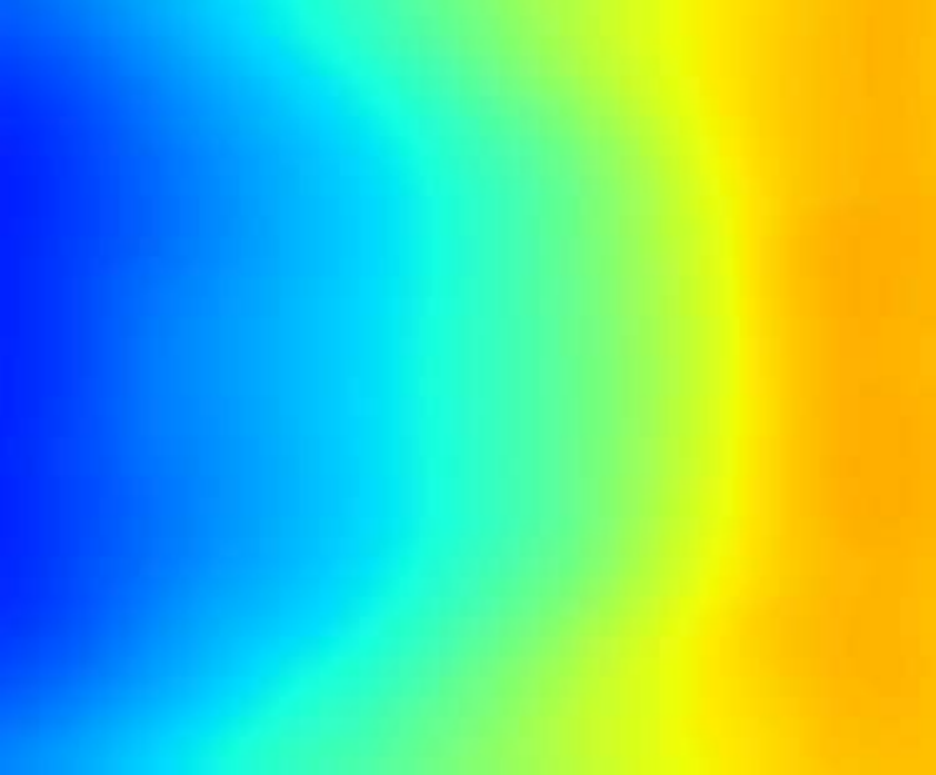}\quad
                          \includegraphics[scale=\figurescale]{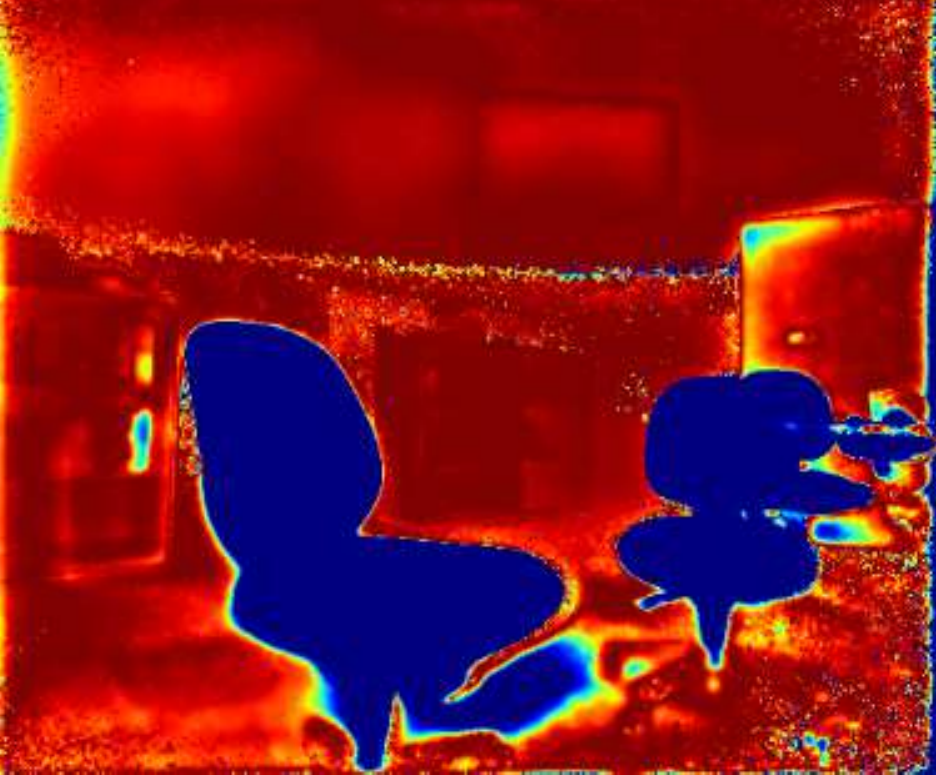}\quad
                          \includegraphics[scale=\figurescale]{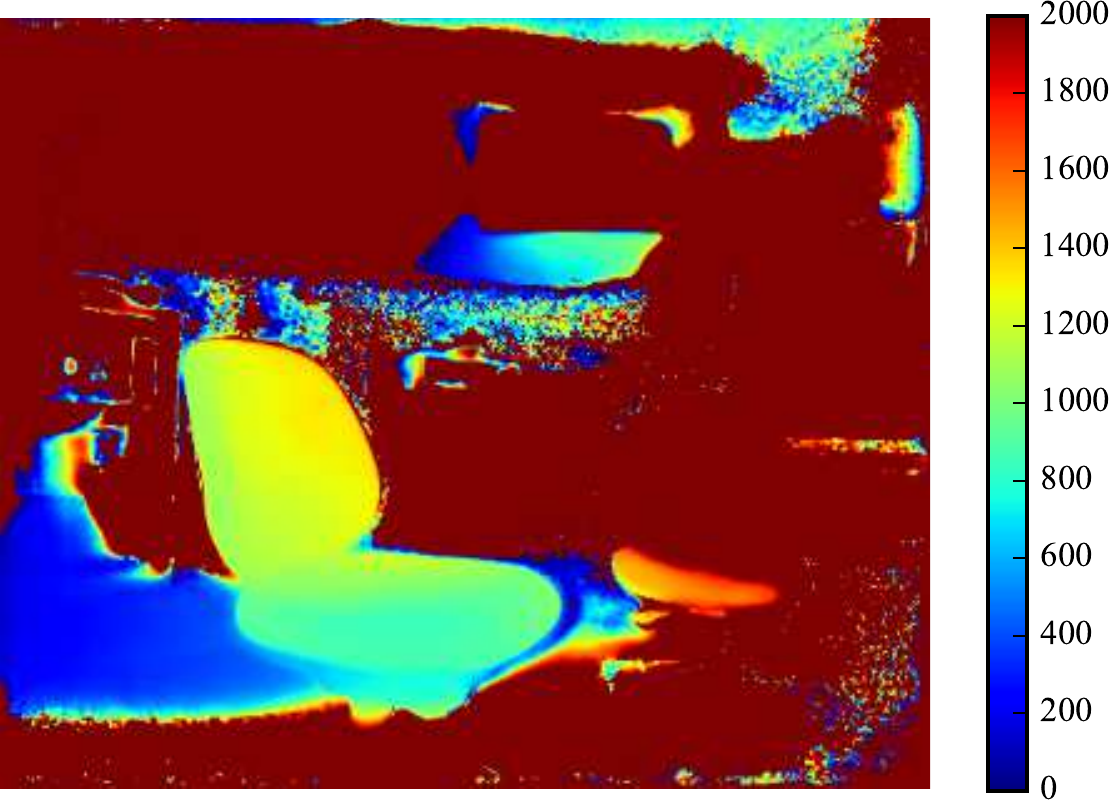}}
      \end{minipage}
    \end{tabular}
\caption{(a) Target scene. (b)(c) Left to right: input image, estimated scattering component and weight of coarse level, and of fine level for the amplitude and phase image, respectively. (d) Left to right: depth without fog, depth with fog, reconstructed depth, masked depth, and estimated object mask. (e) Result of applying only fine optimization. The estimated scattering component and the weight for the amplitude image, for the phase image, and the reconstructed depth from left to right.}
\label{fig:chair_result}
\medskip
    \begin{tabular}{c}
      \begin{minipage}{0.16\hsize}
        \centering
          \subfloat[Scene]{\includegraphics[scale=\figurescale]{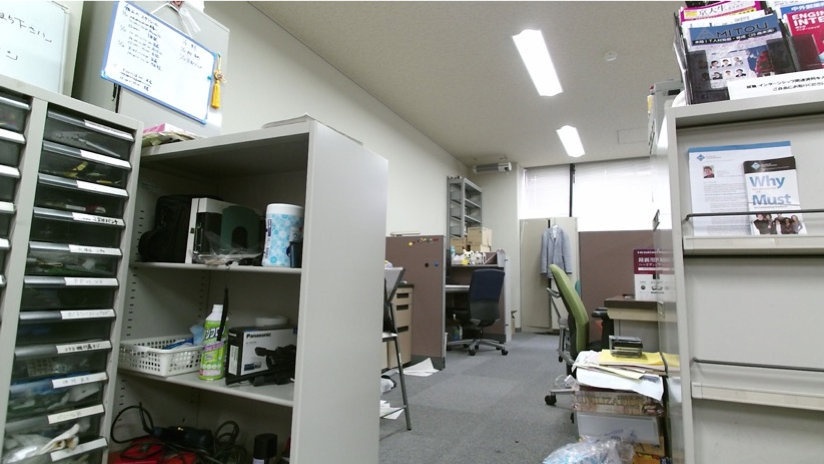}}
      \end{minipage}
      \begin{minipage}{0.8\hsize}
        \centering
          \subfloat[Amplitude]{\includegraphics[scale=\figurescale]{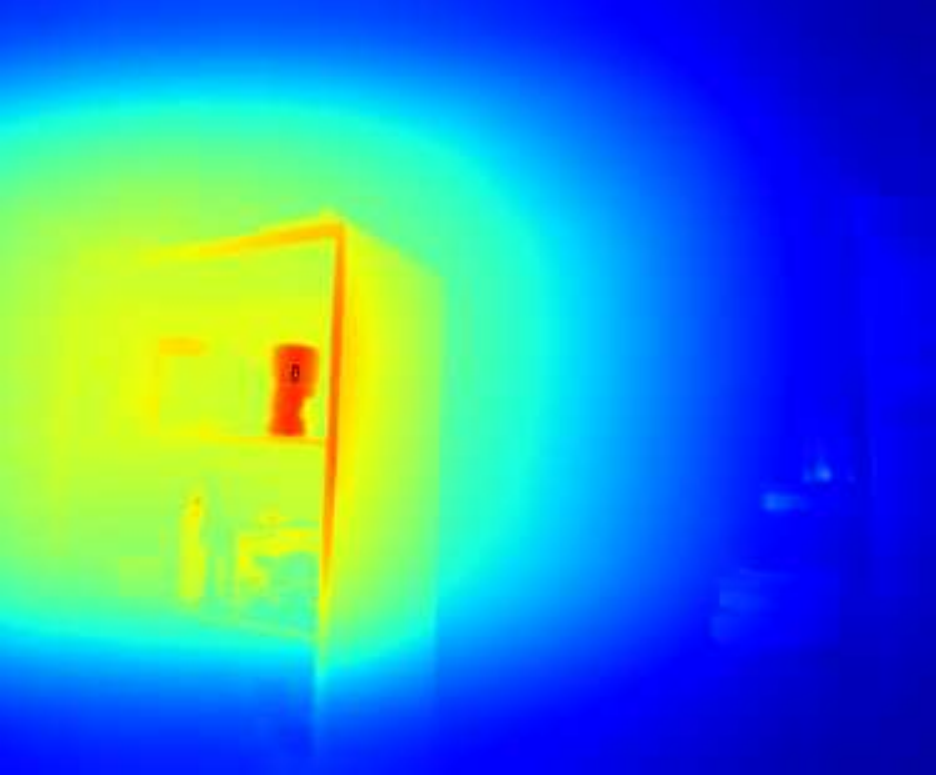}\quad
                          \includegraphics[scale=\figurescale]{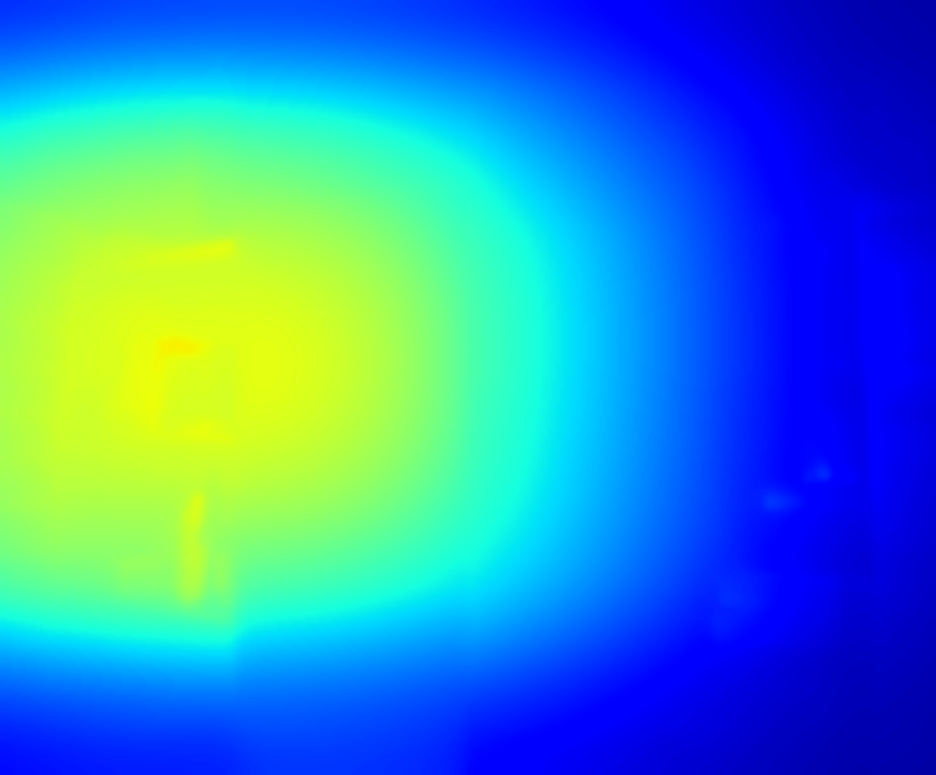}\quad
                          \includegraphics[scale=\figurescale]{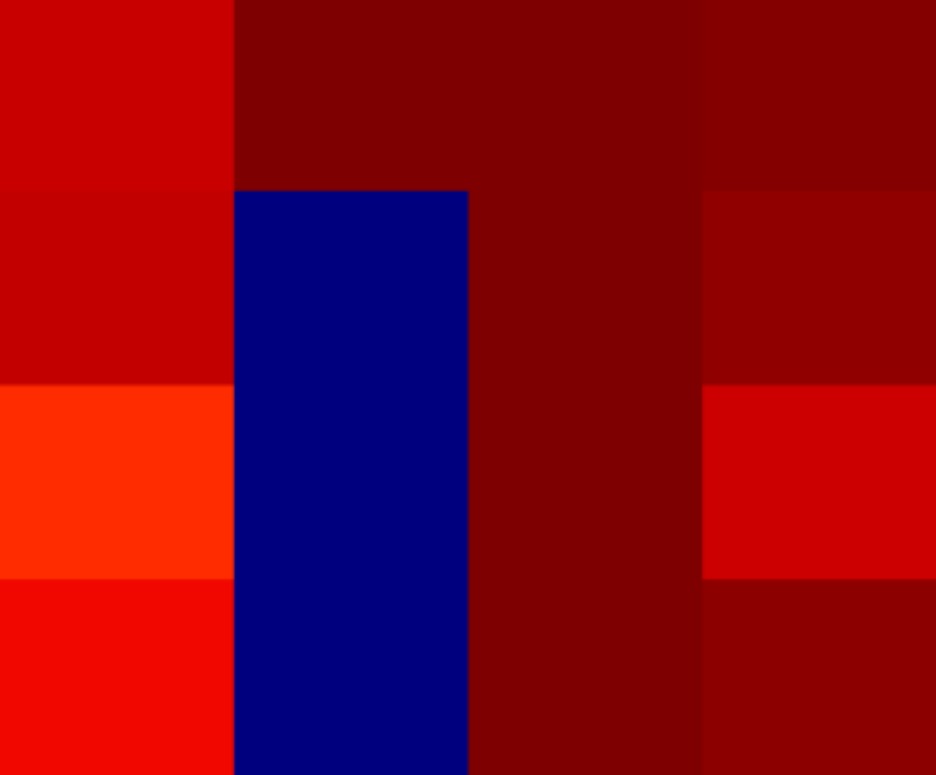}\quad
                          \includegraphics[scale=\figurescale]{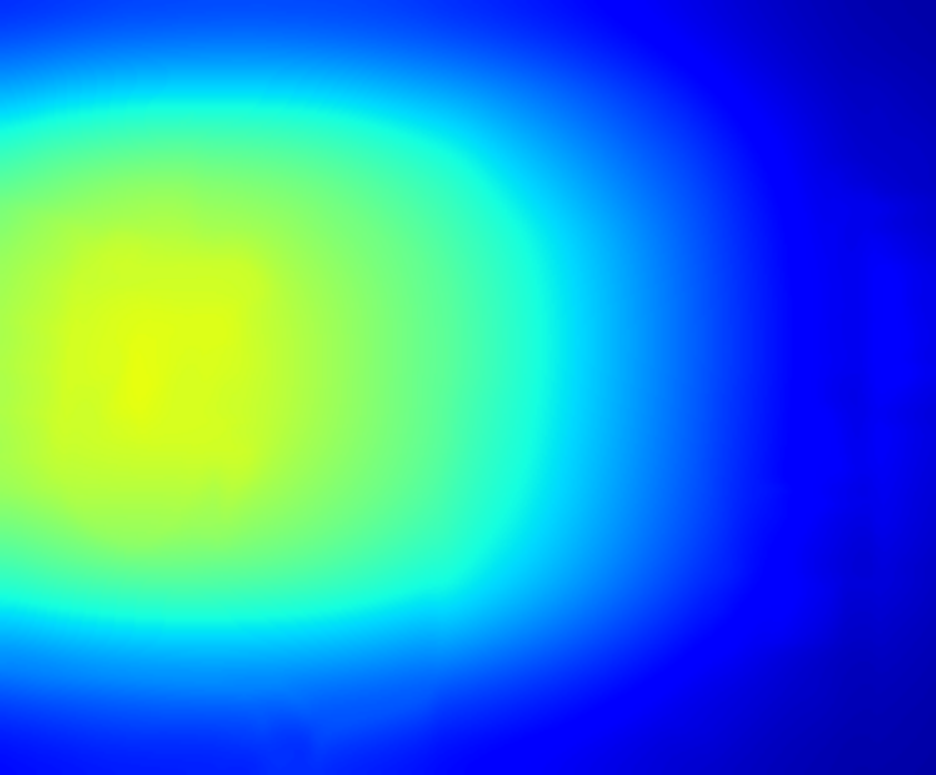}\quad
                          \includegraphics[scale=\figurescale]{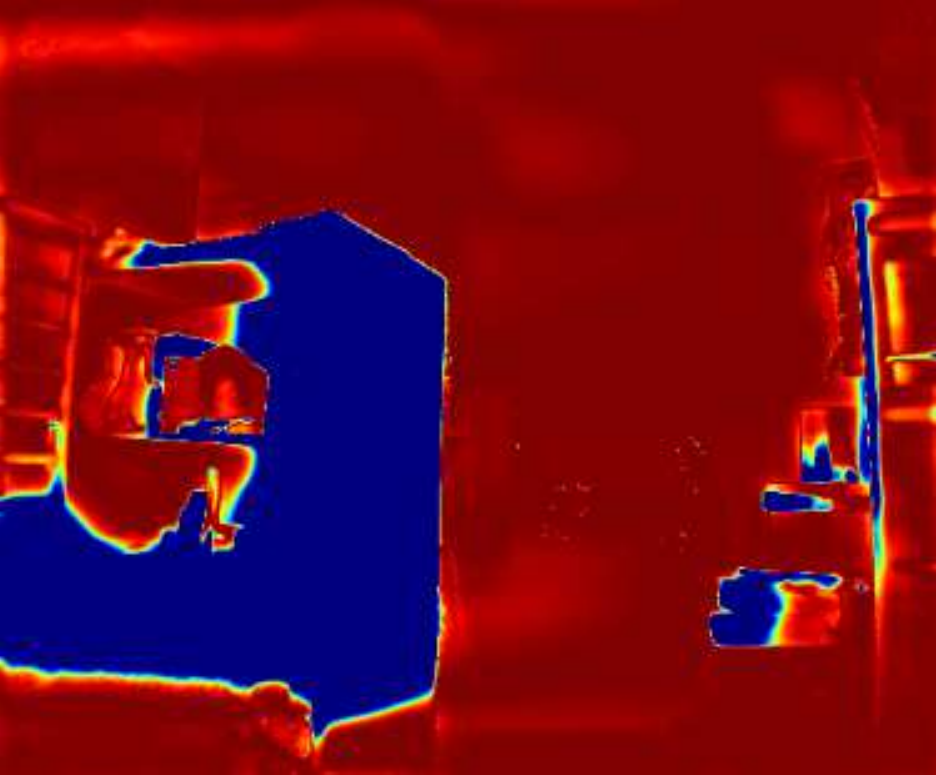}}\\[-0.1ex]
          \subfloat[Phase]{\includegraphics[scale=\figurescale]{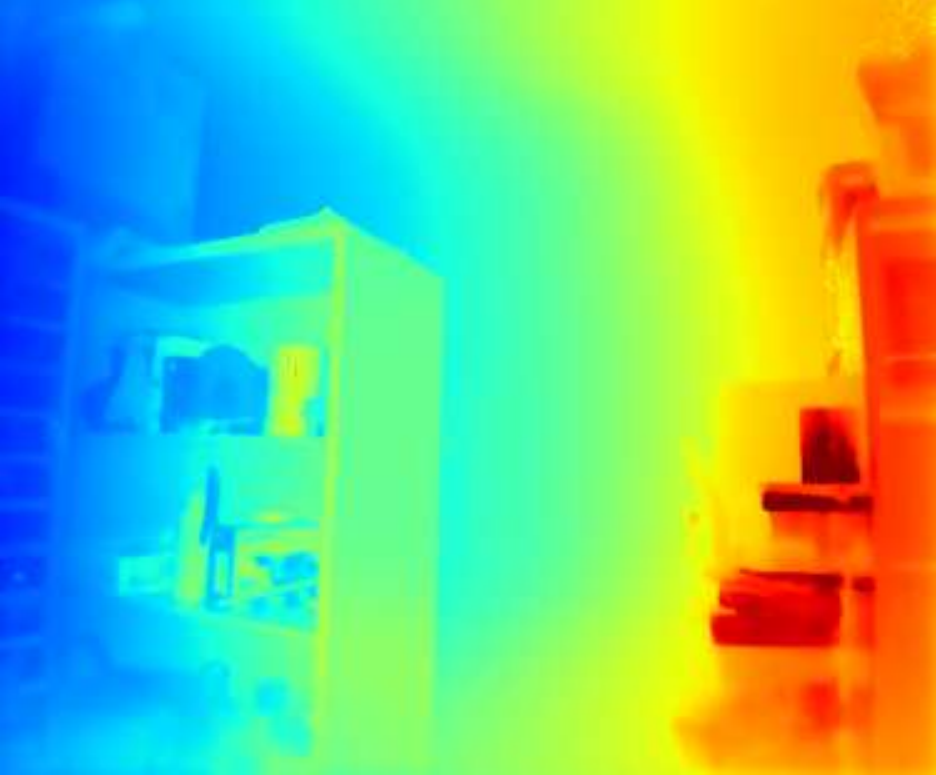}\quad
                          \includegraphics[scale=\figurescale]{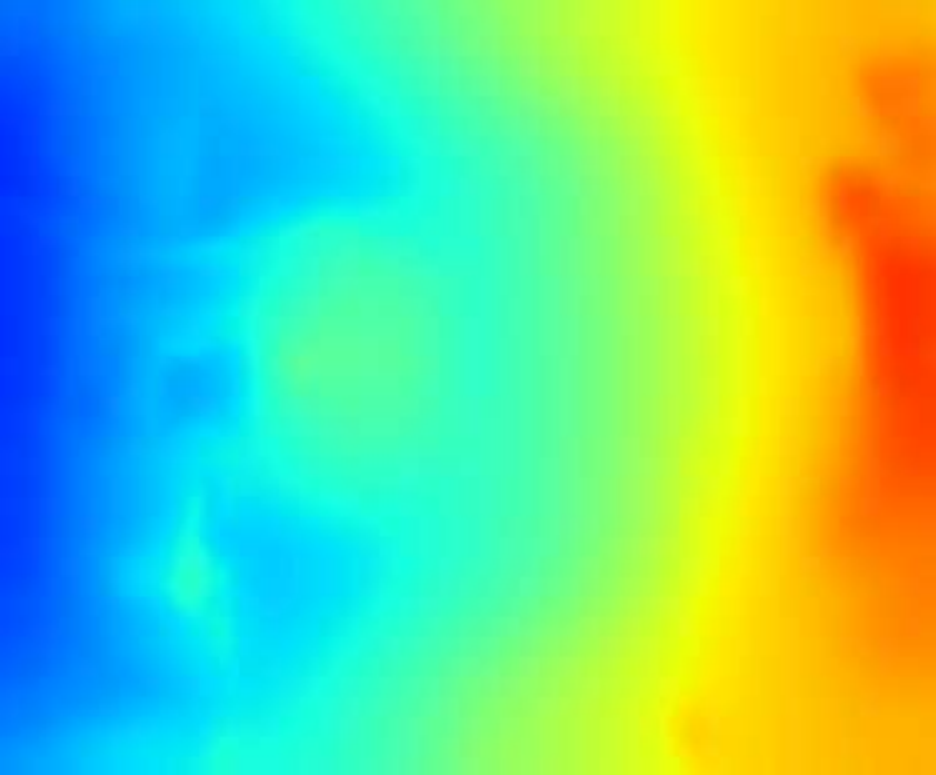}\quad
                          \includegraphics[scale=\figurescale]{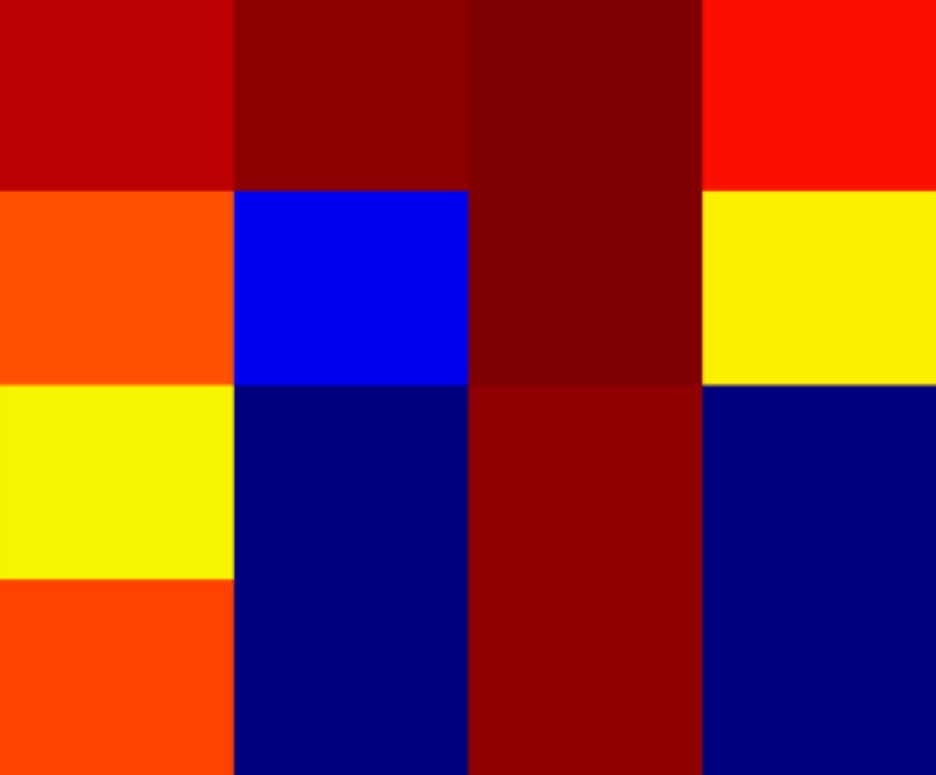}\quad
                          \includegraphics[scale=\figurescale]{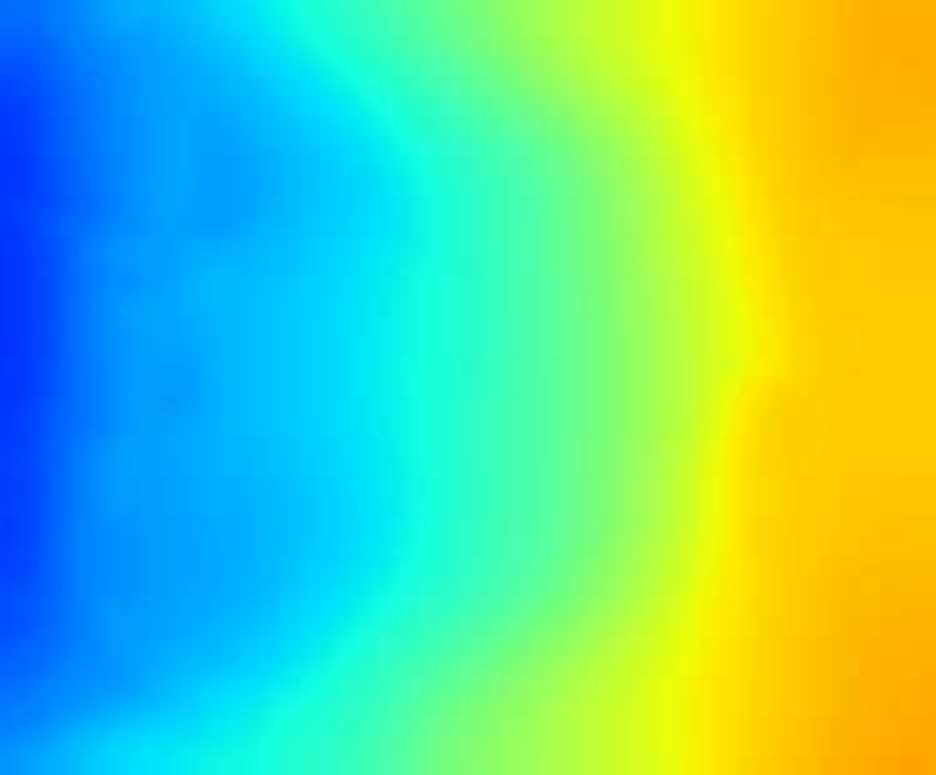}\quad
                          \includegraphics[scale=\figurescale]{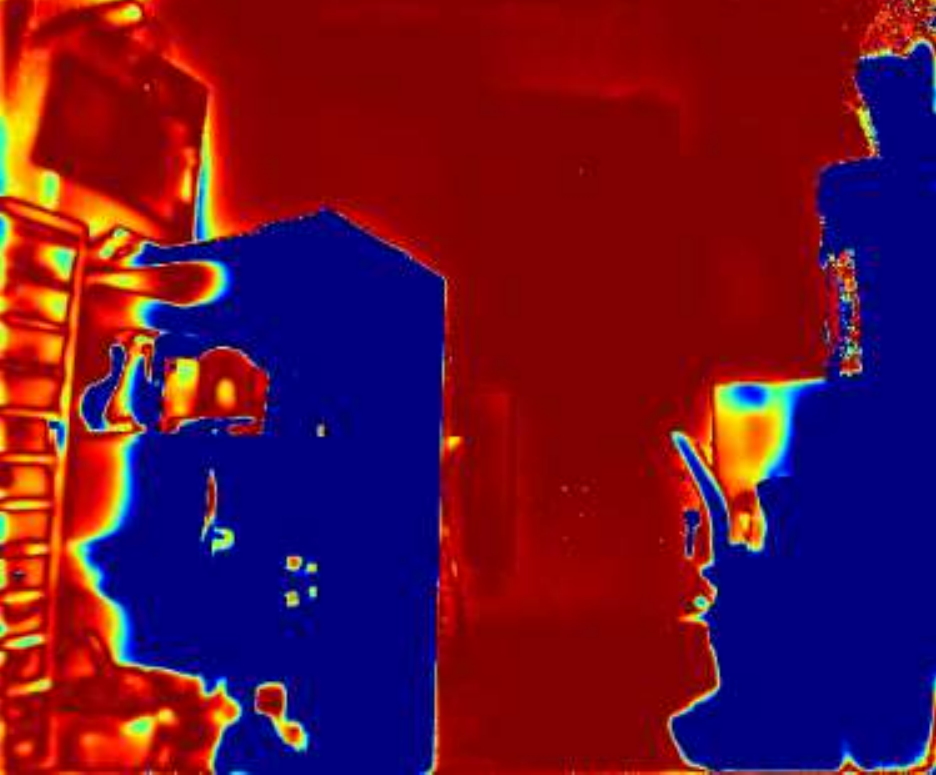}}\\[-0.1ex]
          \subfloat[Depth reconstruction]{\includegraphics[scale=\figurescale]{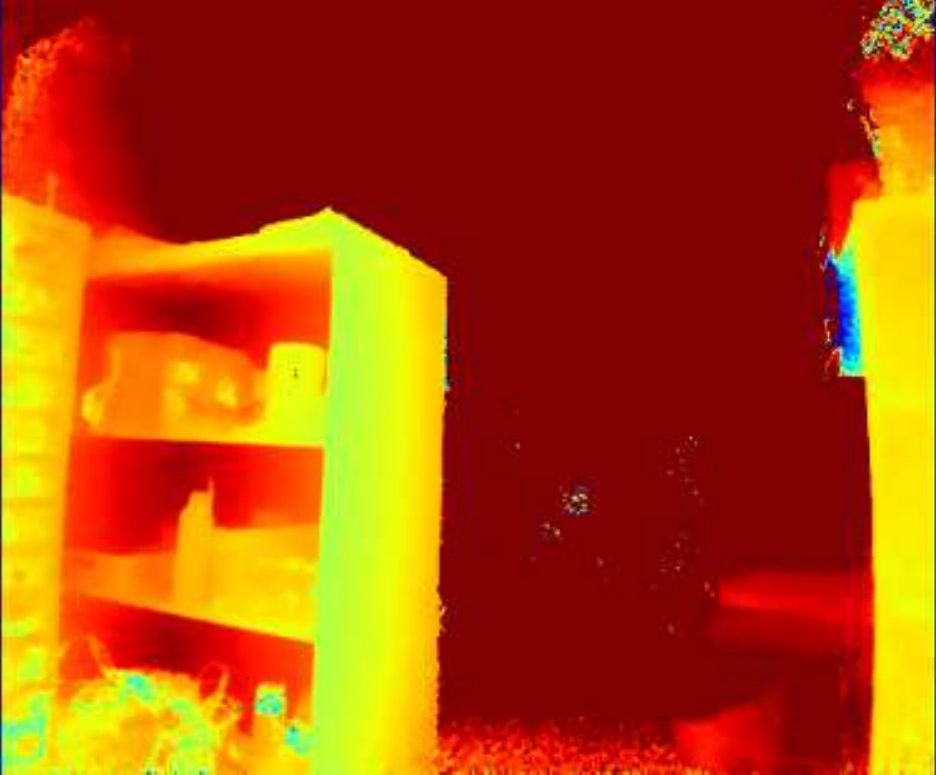}\quad
                          \includegraphics[scale=\figurescale]{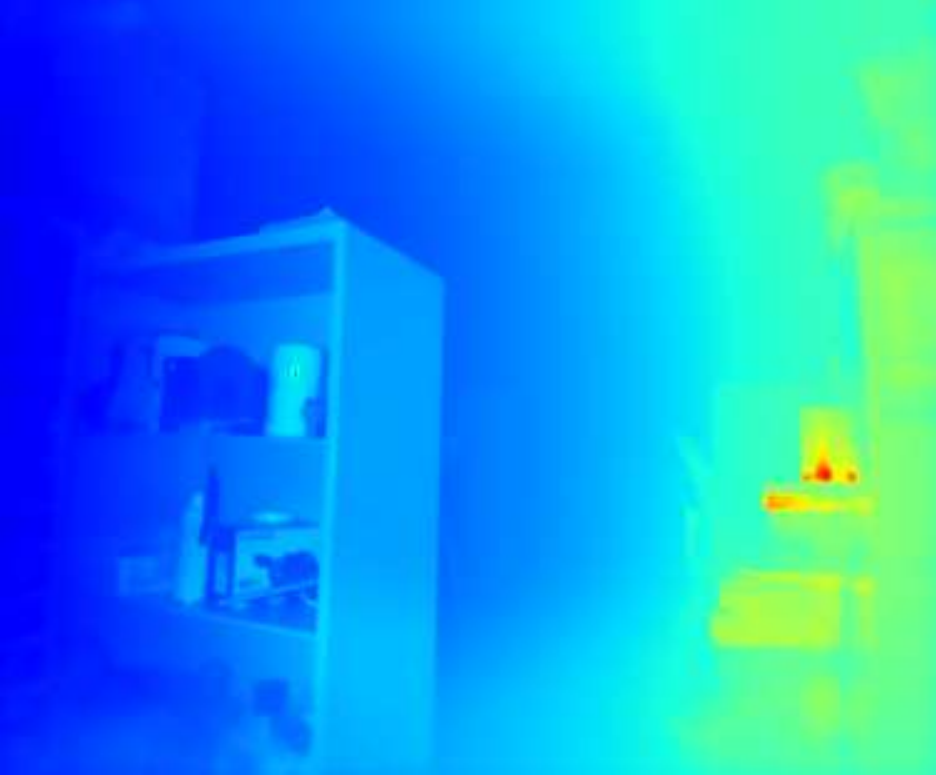}\quad
                          \includegraphics[scale=\figurescale]{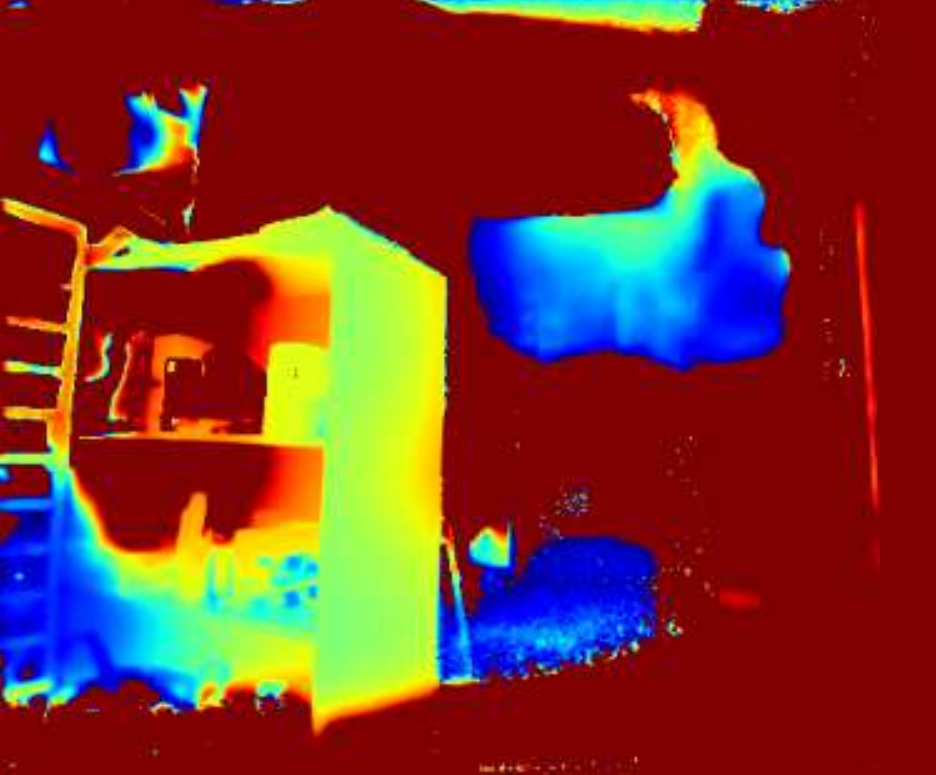}\quad
                          \includegraphics[scale=\figurescale]{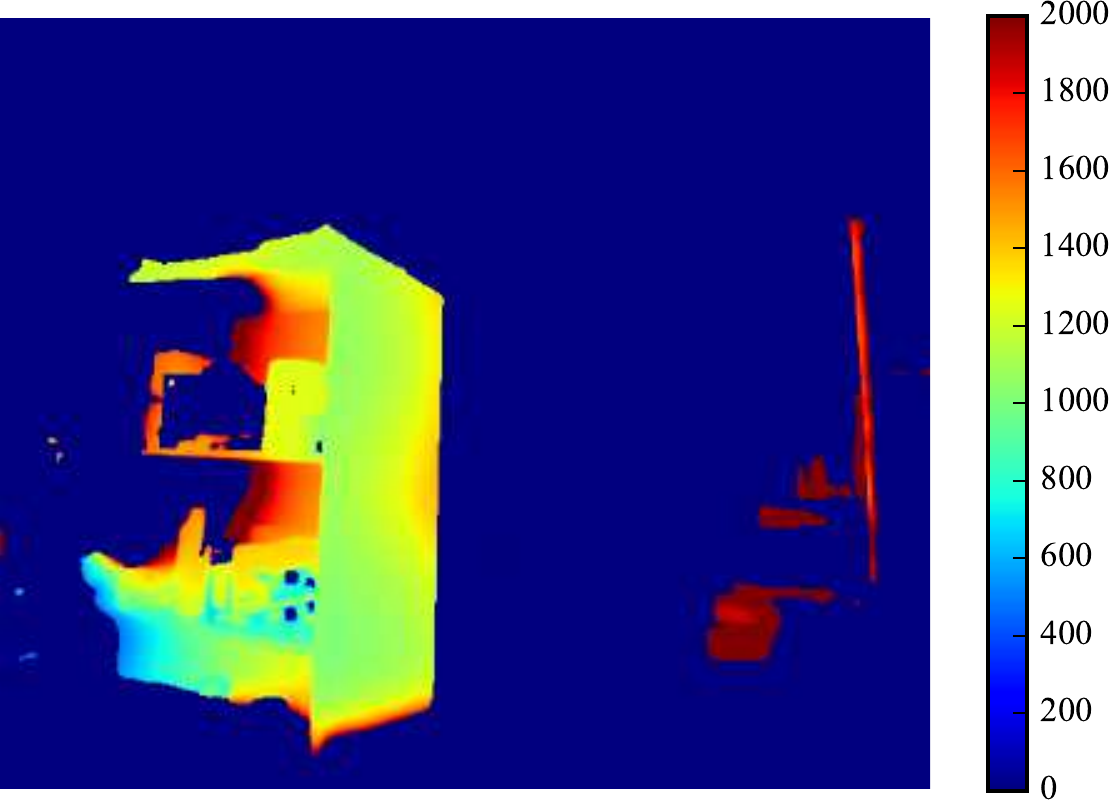}\quad
                          \includegraphics[scale=\figurescale]{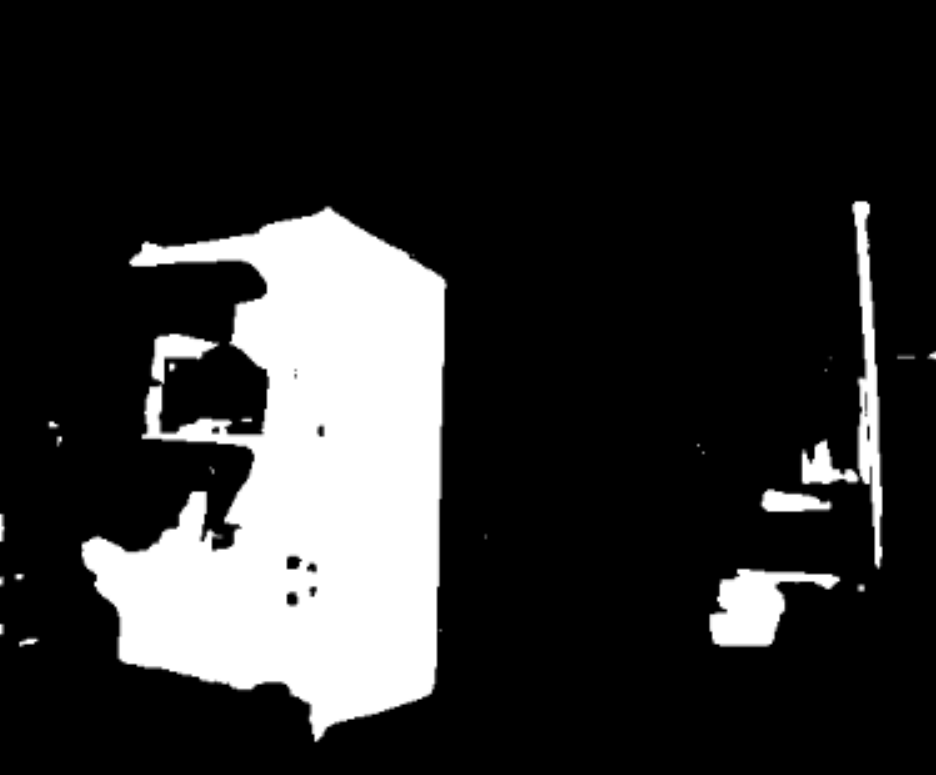}}\\[-0.1ex]
          \subfloat[Only fine optimization]{\includegraphics[scale=\figurescale]{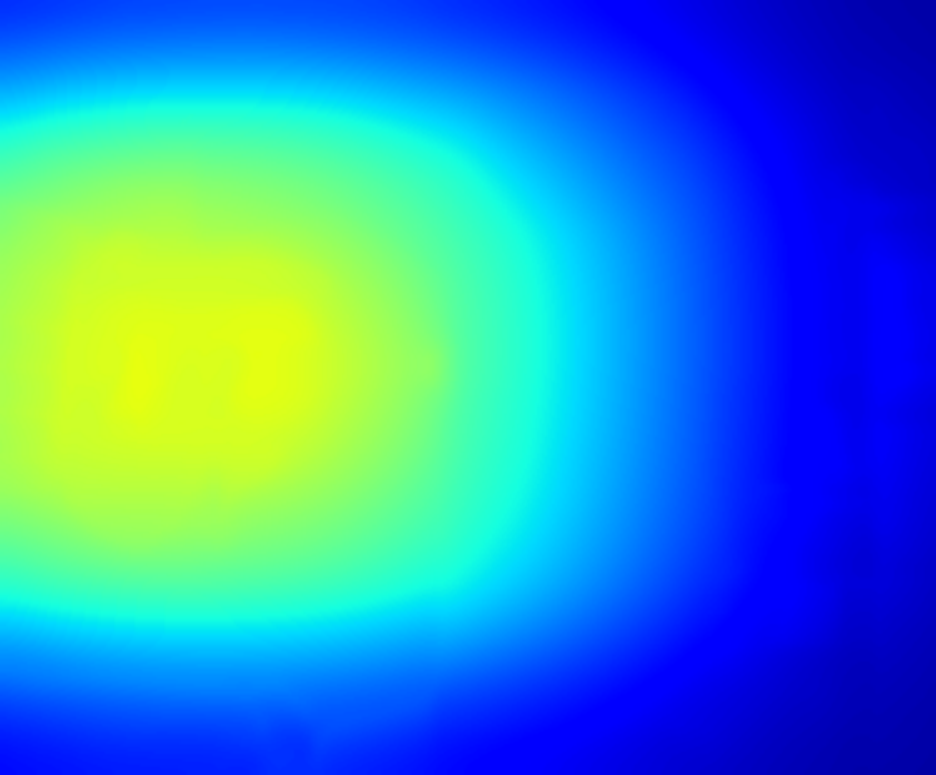}\quad
                          \includegraphics[scale=\figurescale]{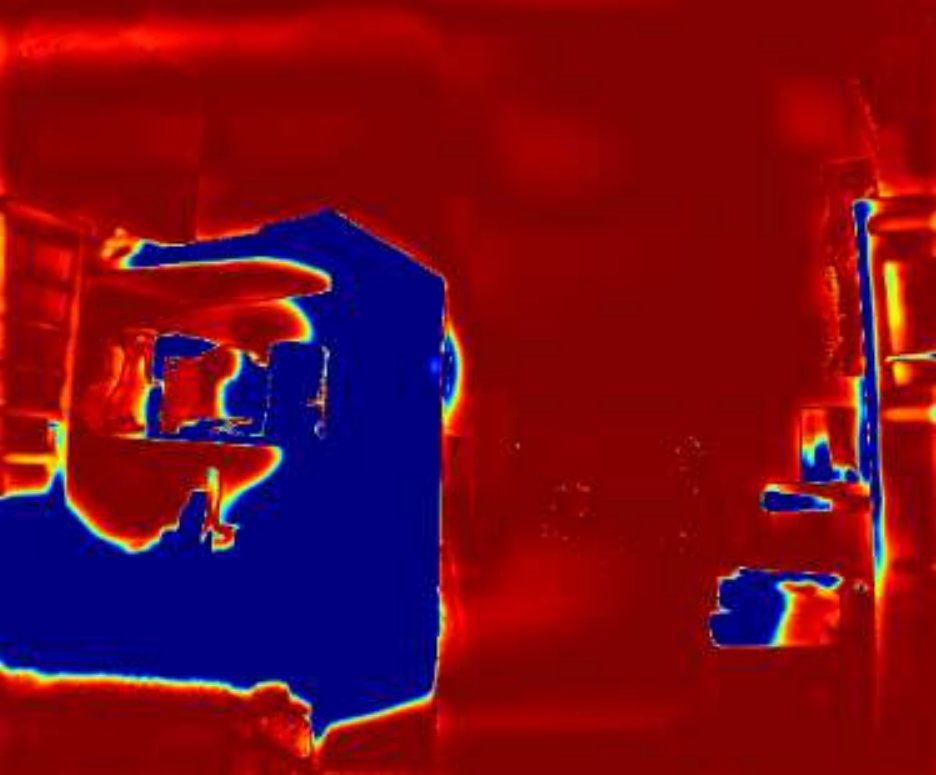}\quad
                          \includegraphics[scale=\figurescale]{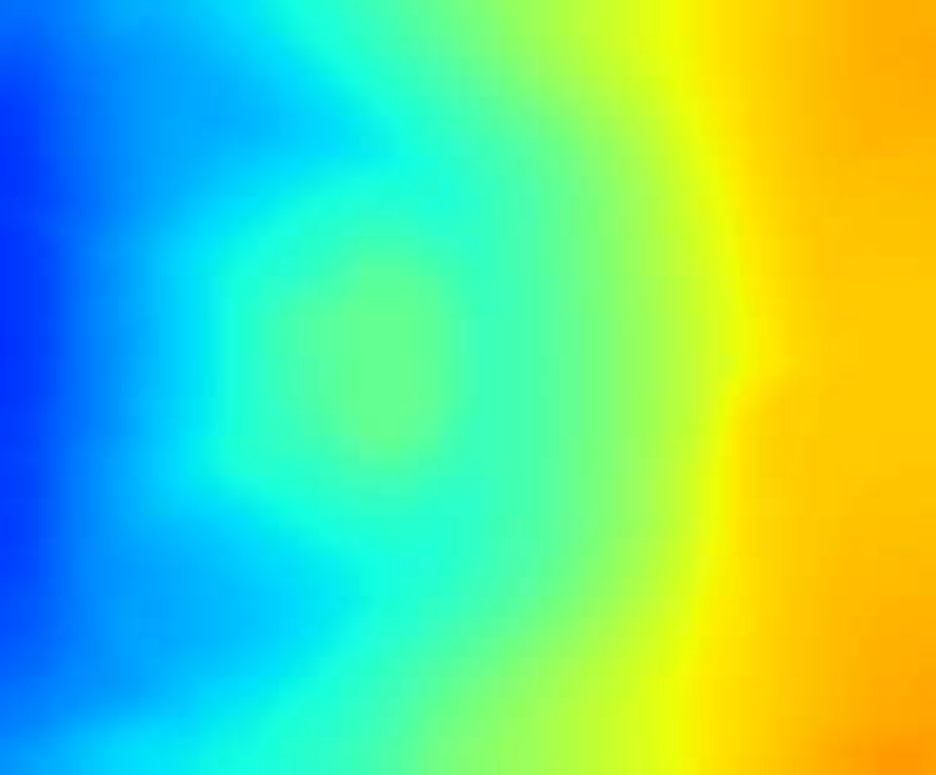}\quad
                          \includegraphics[scale=\figurescale]{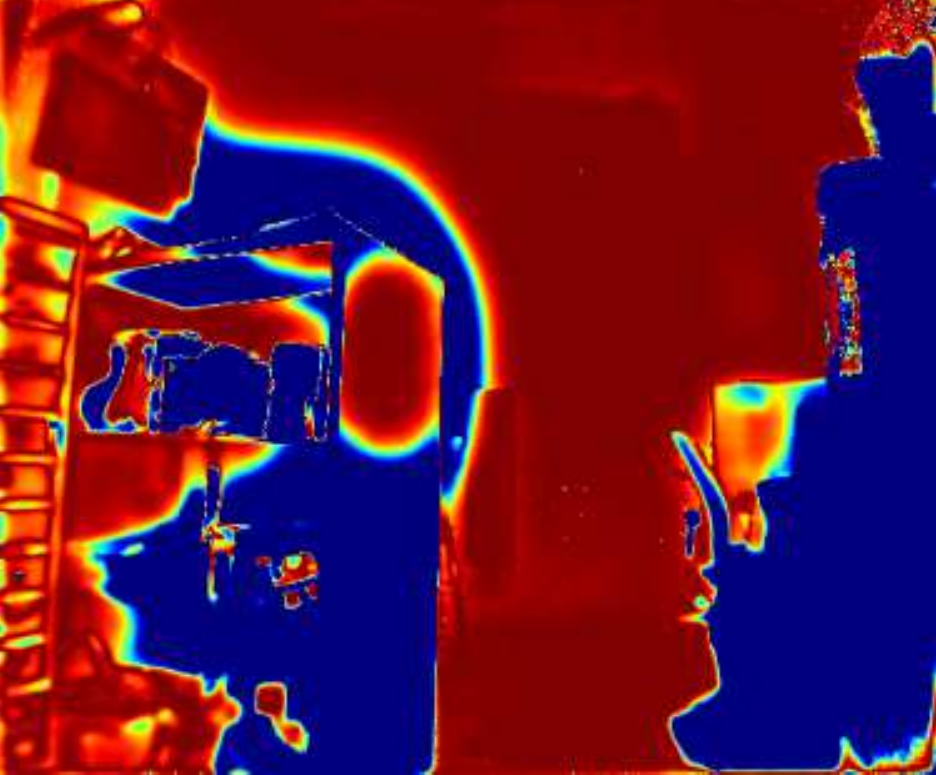}\quad
                          \includegraphics[scale=\figurescale]{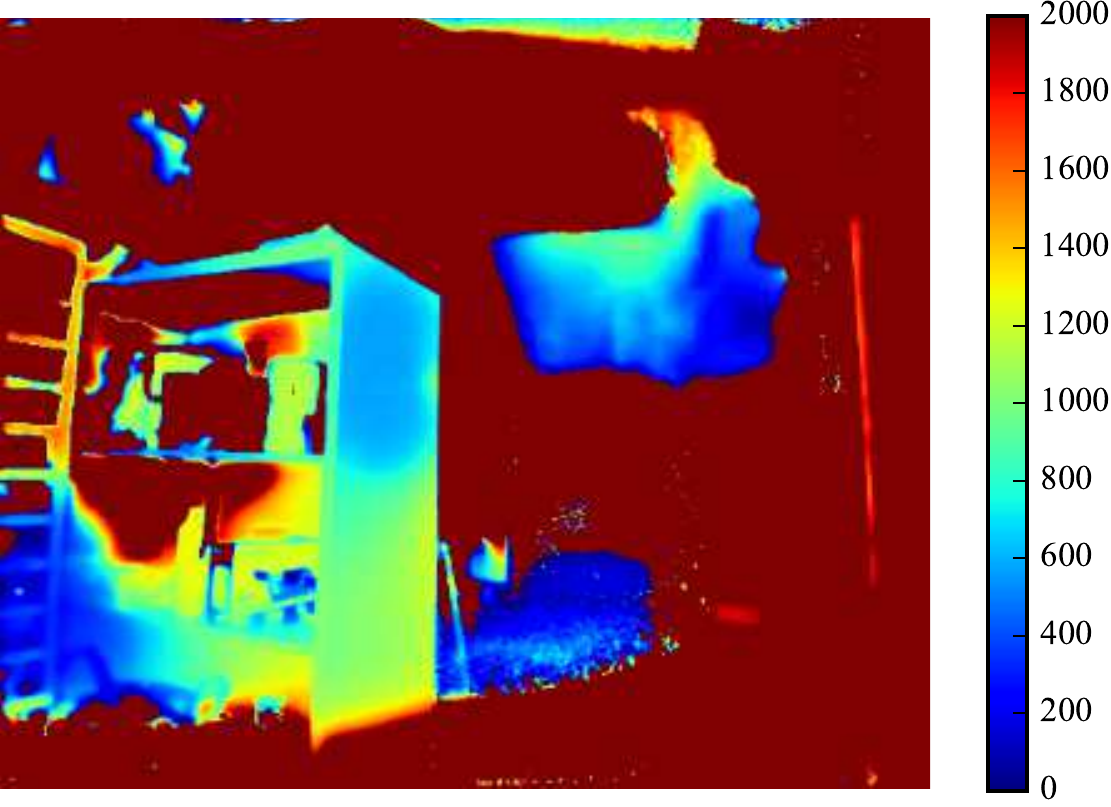}}
      \end{minipage}
    \end{tabular}
\caption{(a) Target scene. (b)(c) Left to right: input image, estimated scattering component and weight of coarse level, and of fine level for the amplitude and phase image, respectively. (d) Left to right: depth without fog, depth with fog, reconstructed depth, masked depth, and estimated object mask. (e) Result of applying only fine optimization. The estimated scattering component and the weight for the amplitude image, for the phase image, and the reconstructed depth from left to right.}
\label{fig:aisle_result}
\end{figure*}

\begin{figure*}[p]
\centering
    \begin{tabular}{c}
      \begin{minipage}{0.16\hsize}
        \centering
          \subfloat[Scene]{\includegraphics[scale=\figurescale]{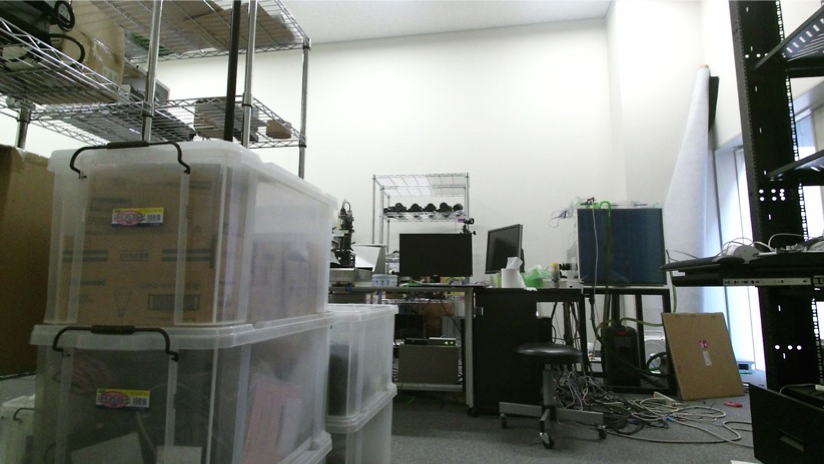}}
      \end{minipage}
      \begin{minipage}{0.8\hsize}
        \centering
          \subfloat[Amplitude]{\includegraphics[scale=\figurescale]{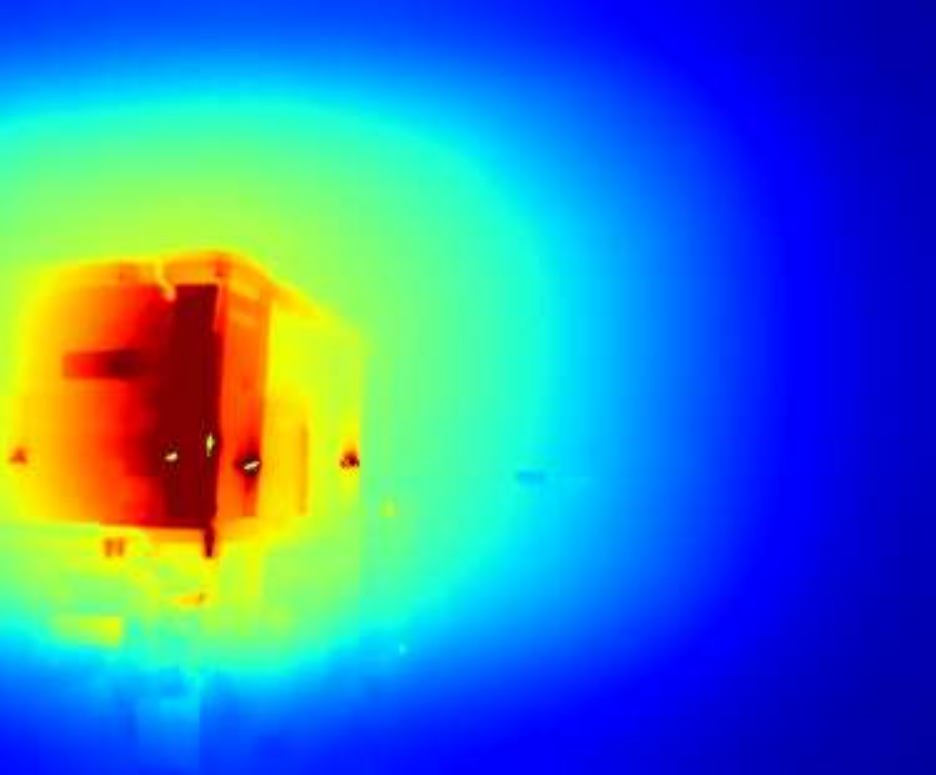}\quad
                          \includegraphics[scale=\figurescale]{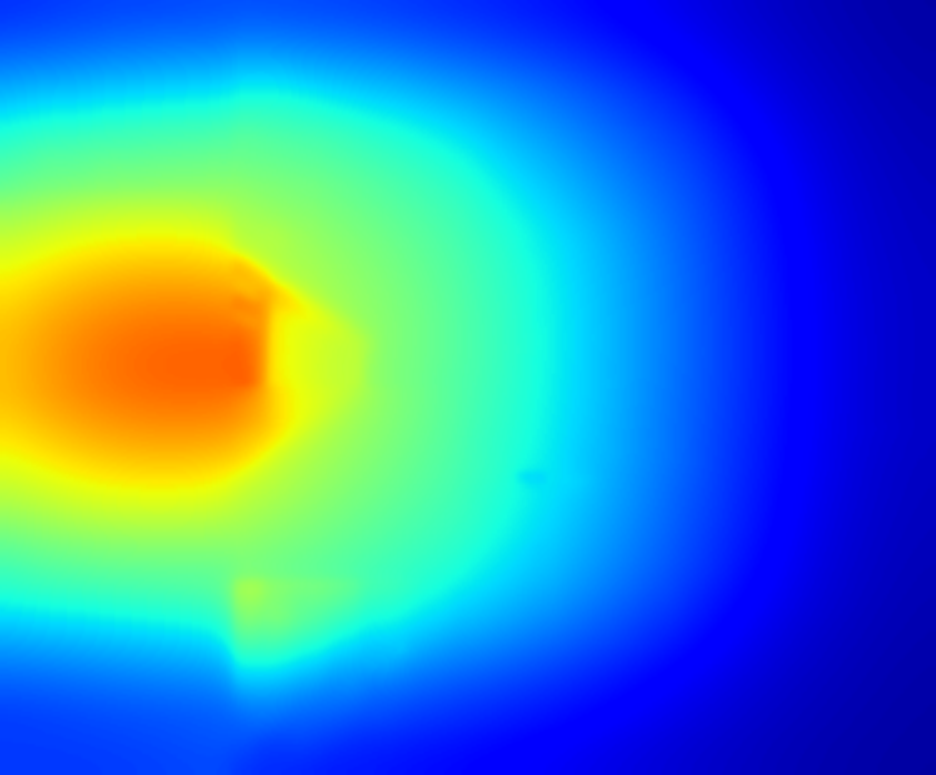}\quad
                          \includegraphics[scale=\figurescale]{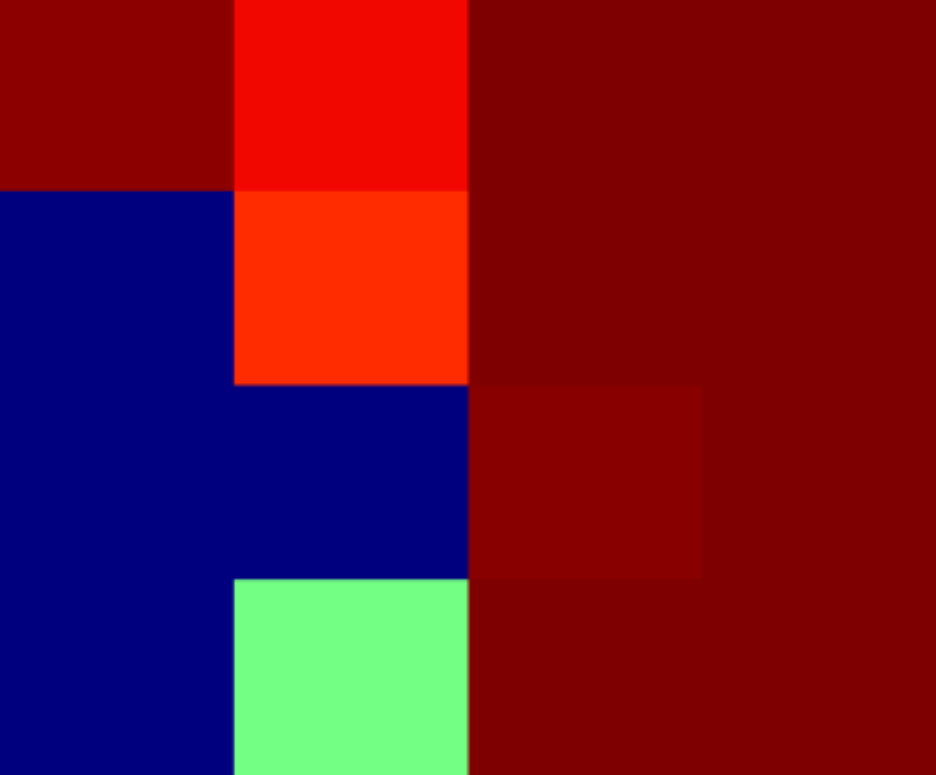}\quad
                          \includegraphics[scale=\figurescale]{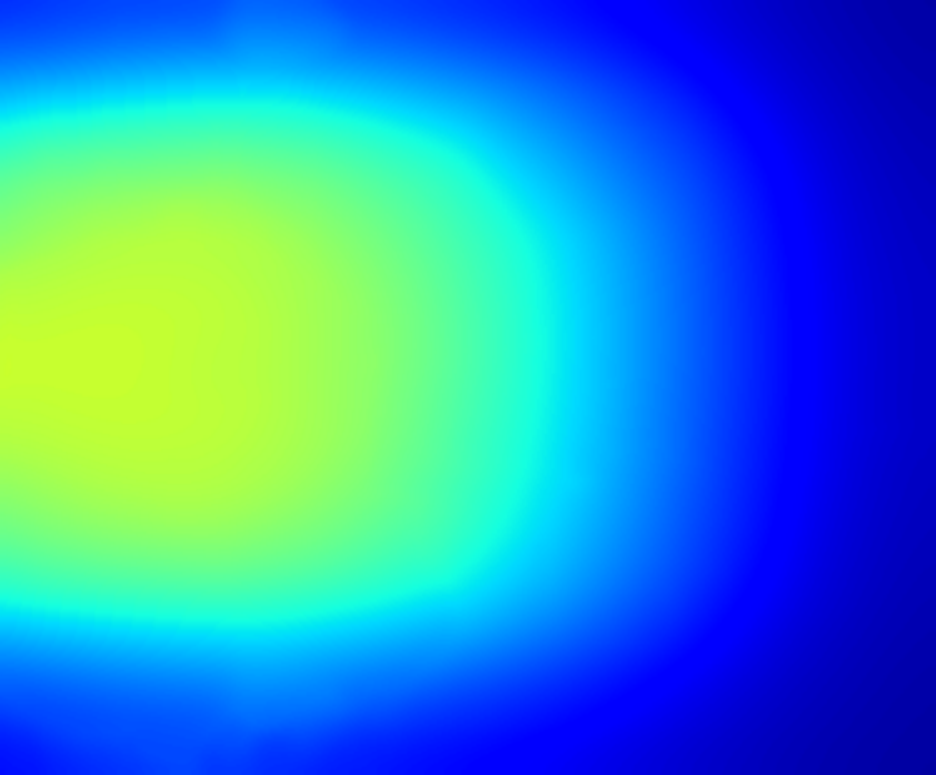}\quad
                          \includegraphics[scale=\figurescale]{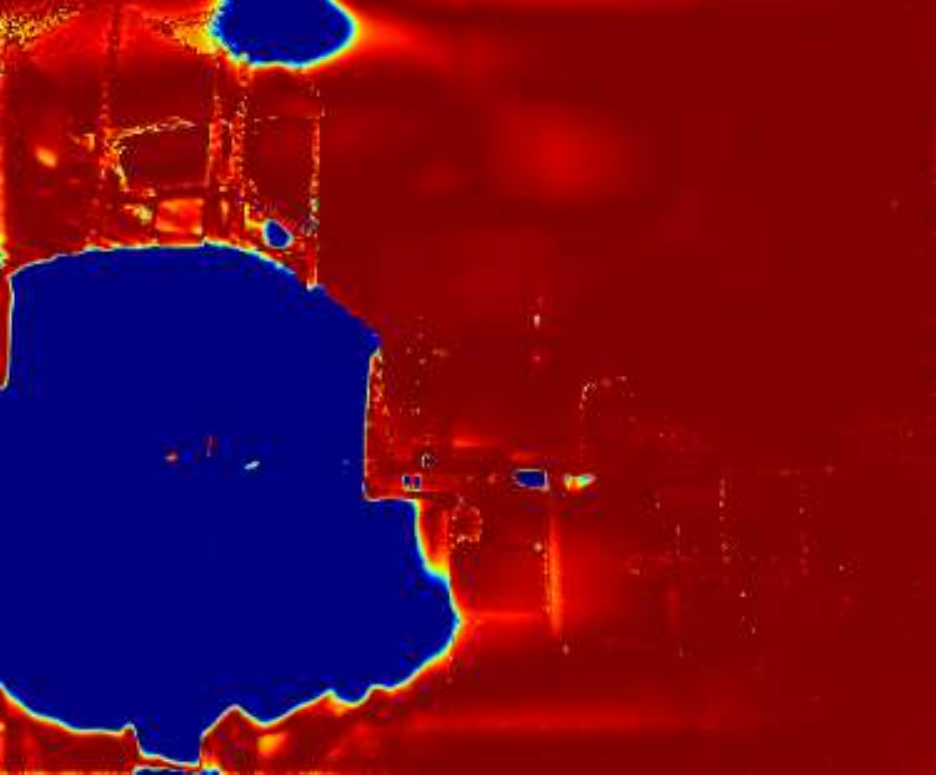}}\\[-0.1ex]
          \subfloat[Phase]{\includegraphics[scale=\figurescale]{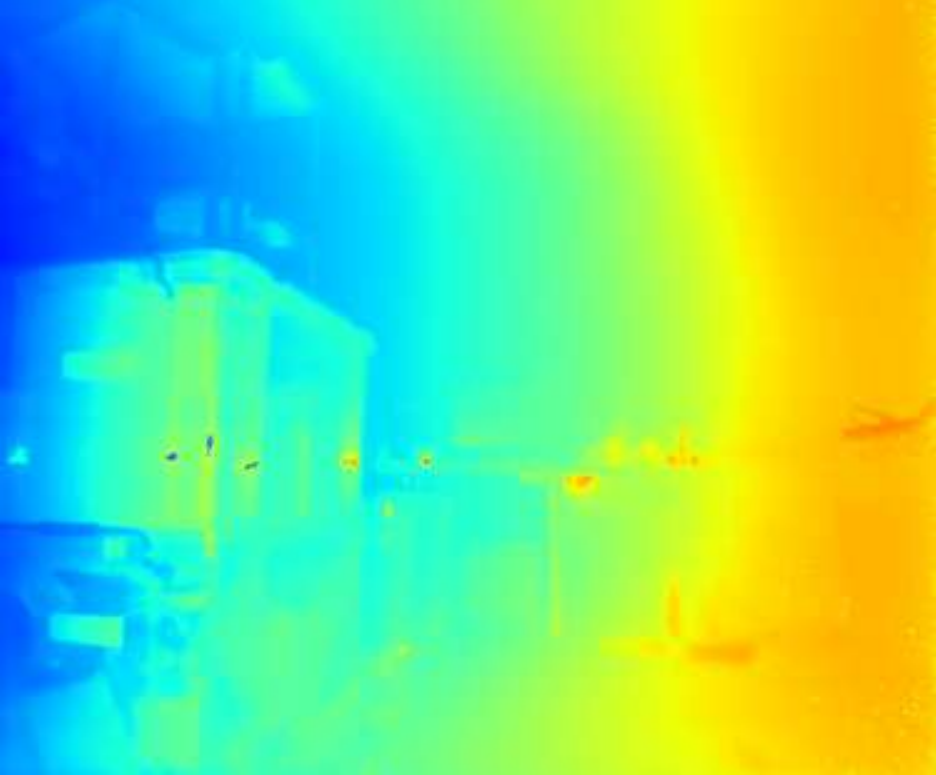}\quad
                          \includegraphics[scale=\figurescale]{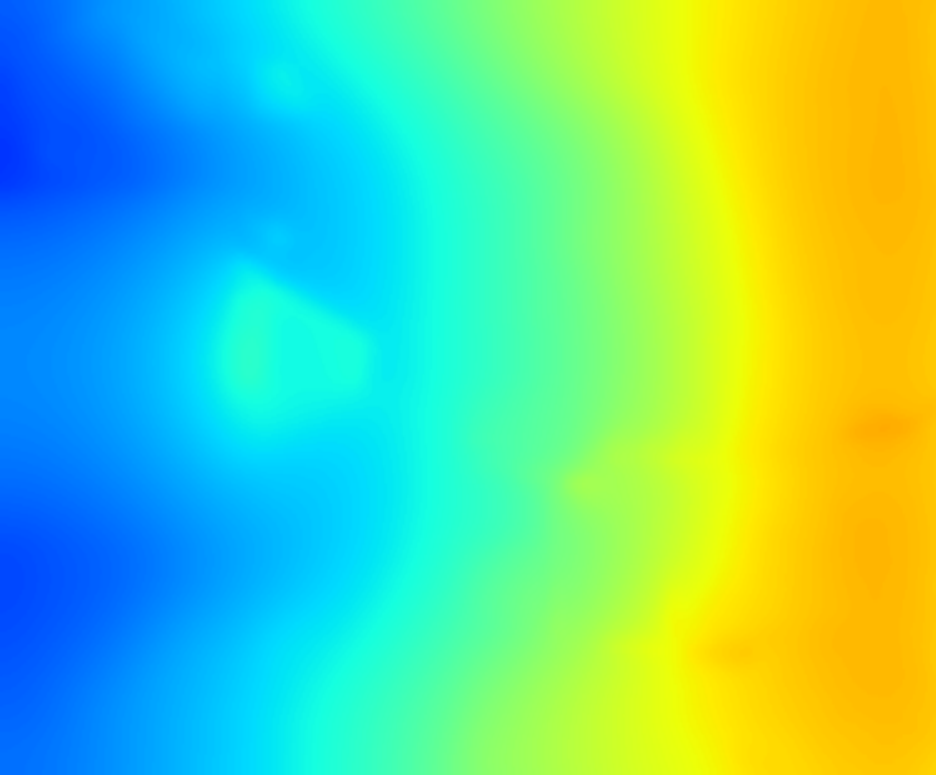}\quad
                          \includegraphics[scale=\figurescale]{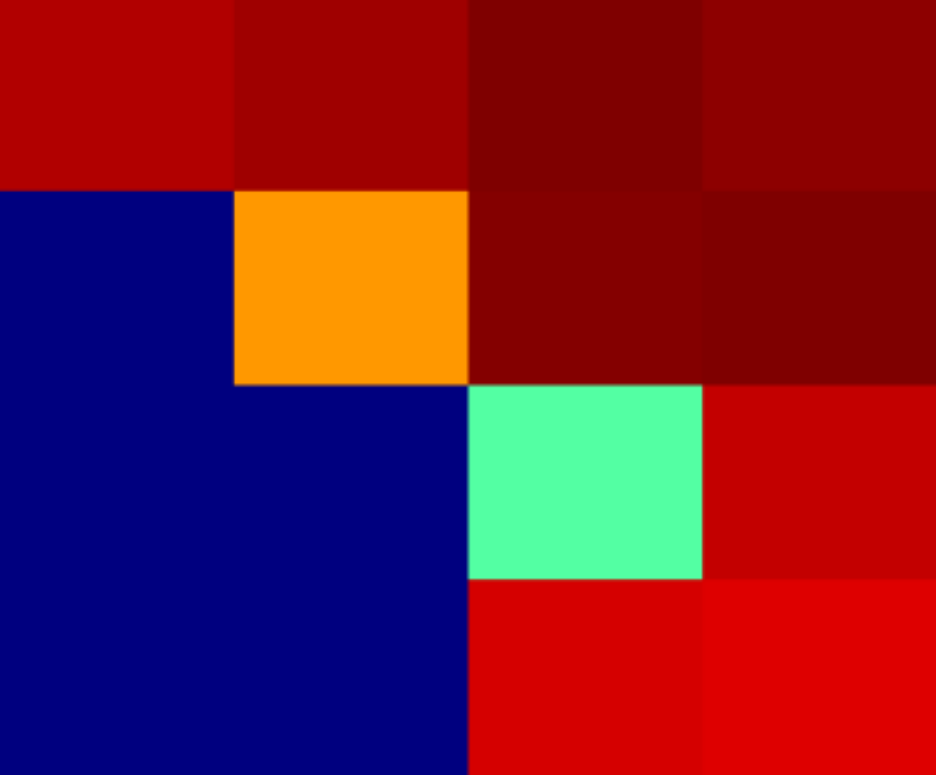}\quad
                          \includegraphics[scale=\figurescale]{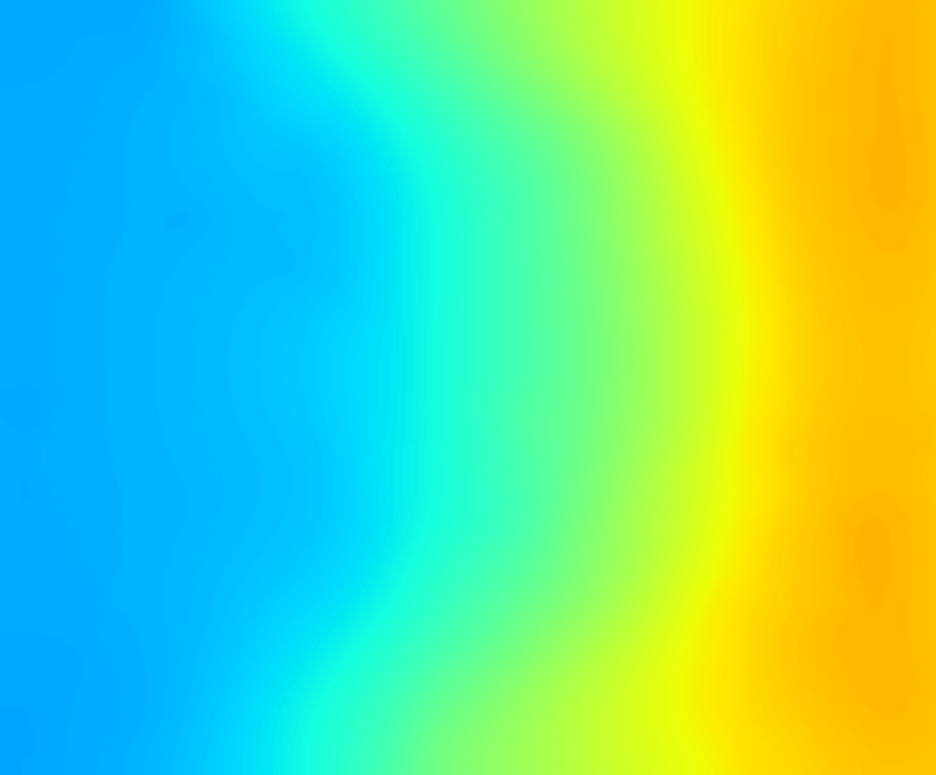}\quad
                          \includegraphics[scale=\figurescale]{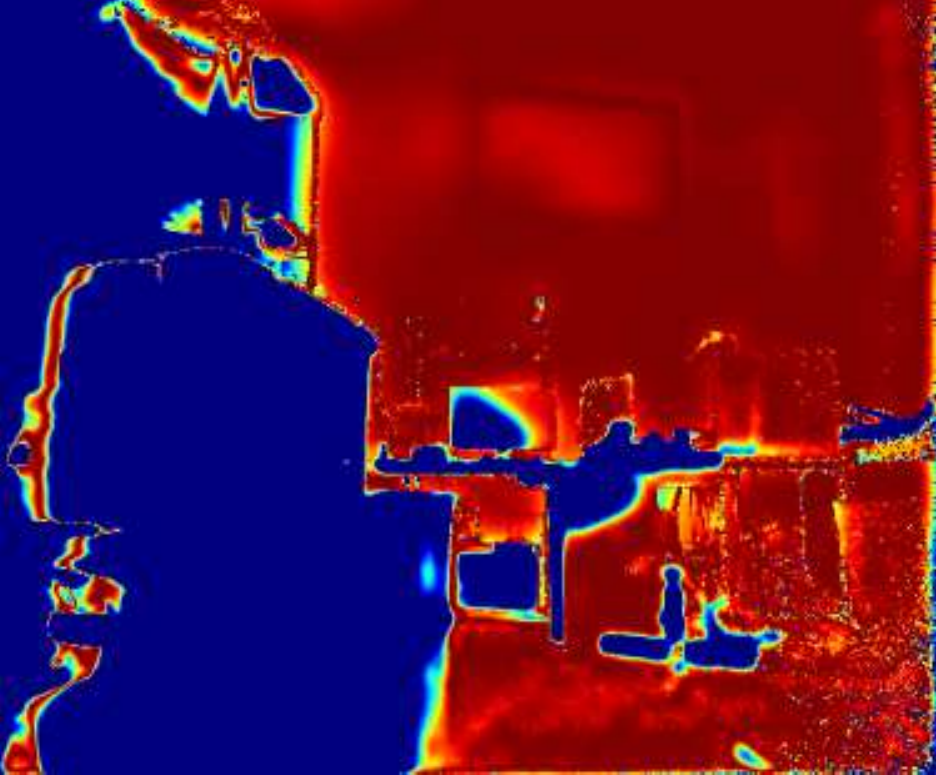}}\\[-0.1ex]
          \subfloat[Depth reconstruction]{\includegraphics[scale=\figurescale]{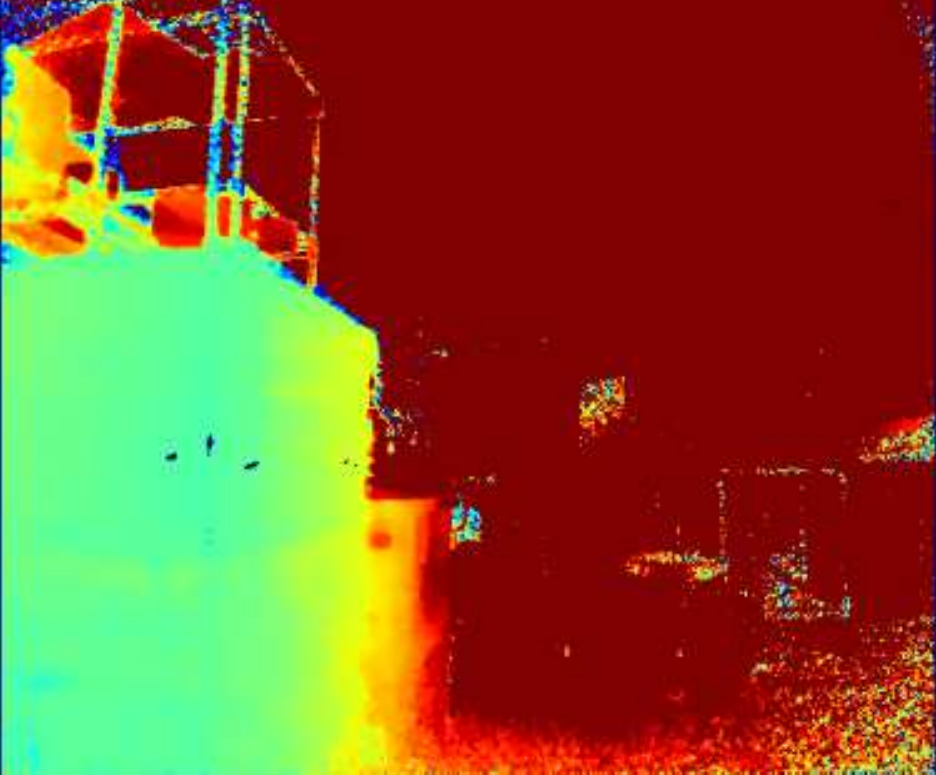}\quad
                          \includegraphics[scale=\figurescale]{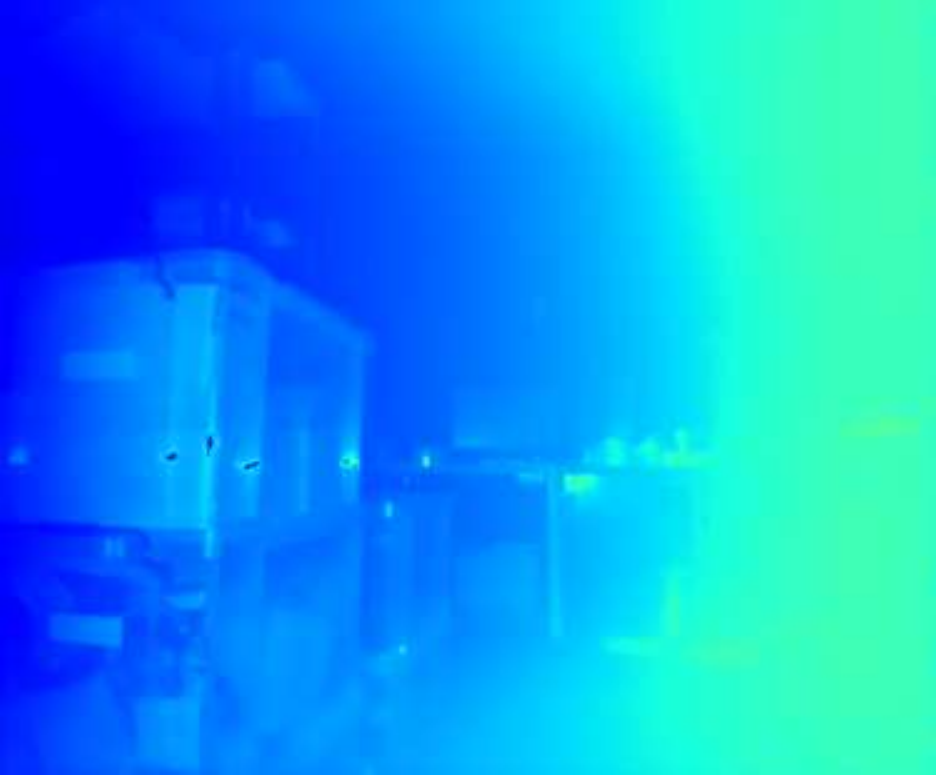}\quad
                          \includegraphics[scale=\figurescale]{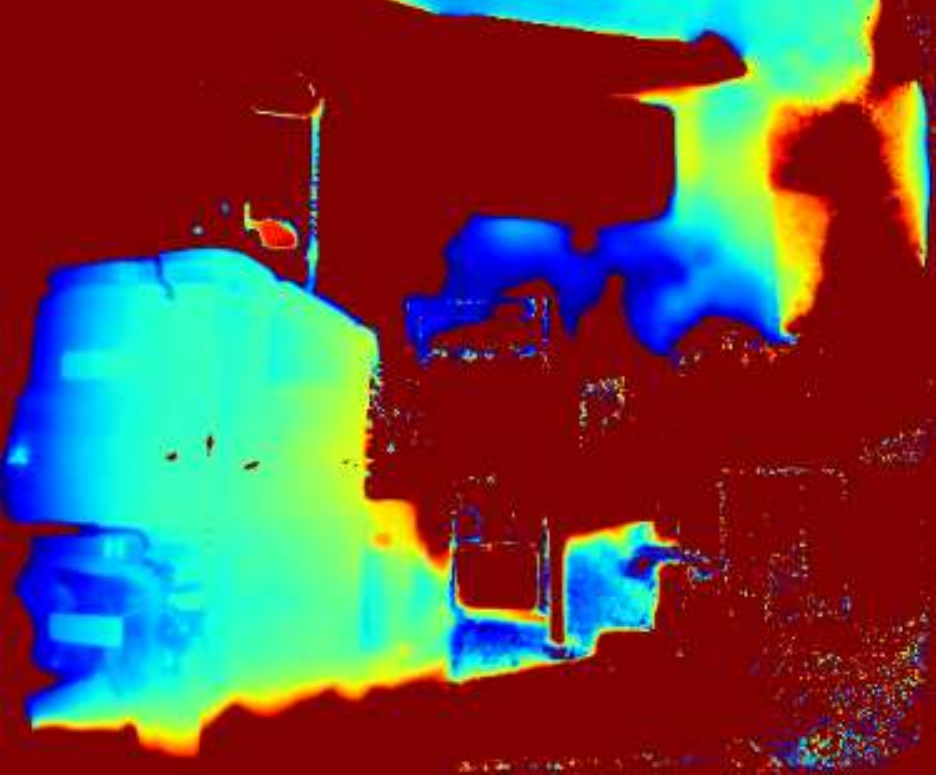}\quad
                          \includegraphics[scale=\figurescale]{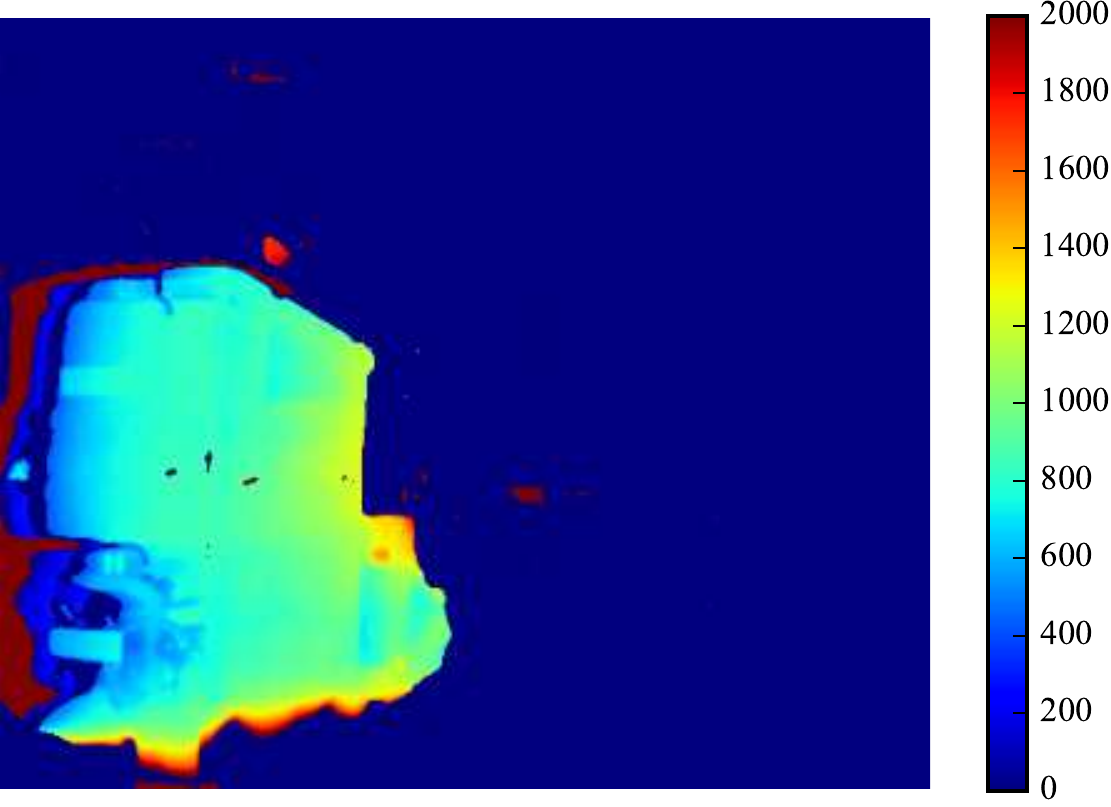}\quad
                          \includegraphics[scale=\figurescale]{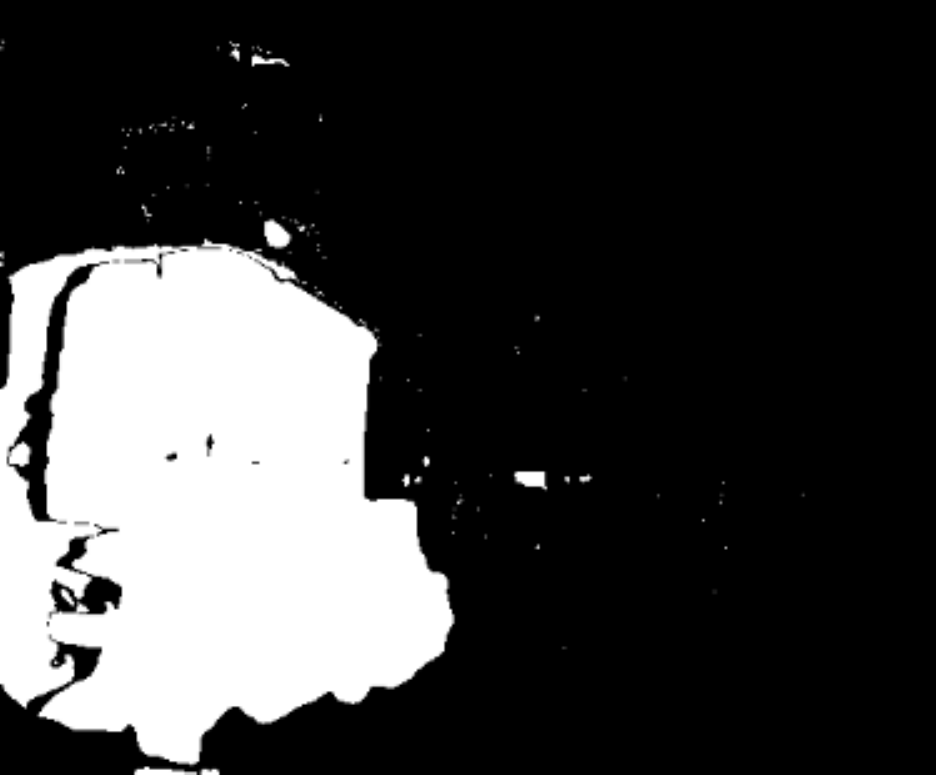}}\\[-0.1ex]
          \subfloat[Only fine optimization]{\includegraphics[scale=\figurescale]{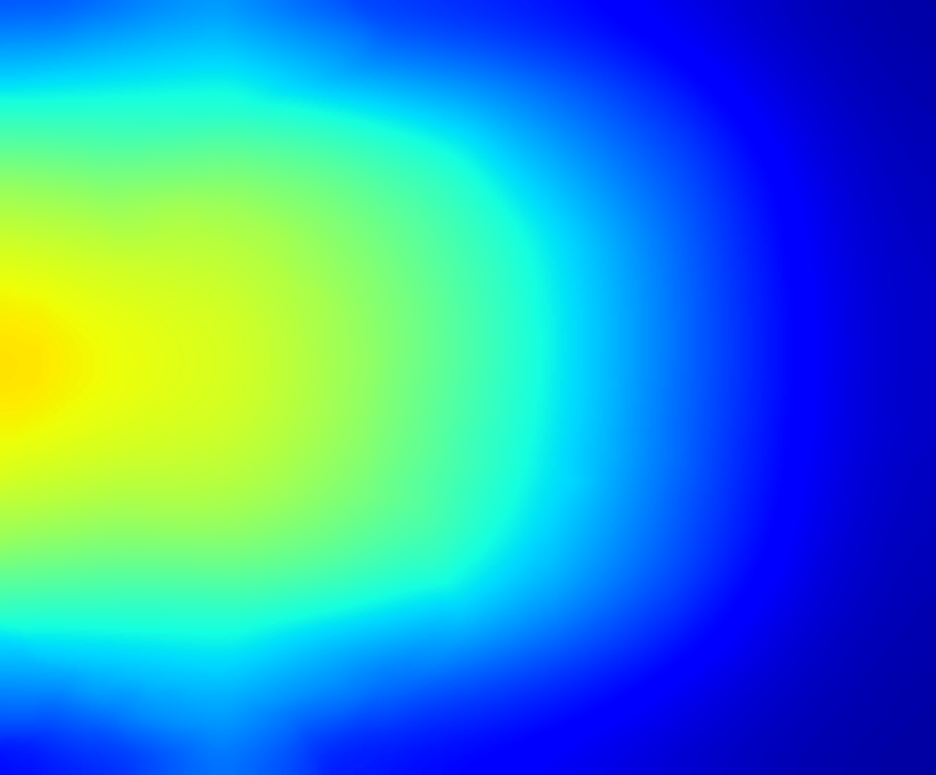}\quad
                          \includegraphics[scale=\figurescale]{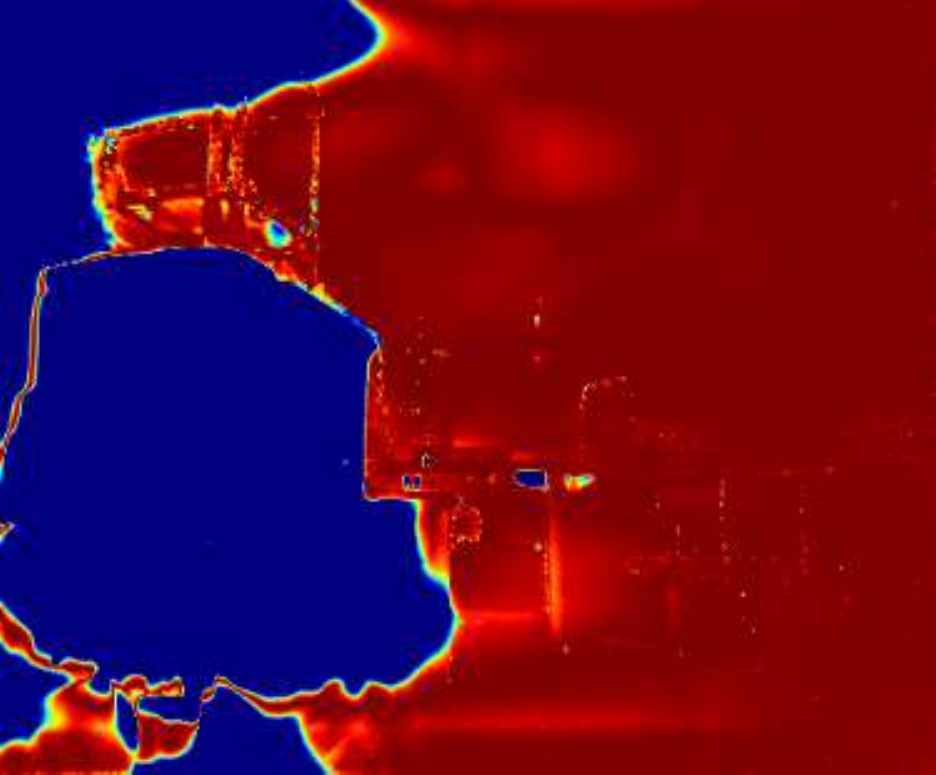}\quad
                          \includegraphics[scale=\figurescale]{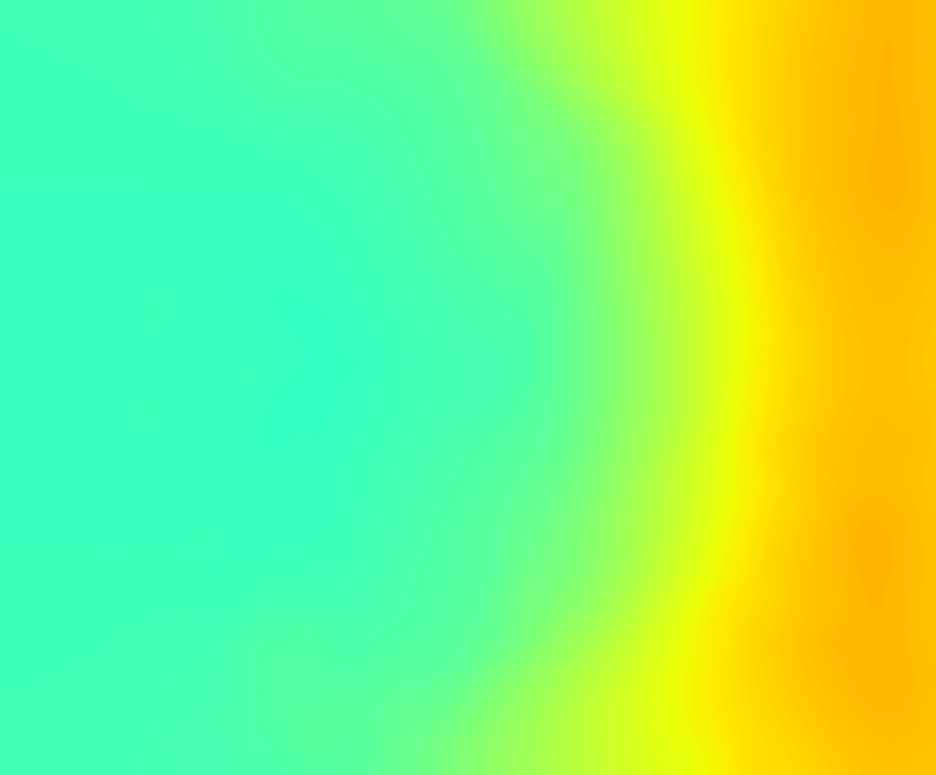}\quad
                          \includegraphics[scale=\figurescale]{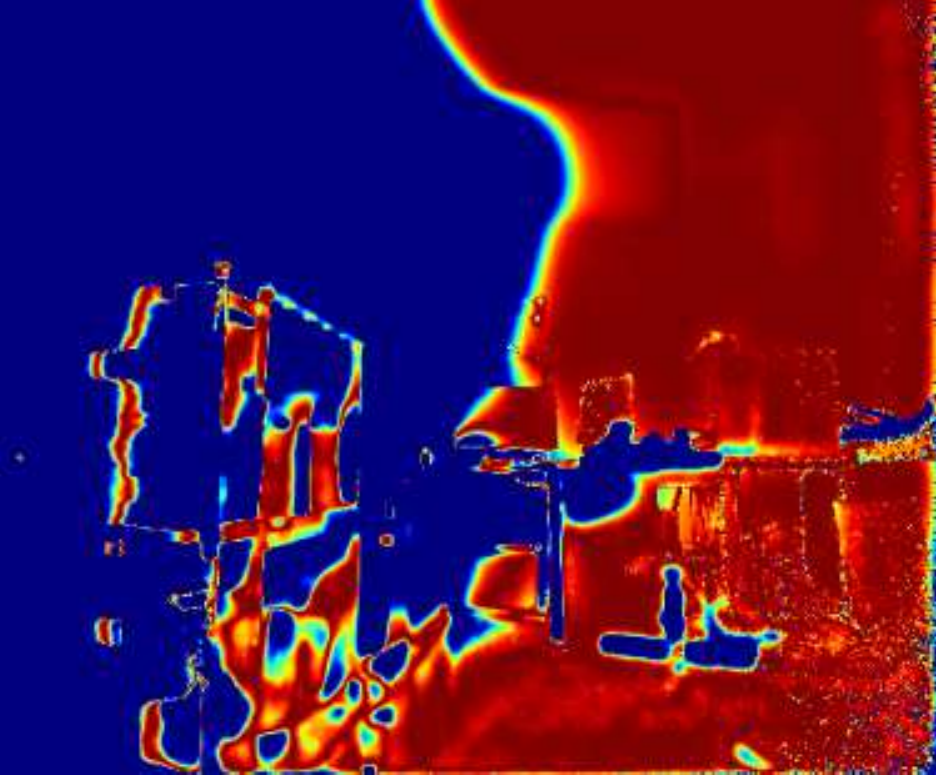}\quad
                          \includegraphics[scale=\figurescale]{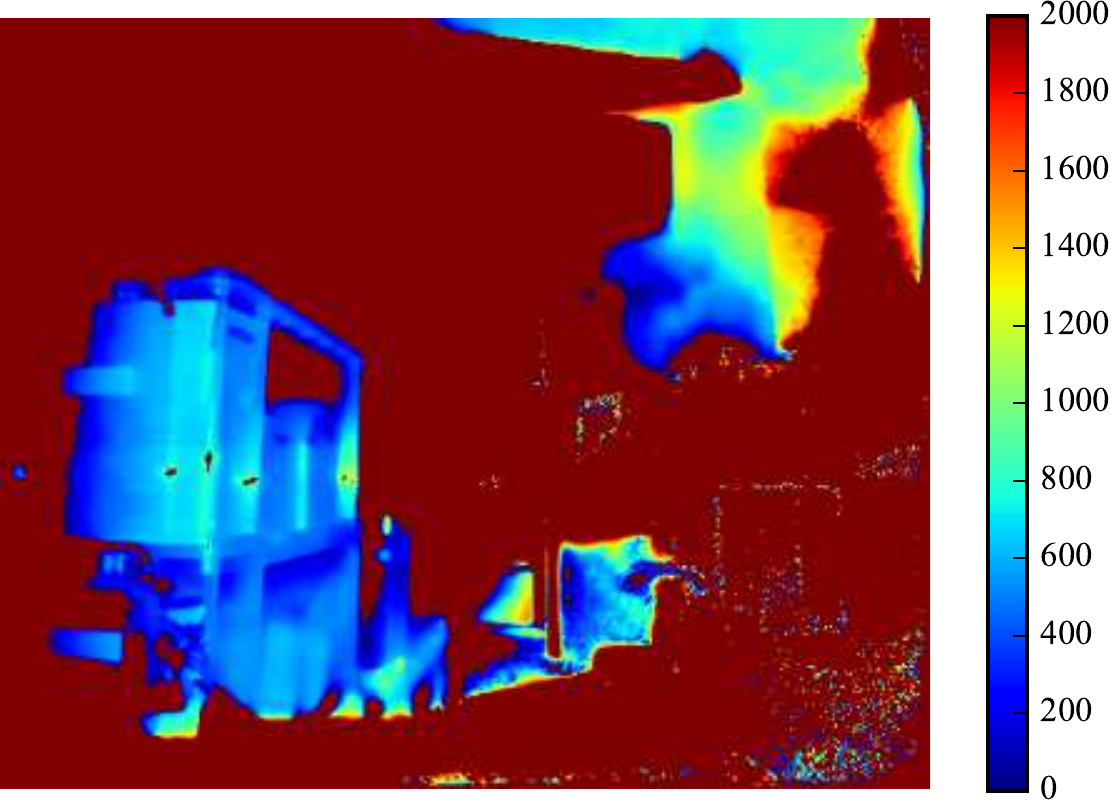}}
      \end{minipage}
    \end{tabular}
\caption{(a) Target scene. (b)(c) Left to right: input image, estimated scattering component and weight of coarse level, and of fine level for the amplitude and phase image, respectively. (d) Left to right: depth without fog, depth with fog, reconstructed depth, masked depth, and estimated object mask. (e) Result of applying only fine optimization. The estimated scattering component and the weight for the amplitude image, for the phase image, and the reconstructed depth from left to right.}
\label{fig:box_result}
\bigskip
	\begin{tabular}{c}
		\begin{minipage}{0.3\hsize}
			\centering
				\subfloat[Scene]{\includegraphics[scale=\figurescale]{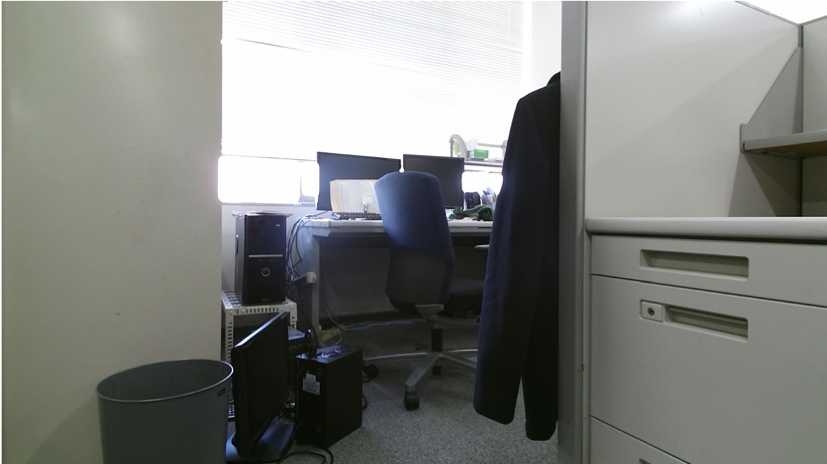}}
		\end{minipage}
		\begin{minipage}{0.6\hsize}
			\centering
				\subfloat[Amplitude]{\includegraphics[scale=\figurescaleR]{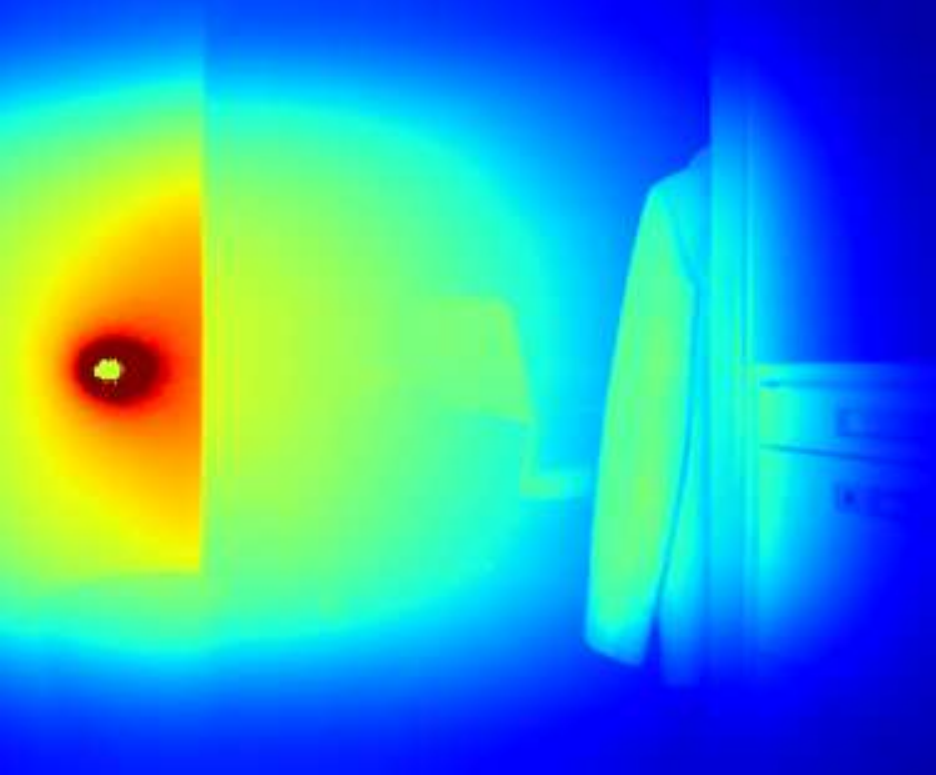}\quad
                          \includegraphics[scale=\figurescaleR]{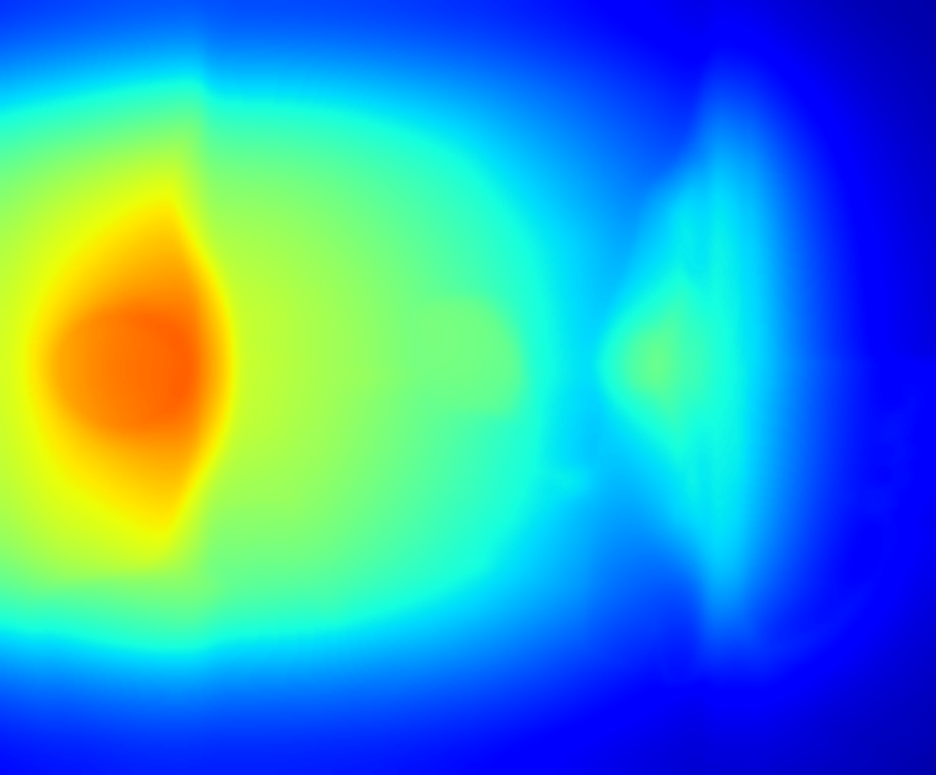}\quad
                          \includegraphics[scale=\figurescaleR]{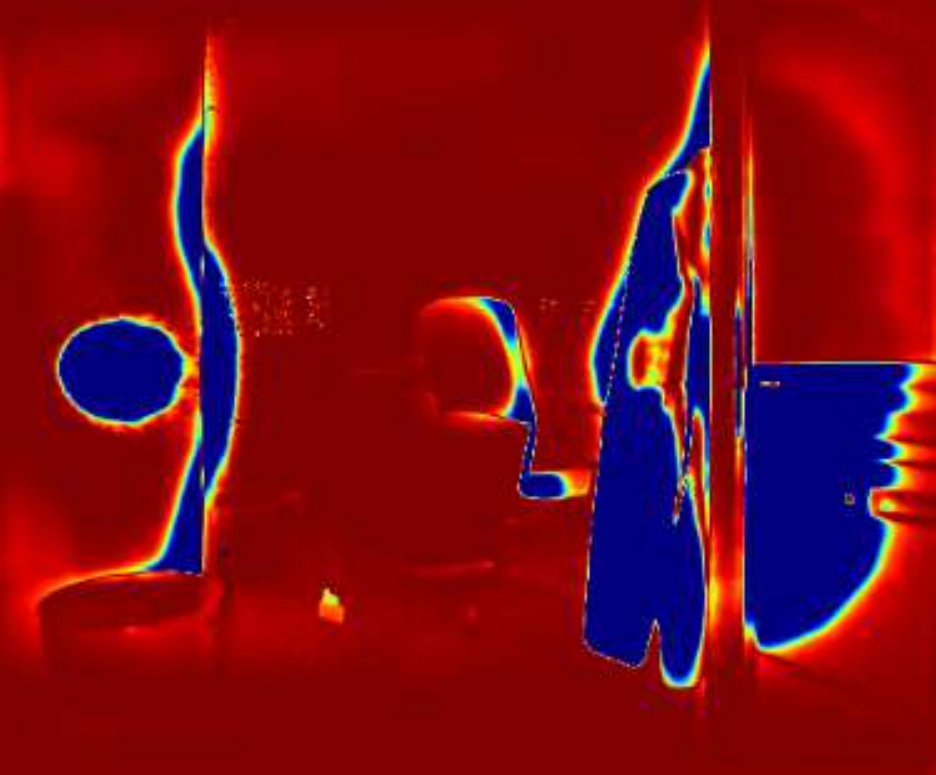}}\\
    			\subfloat[Phase]{\includegraphics[scale=\figurescaleR]{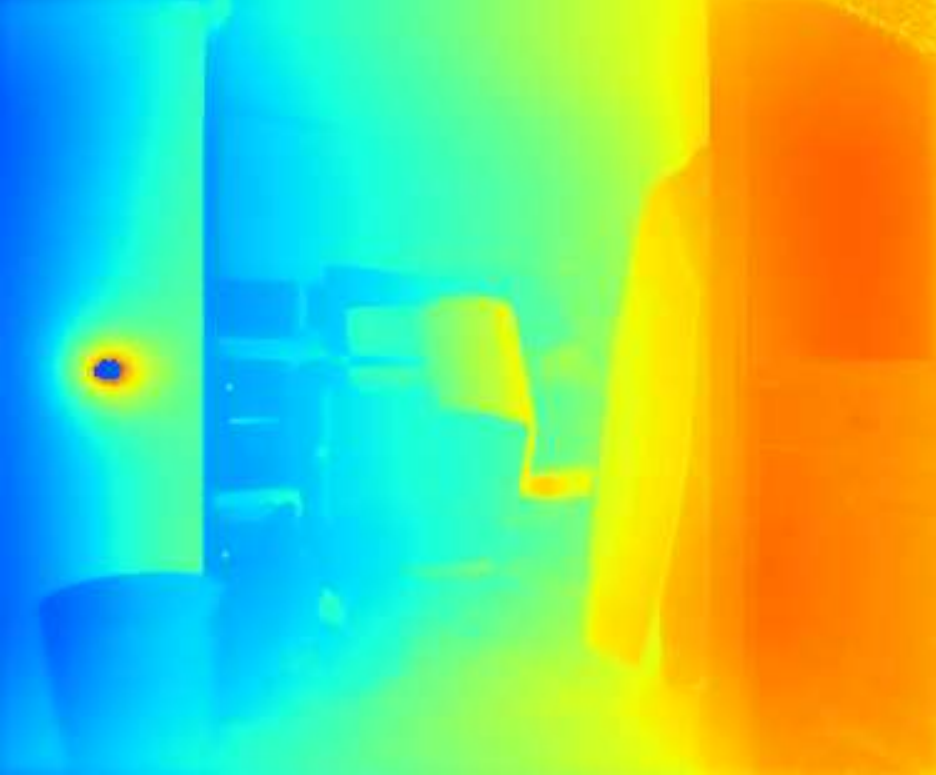}\quad
                         \includegraphics[scale=\figurescaleR]{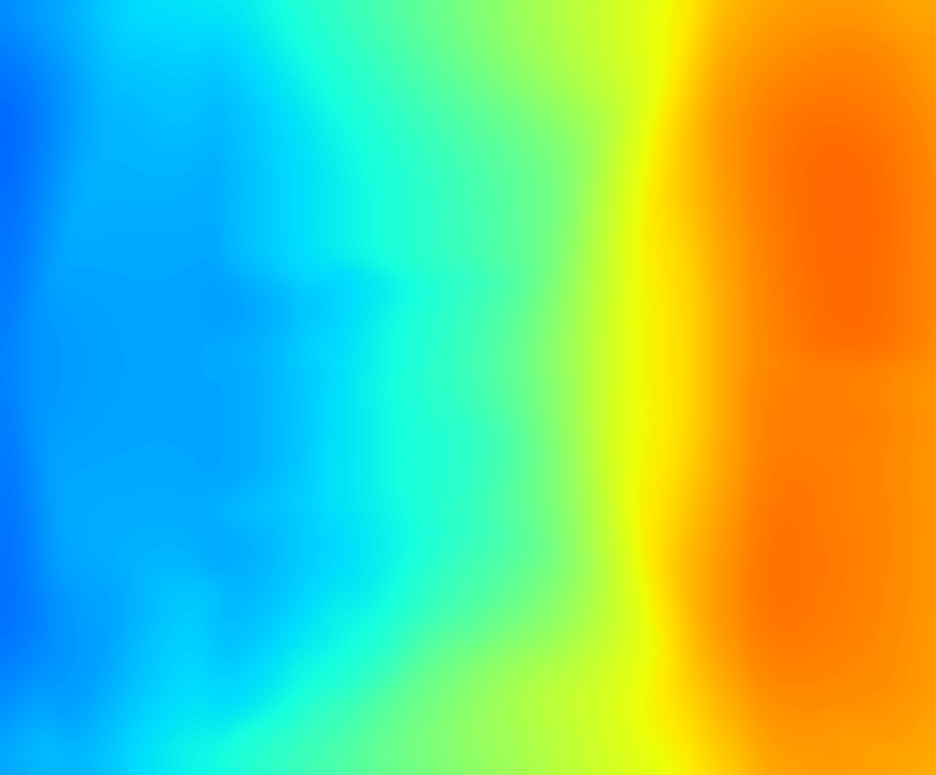}\quad
                         \includegraphics[scale=\figurescaleR]{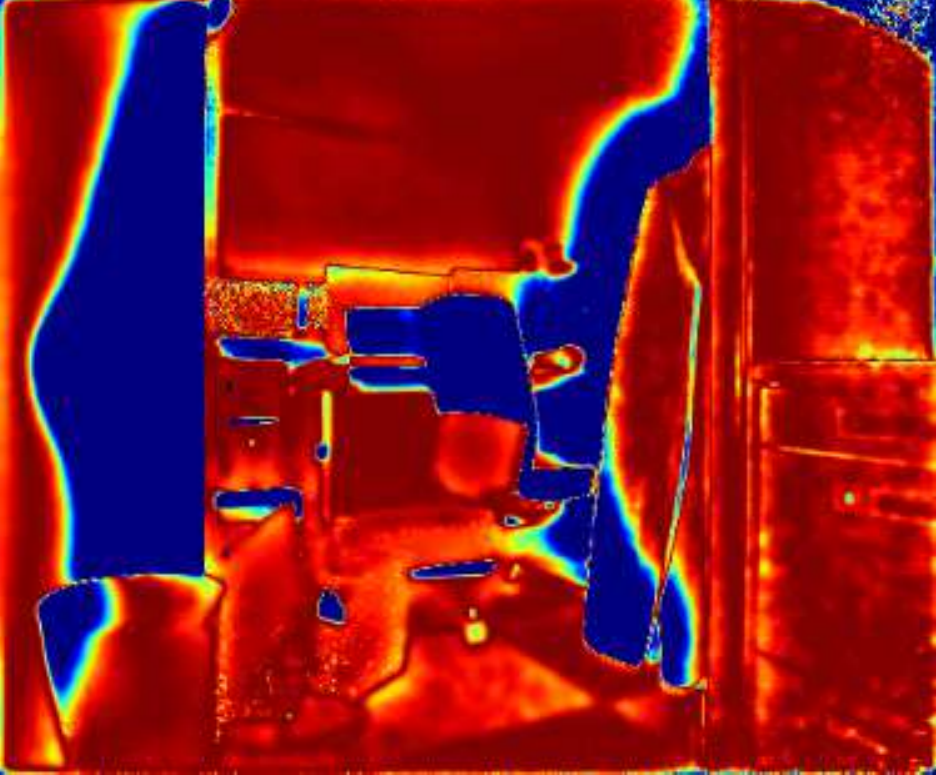}}\\
    			\subfloat[Depth reconstruction]{\includegraphics[scale=\figurescaleR]{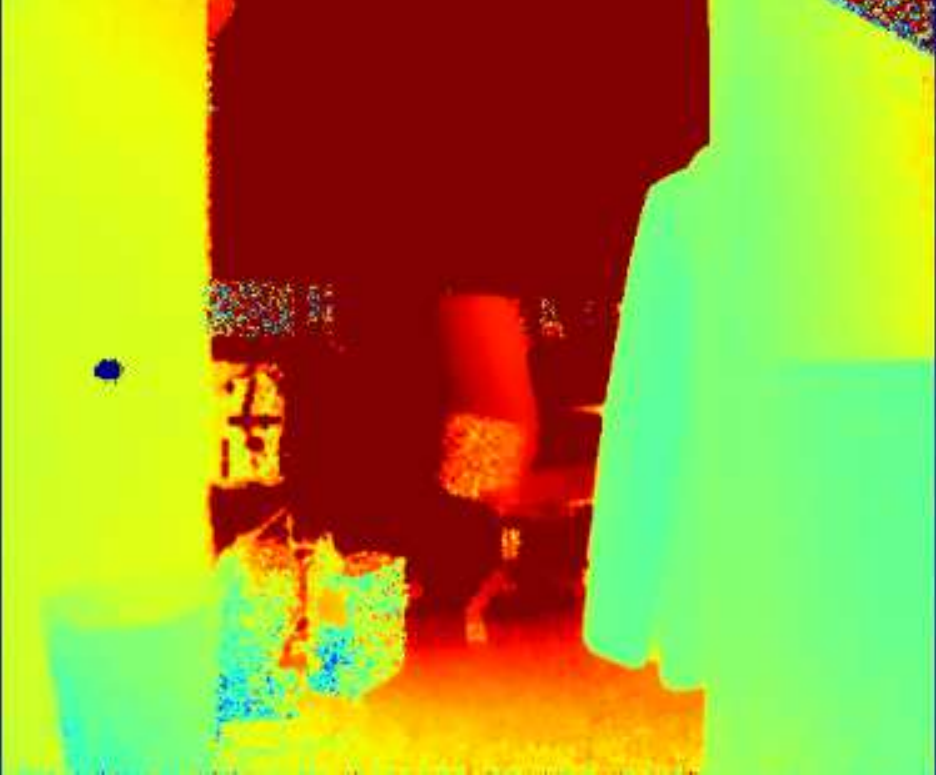}\quad
                          \includegraphics[scale=\figurescaleR]{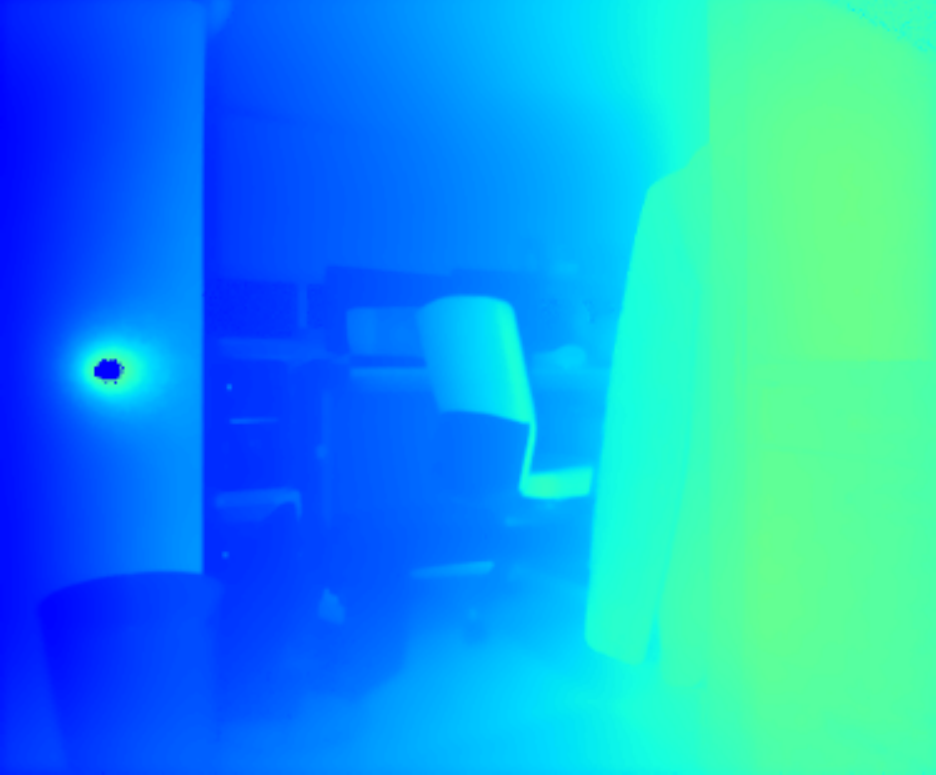}\quad
                          \includegraphics[scale=\figurescaleR]{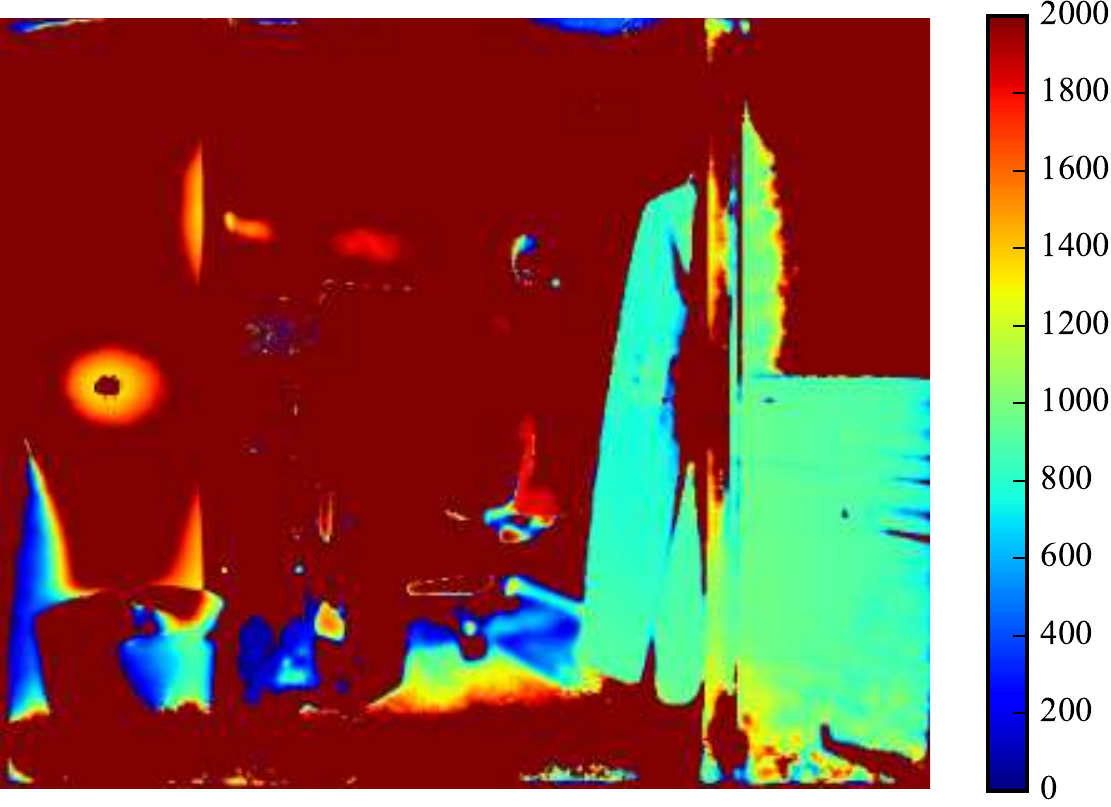}}
		\end{minipage}
	\end{tabular}
\caption{Failure case. (a) Target scene. (b)(c) Result for the amplitude and phase image, respectively. From left to right: input image, estimated scattering component, and weight. (d) Result of the depth reconstruction.}
\label{fig:desk_result}
\end{figure*}

\section{Experiments}\label{sec:experiments}
\subsection{Controlled scene}
\paragraph{Experimental setup.}
We first evaluated the effectiveness of the proposed method in a controlled environment.
The experimental environment is shown in Fig. \ref{fig:environment}.
We set up a fog generator and a Kinect v2 in a closed space sized 186 $\times$ 161 cm with black walls and floor.
The observation of the wall includes only a scattering component because incident light into the wall is absorbed.
The Kinect v2 has three modulation frequencies: 120, 80, and 16 MHz.
We used images obtained with 16 MHz. 
To acquire an amplitude and phase image, we used the source code given by \cite{tanaka17}.
Their code provides the average image of several frames, and we modified the code so that only a single frame was input.
To compensate for high frequency noise, we used a bilateral filter as preprocessing.

The spatial resolution of an image captured by Kinect v2 is 424 $\times$ 512 pixels.
We divided a captured image into 4 $\times$ 4 patches ($K=16$) for local quadratic prior.
In section \ref{sec:prior_of_scattering_component}, we assumed that the camera and the light source are collocated on a line that runs parallel to the horizontal axis of the image.
In practice, the camera and light source are slightly out of alignment.
Although this violates the symmetry of the scattering component, we found that error due to this misalignment is negligibly small.
In our implementation, we defined $\mathbf{F}$ as a matrix that flips an image with respect to the 200th row of the image.
In addition, we did not use the 24 rows of the lower part of the image for the third term of Eq. (\ref{eq:objective}), as these pixels have no information of global symmetrical prior.
For amplitude images, we set the hyperparameters of the objective function as $[\gamma'_1, \gamma'_2, \gamma'_3] = [0.1,0.1,10]$, and the tuning parameter of the funcion $\rho(x)$ is set as $c=4,7$ in the coarse and fine level optimization, respectively.
For phase images, we set $[\gamma'_1, \gamma'_2, \gamma'_3] = [0.01,0.1,50]$ and $c=2,3$.

\paragraph{Experimental result.}
The results are shown in Fig. \ref{fig:medium_result}.
Figure \ref{fig:medium_result}(a) shows the foggy scene, which has five target objects: ``plane,'' ``chair,'' ``desk,'' ``hand,'' and ``duck.''
Figure \ref{fig:medium_result}(b)(c) show the estimation of a scattering component and a object region for the amplitude and phase image, respectively.
The object region depicted here is the IRLS weight before binarization.
As shown, the proposed method effectively estimated the object region via the weight.
Of particular note is that thin regions such as the legs of the chair could be extracted.
The estimation of the scattering component in the object region was also successful.
Figure \ref{fig:medium_result}(d) shows the results of the depth reconstruction, where we show the depth measurement without and with fog, and the reconstructed depth.
The depth measurement in the foggy scene had large error here due to fog.
In contrast, the proposed method reconstructed the object depth correctly. 

We tested the proposed method under different density conditions.
Figure \ref{fig:various_density_result} shows the results under thin fog and thick fog.
In a highly foggy scene like the one in Fig. \ref{fig:various_density_result}(b), the accuracy of depth reconstruction was reduced where a scattering component had a large effect, but the proposed method could estimate an object region and improve the depth measurement regardless of medium density.
The mean depth error on each object with and without our method in scenes under different density conditions is listed in Table \ref{tab:mean_error}; here, we define the ground truth as the measured depth without fog.
As shown, the proposed method could reduce the error significantly regardless of fog density.

\subsection{Realistic scene}
Next, we tested the proposed method in a more realistic scene.
Figure \ref{fig:realistic_result}(a) shows the target scene.
We artificially generated fog in the same manner as the controlled experimental setting, although the scene in Fig. \ref{fig:realistic_result} has neither dark walls nor floor.
Note that the scene has materials with various types of reflectance, including a lamp made from paper, a glossy vase, and a wooden ornament.
The results of the scattering component and object region for the amplitude and phase image are shown in Fig. \ref{fig:realistic_result}(b)(c), respectively, and the estimation of the depth reconstruction is shown in \ref{fig:realistic_result}(d).
The results of this experiment showed that the proposed method could also extract the object region and improve the depth measurement in a scene that has a general background.

\subsection{Experiments with synthesized data}
\paragraph{Synthesized data.}
To investigate the effectiveness in more varied scenes, we evaluated the proposed method with synthesized data.
The procedure of generating the synthesized data is shown in Fig. \ref{fig:synthesized_overview}.
We assume that a scattering component does not depend on object depth, and thus we observed a direct component and a scattering component separately and then combined them into a synthesized image.

To synthesize an image, we have to know the scattering coefficient in a scene.
First, we captured a foggy scene that includes calibration objects, and the region of the calibration objects was masked manually.
After that, we compensated for the defective region by solving Eq. (\ref{eq:irls}) to estimate the scattering component. 
The weight $\mathbf{w}$ in Eq. (\ref{eq:irls}) corresponds to the mask, and we only used the regularization of the global symmetrical prior and smoothing prior.
Using the estimated scattering component, we can compute the direct component of the amplitude using Eq. (\ref{eq:amp_recon}).
Now, the relationship between the amplitude without fog and the attenuated direct component in a foggy scene is given as
\begin{equation}
\alpha_d(u,v) = e^{-2\beta d(u,v)} \hat{\alpha}(u,v),
\end{equation}
where $\hat{\alpha}(u,v)$ is the amplitude at a pixel $(u,v)$ without fog, $d(u,v)$ is the distance between the camera and calibration object, and $\beta$ is a scattering coefficient.
Therefore, we computed the scattering coefficient as
\begin{equation}
\beta = \frac{1}{|\Omega|}\sum_{(u,v)\in \Omega}\frac{\log \hat{\alpha}(u,v) - \log \alpha_d(u,v) }{2d(u,v)},
\end{equation}
where $\Omega$ denotes the set of pixels in the mask and $|\Omega|$ is the number of pixels in $\Omega$.
In the scene in Fig. \ref{fig:synthesized_overview}, a scattering coefficient was computed as $\beta = 3.5 \times 10^{-4}$ /mm.

We also observed a scene without fog, which was used for the direct component after being attenuated by the scattering coefficient.
We combined the attenuated signal and the scattering component to synthesize amplitude and phase images.

\paragraph{Experimental result.}
The results are shown in Figs. \ref{fig:chair_result}, \ref{fig:aisle_result}, and \ref{fig:box_result}.
In each figure, (b)(c) show the estimated scattering component and the IRLS weight for the amplitude and phase image.
We show the output of both the coarse and fine level optimization.
(d) shows the result of the depth reconstruction.
In each scene, the proposed method effectively extracted the object region and estimated the scattering component.
(e) shows the result of just applying the fine optimization.
In the scenes in Figs. \ref{fig:aisle_result} and \ref{fig:box_result}, performing only the fine-level optimization failed to detect the accurate object region, and this deteriorated the scattering component estimation.
In contrast, the coarse-to-fine approach made the region extraction more robust.

We show a failure case in Fig. \ref{fig:desk_result}. 
In a scene that has a large object region, our method was less effective because a quadratic function also fits to values in the object region.
In Fig. \ref{fig:desk_result}, a large textureless object region exists on the left side.
In addition, the global symmetrical prior did not work in this region because the object occupied the pixels from top to bottom in the image.

\begin{figure}[tb]
  \centering
    \includegraphics[width=0.45\textwidth]{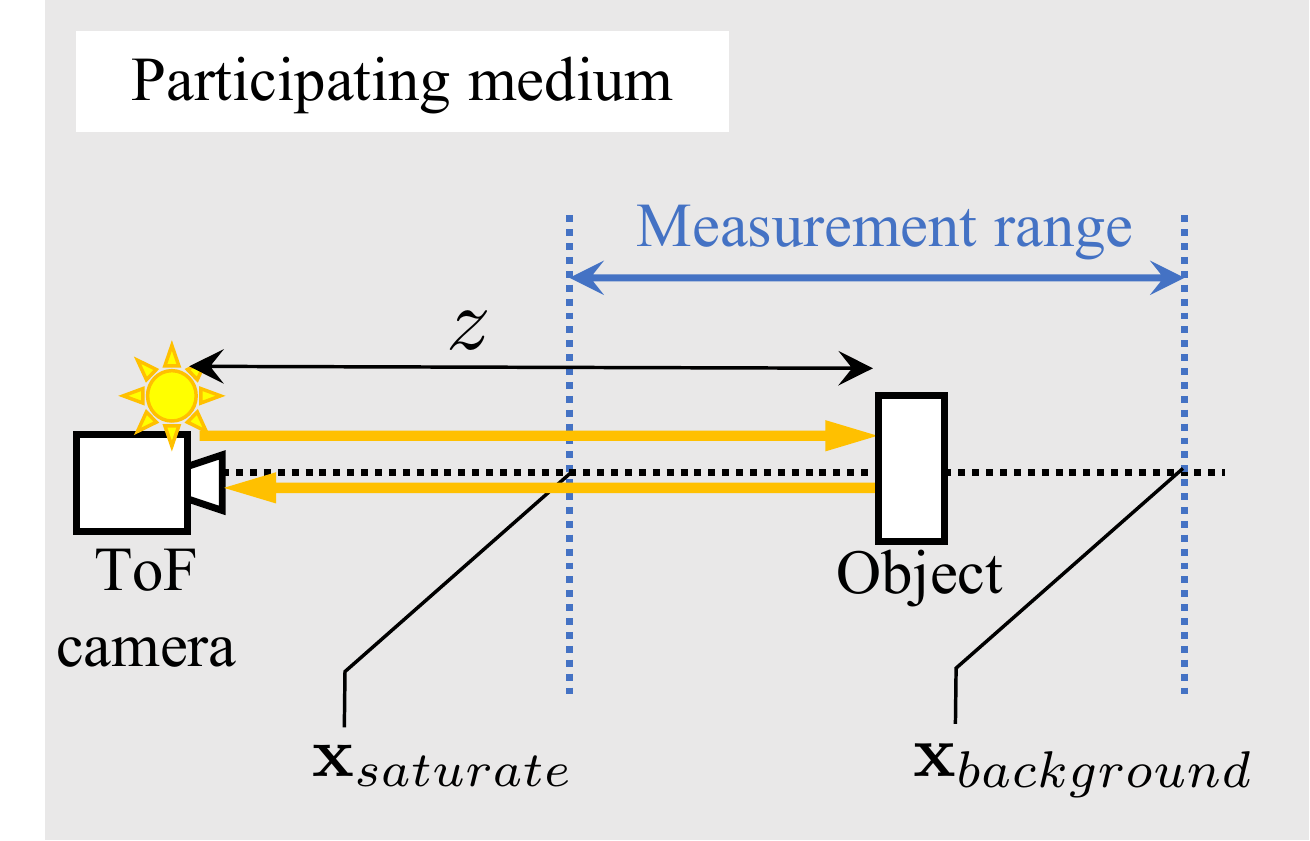}
    \caption{Simulation setting. The measurement range of our method is between $\mathbf{x}_{saturate}$ and $\mathbf{x}_{background}$ where a scattering component is saturated and a direct component remains.}
    \label{fig:measurement_range}
\end{figure}

\begin{figure}[tb]
\centering
	\subfloat[]{\includegraphics[scale=0.38]{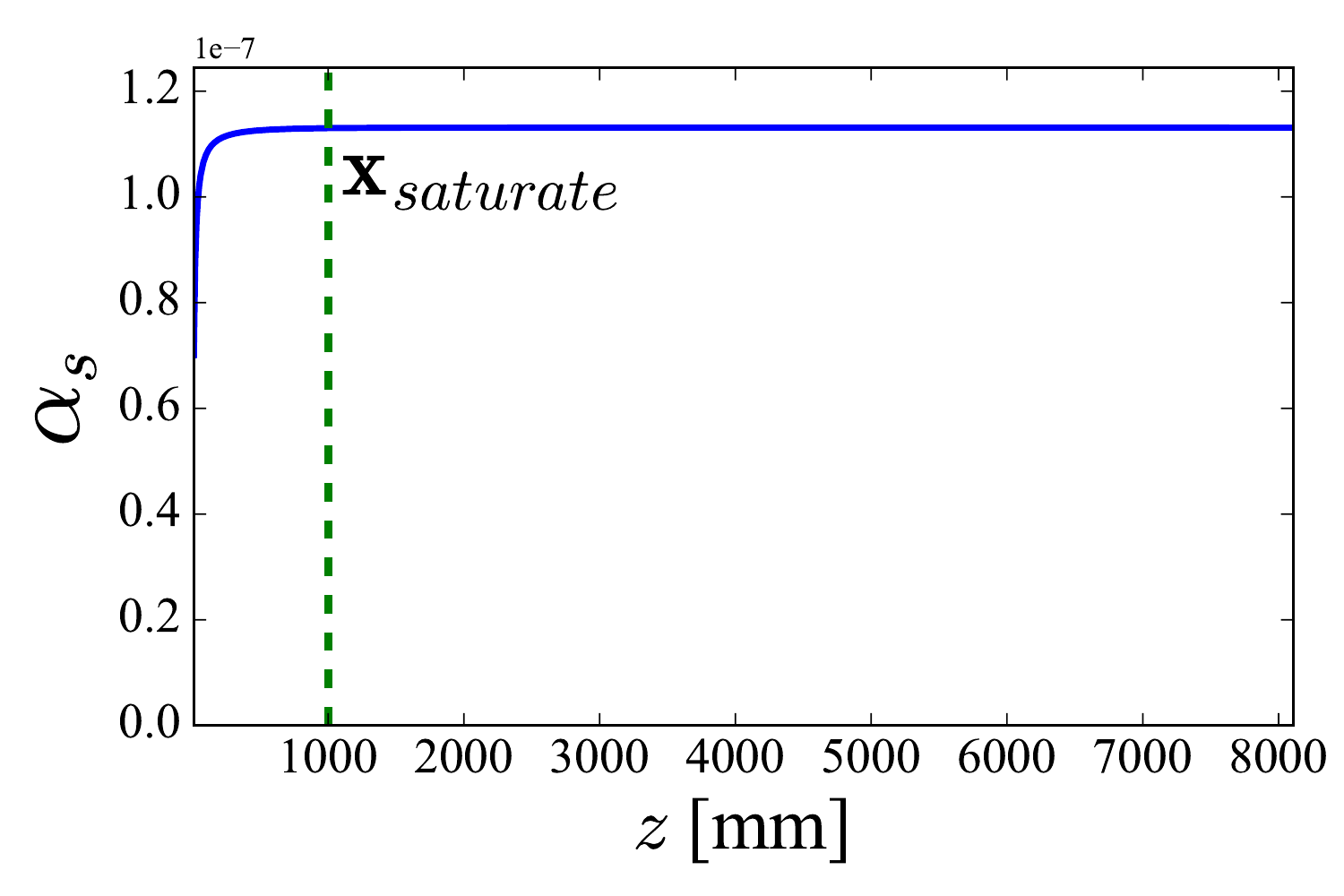}}\\
	\subfloat[]{\includegraphics[scale=0.38]{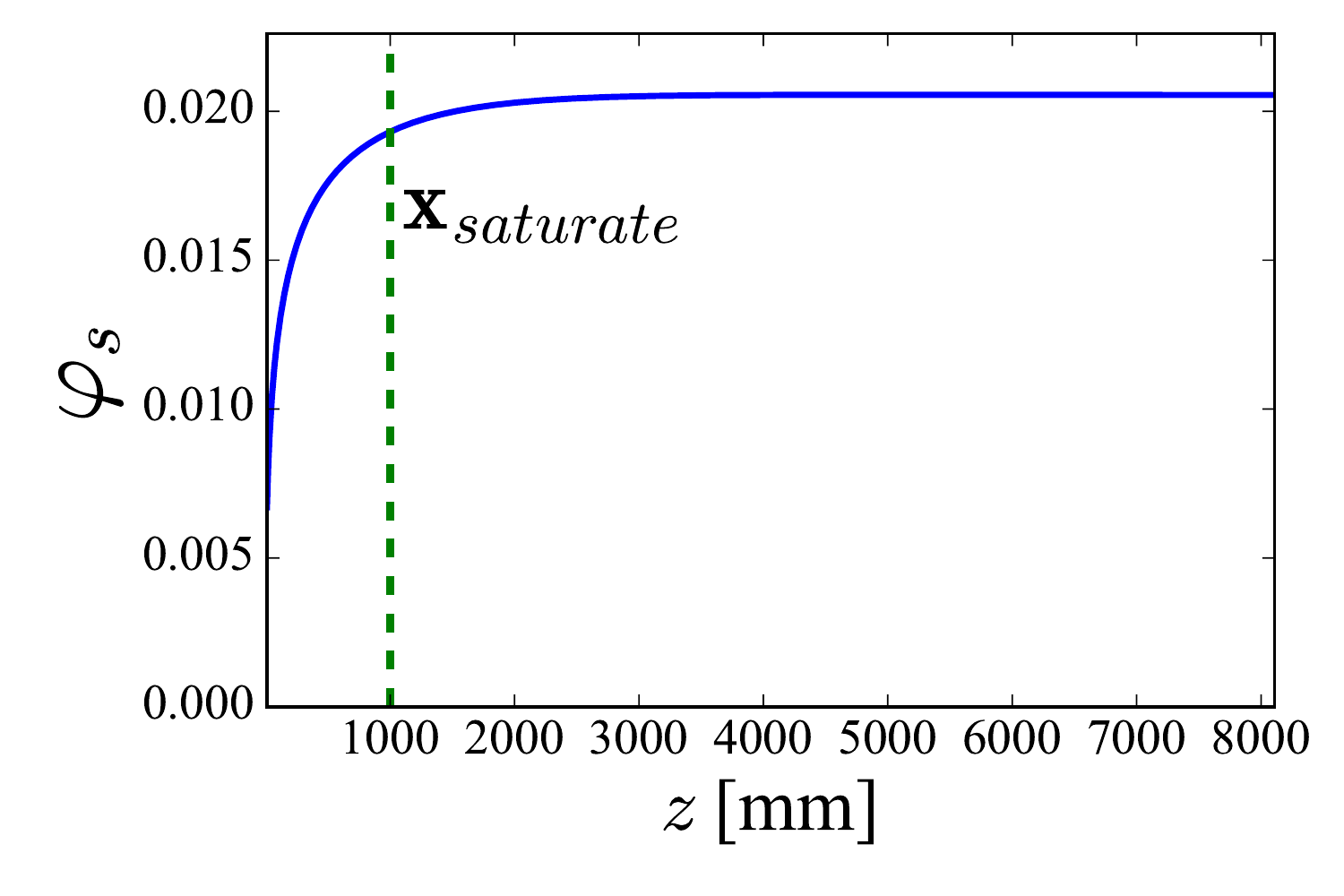}}\\
	\subfloat[]{\includegraphics[scale=0.38]{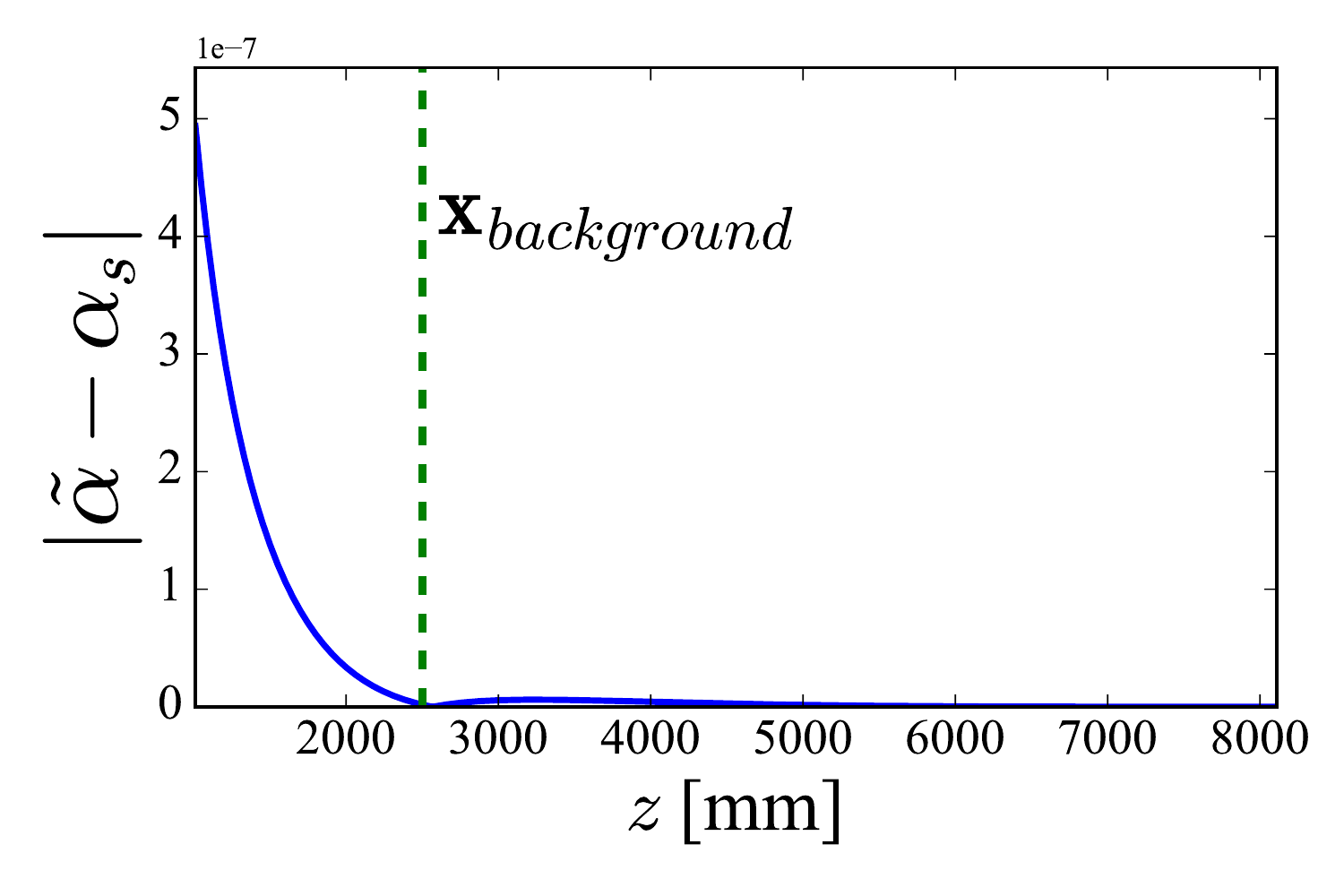}}\\
	\subfloat[]{\includegraphics[scale=0.38]{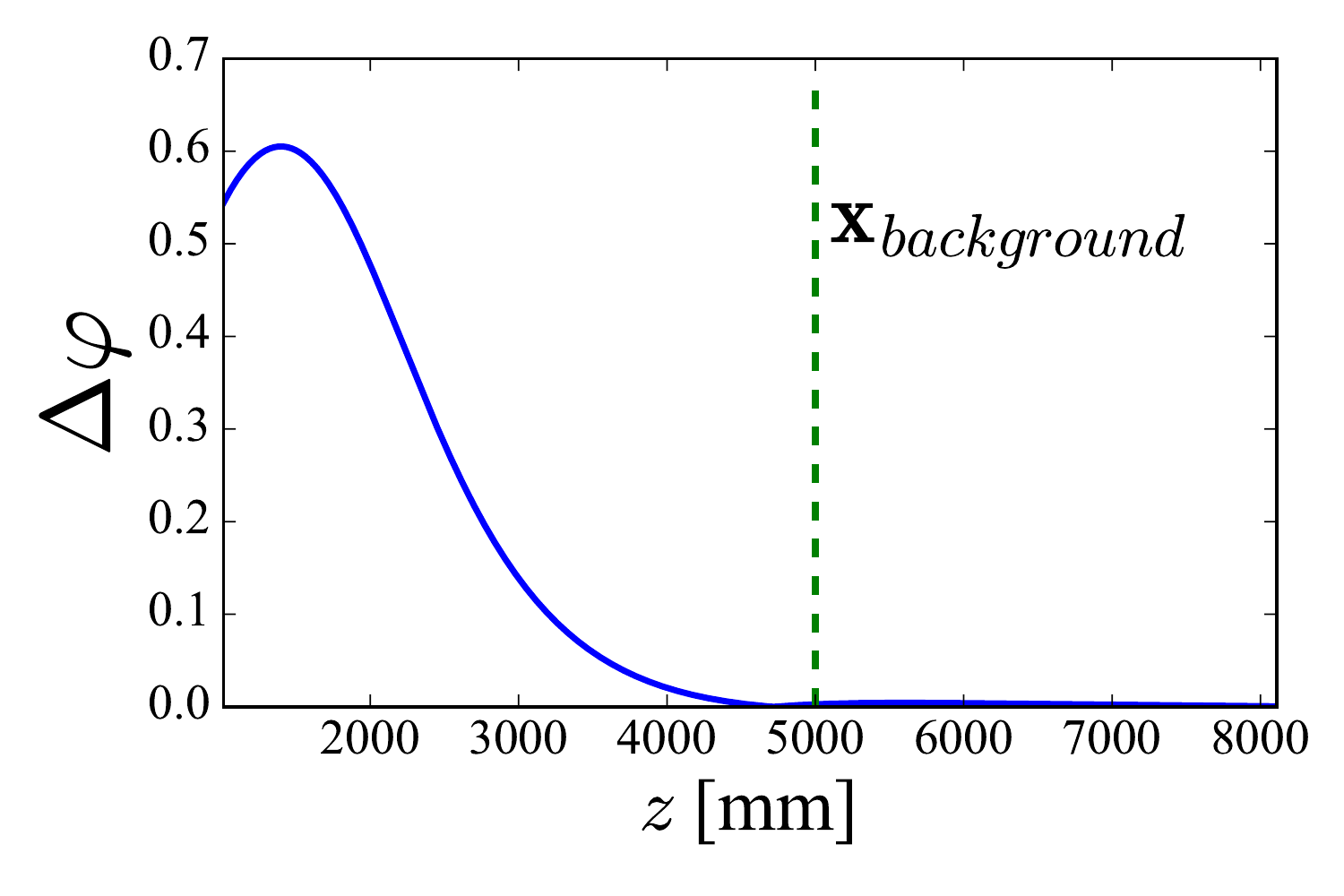}}
\caption{(a)(b) Scattering components for amplitude and phase observed at a scene point whose depth is $z$. (c)(d) Residuals of observation and scattering component, which represent remaining direct components.}
\label{fig:simulation_result}
\end{figure}

\subsection{Discussion of measurement range}\label{sec:discussion_of_measurement_range}
We assume that a scattering component is saturated close to a camera and there exists a background that has only a scattering component.
In this section, we discuss the measurement capability of our method in this context.

As shown in Fig. \ref{fig:measurement_range}, a scene has a saturation point and a background point denoted as $\mathbf{x}_{saturate}$ and $\mathbf{x}_{background}$.
For simplicity, the camera and light source are assumed to be collocated in the same place.
Behind $\mathbf{x}_{saturate}$, a scattering component is constant due to its saturation, while a direct component fades away behind $\mathbf{x}_{background}$.
Therefore, the measurement range of our method is between $\mathbf{x}_{saturate}$ and $\mathbf{x}_{background}$.

We simulated the measurement range to evaluate the capability.
Similarly to the process of synthesizing images, a scattering coefficient was computed for the scene in Fig. \ref{fig:medium_result} to use for the simulation ($\beta = 3.2 \times 10^{-4}$ /mm).
We use the following Henyey-Greenstein phase function for scattering property:
\begin{equation}
P(\theta) = \frac{1}{4\pi}\frac{1-g^2}{(1 + g^2 - 2g \cos \theta)^{3/2}},
\end{equation}
where $\theta$ is a scattering angle.
The parameter $g$ was set as $0.9$ for fog \citep{narasimhan03}.
A scattering component from a camera to depth $z$ is given as
\begin{equation}
\alpha_s(z)e^{j\varphi_s (z)} = \int_{z_{0}}^{z} \frac{1}{z^2} \beta P(\pi) e^{-2 \beta z} e^{j\frac{4\pi f}{c}z}dz.
\end{equation}
We set the starting point of the integral as $z_{0}=10$ mm.
A direct component from depth $z$ is computed as
\begin{equation}
\alpha_d(z)e^{j\varphi_d (z)} = \frac{I}{z^2}e^{-2 \beta z} e^{j\frac{4\pi f}{c}z},
\end{equation}
where $I$ consists of a surface albedo and shading, and we set $I=1$ in this simulation.
The total observation $\tilde{\alpha}(z)e^{\tilde{\varphi}(z)}$ is the sum of these components as with Eq. (\ref{eq:lte}).

Figure \ref{fig:simulation_result} shows the simulation results.
In (a) and (b), the horizontal axis denotes depth $z$ and the vertical axis denotes a scattering component for amplitude and phase.
These figures validate the saturation characteristic.
Meanwhile, in \ref{fig:simulation_result}(c) and (d), the vertical axes denote the residual of the observation and scattering component.
$\Delta \varphi$ is given by the residual angle of $\tilde{\alpha}(z)e^{j\varphi(z)}$ and $\alpha_s(z)e^{j\varphi_s(z)}$ on the complex plane.
These values represent the remaining direct components from depth $z$.

Now, we can set $\| \mathbf{x}_{background}\| = 2500$ mm and $5000$ mm for amplitude and phase from Fig. \ref{fig:simulation_result}(c) and (d) because the direct component is close to zero.
In contrast, in Fig. \ref{fig:simulation_result}(a) and (b), if we set $\|\mathbf{x}_{saturate}\| = 1000$ mm, the estimation error of the scattering component for amplitude due to the saturation assumption can be considered almost zero because $1-\alpha_s(1000)/\alpha_s(8000) \approx 0$, and for phase, the error is $1-\varphi_s(1000)/\varphi_s(8000) \approx 6.0 \%$.

In the experiments with real data, all of the target objects were located between $1000$ mm and $2000$ mm. 
If we assume the measurement range between $\|\mathbf{x}_{saturate}\|=1000$ mm and $\|\mathbf{x}_{background}\| = 2500$ mm, the target objects are located in that range, and from above discussion, we have just $6.0 \%$ error of the scattering component estimation for phase due to the saturation assumption.
As shown in the experiments, we can effectively reconstruct the object depth regardless of the error.
The measurement range depends on the density of a participating medium, and it will get larger under thinner fog.

\section{Conclusion}\label{sec:conclusion}
In this paper, we proposed a method that simultaneously estimates an object region and depth by using a continuous-wave ToF camera.
We leveraged the saturation of a scattering component and the attenuation of a direct component from a distant point in a scene.
The formulation with a robust estimator and the IRLS optimization scheme allow us to estimate the scattering component and object region simultaneously.

The limitation of the proposed method is that we assume the density of the participating medium in the scene to be homogeneous; thus, it cannot be applied to an inhomogeneous medium or a dynamic floating medium.
In these environments, global symmetrical prior does not hold.
In addition, we assume that a scene has a background region, which makes it difficult to apply the method to scenes filled with objects.
In future work, we will address these problems in order to further enhance the real-world applicability of the proposed method.

\begin{acknowledgements}
This work was supported by JSPS KAKENHI Grant Number 18H03263.
\end{acknowledgements}

\bibliographystyle{spbasic}      
\bibliography{template}

\end{document}